\documentclass[11pt,reqno]{amsproc}
\usepackage{titlesec}
\titleformat{\section}
{\normalfont\Large\bfseries}{\thesection}{1em}{}
\titleformat{\subsection}
{\normalfont\large\bfseries}{\thesubsection}{1em}{}
\titleformat{\subsubsection}
{\normalfont\normalsize\bfseries}{\thesubsubsection}{1em}{}
\titleformat{\paragraph}[runin]
{\normalfont\normalsize\bfseries}{\theparagraph}{1em}{}
\titleformat{\subparagraph}[runin]
{\normalfont\normalsize\bfseries}{\thesubparagraph}{1em}{}

\usepackage{booktabs} 
\usepackage[colorinlistoftodos]{todonotes}
\usepackage[T1]{fontenc}
\usepackage{amsmath,amscd,amssymb}
\usepackage{cite}
\usepackage{color}
\usepackage{graphicx}
\usepackage{psfrag,epsfig}
\usepackage{mathrsfs}
\usepackage{siunitx,array,multirow}
\usepackage{rotating}
\usepackage[letterpaper,margin=1in]{geometry}
\usepackage{subfigure}
\usepackage{enumerate}
\usepackage{comment}
\usepackage{lineno}
\usepackage{algorithmic}
\usepackage{algorithm}
\usepackage[small]{caption}
\usepackage{bbold}
\usepackage[normalem]{ulem}
\usepackage[debug=false, colorlinks=true, pdfstartview=FitV, linkcolor=blue, citecolor=blue, urlcolor=blue]{hyperref}
\hypersetup{
    colorlinks=true,
    linkcolor=blue,
    filecolor=magenta,      
    urlcolor=blue,
}
\usepackage{chemarr}
\usepackage[version=3]{mhchem}
\newcommand*\patchAmsMathEnvironmentForLineno[1]{%
  \expandafter\let\csname old#1\expandafter\endcsname\csname #1\endcsname
  \expandafter\let\csname oldend#1\expandafter\endcsname\csname end#1\endcsname
  \renewenvironment{#1}%
     {\linenomath\csname old#1\endcsname}%
     {\csname oldend#1\endcsname\endlinenomath}}%
\newcommand*\patchBothAmsMathEnvironmentsForLineno[1]{%
  \patchAmsMathEnvironmentForLineno{#1}%
  \patchAmsMathEnvironmentForLineno{#1*}}%
\AtBeginDocument{%
\patchBothAmsMathEnvironmentsForLineno{equation}%
\patchBothAmsMathEnvironmentsForLineno{align}%
\patchBothAmsMathEnvironmentsForLineno{flalign}%
\patchBothAmsMathEnvironmentsForLineno{alignat}%
\patchBothAmsMathEnvironmentsForLineno{gather}%
\patchBothAmsMathEnvironmentsForLineno{multline}%
}
\newcommand{\minitab}[2][l]{\begin{tabular}{#1}#2\end{tabular}}

\title[Machine Learning Emulators for Reactive Mixing]{A Comparative Study of Machine Learning Models for Predicting the State of Reactive Mixing}
\author[B.~Ahmmed, et.al.]{B.~Ahmmed$^{1,2,*}$, M.~K.~Mudunuru$^1$, S.~Karra$^1$, S.~C.~James$^{2,3}$, and V.~V.~Vesselinov$^1$ \\
{\small $^1$Computational Earth Science Group, Earth and Environmental Sciences Division, Los Alamos National Laboratory, Los Alamos, NM 87545} \\
\small $^2$Department of Geosciences, Baylor University, Waco, TX 76706 \\
\small $^3$Departments of Geosciences and Mechanical Engineering, Baylor University, Waco, TX 76706 \\
}
\thanks{$^*$Corresponding author:~Bulbul Ahmmed, Email:~\texttt{ahmmedb@lanl.gov}, \texttt{bulbul\_ahmmed@baylor.edu}, Baylor University, Waco, TX 76706}
\date{\today}
\begin{document}
\maketitle
\let\thefootnote\relax\footnotetext{\textbf{Authorship statement:} B.\,Ahmmed developed the framework, ran machine learning models, and drafted the original manuscript. M.\,K. Mudunuru formulated the idea, generated data, and helped to draft the manuscript. S.\,Karra wrote the FEM code to generate data and critically revised the manuscript. S.\,C. James supervised, participated in drafting, and critically revised the manuscript. V.\,V. Vesselinov critically revised the manuscript.}
%
\section*{Abstract}
Mixing phenomena are important mechanisms controlling flow, species transport, and reaction processes in fluids and porous media.
Accurate predictions of reactive mixing are critical for many Earth and environmental science problems such as contaminant fate and remediation, macroalgae growth, and plankton biomass evolution.
To investigate mixing dynamics over time under different scenarios (e.g., anisotropy, fluctuating velocity fields), a high-fidelity, finite-element-based numerical model is built to solve the fast, irreversible bimolecular reaction-diffusion equations to simulate a range of reactive-mixing scenarios.
A total of 2,315 simulations are performed using different sets of model input parameters comprising various spatial scales of vortex structures in the velocity field, time-scales associated with velocity oscillations, the perturbation parameter for the vortex-based velocity, anisotropic dispersion contrast (i.e., ratio of longitudinal-to-transverse dispersion), and molecular diffusion.
Outputs comprise concentration profiles of the reactants and products.
The inputs and outputs of these simulations are concatenated into feature and label matrices, respectively, to train 20 different machine learning (ML) emulators to approximate system behavior.
The 20 ML emulators based on linear methods, Bayesian methods, ensemble learning methods, and multilayer perceptron (MLP), are compared to assess these models.
The ML emulators are specifically trained to classify the state of mixing and predict three quantities of interest (QoIs) characterizing species production, decay (i.e., average concentration, square of average concentration), and degree of mixing (i.e., variances of species concentration).
Linear classifiers and regressors fail to reproduce the QoIs; however, ensemble methods (classifiers and regressors) and the MLP accurately classify the state of reactive mixing and the QoIs.
Among ensemble methods, random forest and decision-tree-based AdaBoost faithfully predict the QoIs.
At run time, trained ML emulators are $\approx10^5$ times faster than the high-fidelity numerical simulations.
Speed and accuracy of the ensemble and MLP models facilitate uncertainty quantification, which usually requires 1,000s of model run, to estimate the uncertainty bounds on the QoIs.
\newline
\newline
\textbf{Keywords:}~Surrogate modeling, machine learning, reaction-diffusion equations, random forests, ensemble methods, artificial neural networks.
%

\section{Introduction}
\label{Sec:S1_Intro}
Reactive-mixing phenomena dictate the distribution of chemical species in fluids (e.g., coastal waters) and subsurface porous media.
Accurate quantification of species concentration is critical to remediation applications such as nuclear remediation, spill distribution, algal-bloom forecasting, etc \cite{R1_Lagneau2019industry,R2_Cama2019acid_water,R3_Rolle2019reactive_fronts,R4_sin2019_multiphase,R5_molins2019_multiscale,lichtner2019reactive,lichtner2015pflotran,chen2018pore,ozturk2015plants,ahmmedMS2015}.
Parameters that influence reactive-mixing in fluids and subsurface porous media include the structure of the flow field (e.g., chaotic advection), fluid injection/extraction (i.e., location of wells, injection/extraction rates), subsurface heterogeneity, dispersion, and anisotropy \cite{VESSELINOV201985ntfk,mudunuru2019mixing}. 
These parameters have variable impacts on important quantities of interest (QoIs) such as species production and decay (e.g., average and squared average species concentrations) and degree of mixing (i.e., variances of species concentrations).
For QoIs, nonlinear partial differential equations are solved using high-fidelity numerical methods (e.g., finite-difference, -element,or -volume methods) that can take hours to days (for $\approx \mathcal{O}(10^6) - \mathcal{O}(10^9)$ degrees-of-freedom) on state-of-the-art, high-performance computing (HPC) machines.
Such computation times preclude real-time predictions, which can be critical to decision making for remediation activities. 
Hence, alternative faster  approaches are needed and machine learning (ML)-based emulators show promise \cite{wu2018seismic,hulbert2019similarity,viswanathan2018graph,srinivasan2018robust,camps2018physics}. 
Here, we build and compare various ML emulators to predict reactive-mixing QoIs.
The ML emulators are trained and tested using data from high-fidelity, finite-element numerical simulations, which expressly reflect the underlying reaction-diffusion physics in anisotropic porous media. 

Given sufficient data, ML models can successfully detect, quantify, and predict different types of phenomena in the geosciences \cite{bergen2019machine,reichstein2019deep}.
Applications include remote sensing \cite{camps2018physics,requena2018predicting}, ocean wave forecasting \cite{james2018_1,james2019_2,james2018_3}, seismology \cite{rouet2017machine,reynen2017supervised,wu2018seismic,magana2017explosion,hulbert2019similarity,yuan2019using}, hydrogeology \cite{BARZEGAR2018697,srinivasan2018robust,viswanathan2018graph}, and geochemistry \cite{RODRIGUEZGALIANO2015804,KIRKWOOD201649,Zuo2017,OONK201580,cracknell2014mapping}. 
ML emulators (also known as surrogate models or reduced-order models) can be fast, reliable, and robust when trained on large datasets \cite{reichstein2019deep,bergen2019machine,salah2018machine}. 
ML emulators are constructed using training data (e.g., features and labels), which include inputs and outputs either from field data, experimental data, high-fidelity numerical simulations, or any combination of these \cite{salah2018machine,brunton2019data}. 
In this paper, we compare emulators based on generalized linear methods \cite{marquaridt1970generalized,hastie2009elements}, Bayesian methods \cite{murphy2012machine,byr_2tipping2001}, ensemble methods \cite{breiman1996bagging,adab4_freund1997}, and an MLP \cite{rumelhart1988learning,rumelhart1988learning,montavon2018methods} to predict various QoIs. 

Previous researchers have used unsupervised and supervised ML methods to reproduce reactive-mixing QoIs.
Vesselinov et~al. \cite{VESSELINOV201985ntfk} used non-negative tensor factorization with custom $k$-means clustering (unsupervised ML) to identify hidden features in the solutions to reaction-diffusion equations.
They determined that anisotropy features (i.e., longitudinal and transverse dispersion)  govern reactive mixing at early to middle times while molecular diffusion controls product formation at late times.
They also quantified the effects of longitudinal and transverse dispersion and molecular diffusion on species production and decay over time. 
Mudunuru and Karra \cite{mudunuru2019mixing} ranked the importance of input parameters/features on reactive-mixing QoIs. 
Also, they developed support vector machine (SVM) and support vector regressor (SVR) emulators to classify the degree of mixing and to predict QoIs. 
However, SVM/SVR training times increase significantly with the size of the training data set \cite{mudunuru2019mixing}.   
To obviate this problem, in the present paper, we build ML emulators whose training times are $\approx 10^{5}$ times faster than SVM and SVR without compromising accuracy.

Specifically, we compare one linear classifier, two Bayesian classifiers, an ensemble classifier, an MLP classifier, seven linear regressors, six ensemble regressors, and an MLP regressor.
Emulator performance is assessed according to training and testing scores, training time, and $R^2$ score on the QoIs from a blind data set.
The blind data set includes six realizations that are not seen during training and testing phases. 
\emph{This study addresses the following questions:~(1) Can ML emulators accurately classify the mixing state of the anisotropic reaction-diffusion system? (2) How accurately do they predict QoIs of reactive mixing? (3) How fast can they be trained? (4) How does each emulator rank overall?}

\section{Governing Equations for Reactive Mixing}
\label{Sec:S2_ROM_GE}
Let $\Omega \subset {\rm I\!R}^{d}$ be an open bounded domain, where $d$ indicates the number of spatial dimensions. 
The boundary is denoted by $\partial \Omega$, which is assumed to be piece-wise smooth. 
Let $\overline{\Omega}$ be the set closure of $\Omega$ and let spatial point $\mathbf{x} \in \overline{\Omega}$.
The divergence and gradient operators with respect to $\mathbf{x}$ are denoted by $\mathrm{div}[\bullet]$ and $\mathrm{grad}[\bullet]$, respectively. 
Let $\mathbf{n}(\mathbf{x})$ be the unit outward normal to $\partial \Omega$. 
Let $t \in \, ]0, \mathcal{I}[$ denote time, where $\mathcal{I}$ is the length of time of interest.
The governing equations are posed on $\Omega \times ]0, \mathcal{I}[$ and the initial condition is specified on $\overline{\Omega}$.
Consider the fast bimolecular reaction where species $A$ and $B$ react irreversibly to yield product $C$:
\begin{align}
  \label{Eqn:Bimolecular_fast_reaction}
    n_{A} \, A \, + \, n_{B} \, B \longrightarrow n_{C} \, C.
\end{align}
The governing equations for this fast bimolecular reaction without volumetric sources/sinks are: 
\begin{subequations}
  \label{Eqn:DRs_for_A_B_C}
  \begin{align}
    \label{Eqn:DRs_for_A}
    &\frac{\partial c_A}{\partial t} - \mathrm{div}[\mathbf{D}
    (\mathbf{x},t) \, \mathrm{grad}[c_A]] =  - 
    n_{\small{A}} \, k_{AB} c_A c_B \quad \mathrm{in} \; \Omega \times ]0, 
    \mathcal{I}[, \\ 
    \label{Eqn:DRs_for_B} 
    &\frac{\partial c_B}{\partial t} - \mathrm{div}[\mathbf{D}
    (\mathbf{x},t) \, \mathrm{grad}[c_B]] =  - 
    n_{\small{B}} \, k_{AB} c_A c_B \quad \mathrm{in} \; \Omega \times ]0, 
    \mathcal{I}[, \\
    \label{Eqn:DRs_for_C} 
    &\frac{\partial c_C}{\partial t} - \mathrm{div}[\mathbf{D}
    (\mathbf{x},t) \, \mathrm{grad}[c_C]] = +
    n_{\small{C}} \, k_{AB} c_A c_B \quad \mathrm{in} \; \Omega \times ]0, 
    \mathcal{I}[, \\
    \label{Eqn:DRs_for_Dirchlet}
    &c_i(\mathbf{x},t) = c^{\mathrm{p}}_i(\mathbf{x},t) \quad 
    \mathrm{on} \; \Gamma^{\mathrm{D}}_{i} \times ]0, 
    \mathcal{I}[ \quad (i = A, \, B, \, C), \\
    \label{Eqn:DRs_for_Neumann}
    & \left(-\mathbf{D} (\mathbf{x},t) \, \mathrm{grad}
    [c_i] \right) \cdot \mathbf{n}(\mathbf{x}) = h^
    {\mathrm{p}}_i(\mathbf{x},t) \quad \mathrm{on} \; 
    \Gamma^{\mathrm{N}}_{i} \times ]0, \mathcal{I}[ 
    \quad (i = A, \, B, \, C), \\
    \label{Eqn:DRs_for_IC}
    &c_i(\mathbf{x},t=0) = c^{0}_i(\mathbf{x}) \quad 
    \mathrm{in} \; \overline{\Omega} \quad (i = A, \, B, \, C).
  \end{align}
\end{subequations}
Traditional numerical formulations for Eqs.~(\ref{Eqn:DRs_for_A})--(\ref{Eqn:DRs_for_IC}) can yield nonphysical solutions for chemical species concentration \cite{nakshatrala2013}. 
Also, when anisotropy dominates, the standard Galerkin formulation produces erroneous concentrations \cite{2015_Mudunuru_etal_ASME,mudunuru2012framework,mudunuru2017mesh,nakshatrala2013}. 
To overcome these problems, a non-negative, finite-element method is used to compute species concentrations \cite{nakshatrala2013}.
This method ensures that concentrations are non-negative and satisfy the discrete maximum principle. 

\subsection{Reaction Tank Problem and Associated QoIs}
\label{SubSec:Reaction_Tank_Problem}
\begin{figure}
  \centering
  \includegraphics{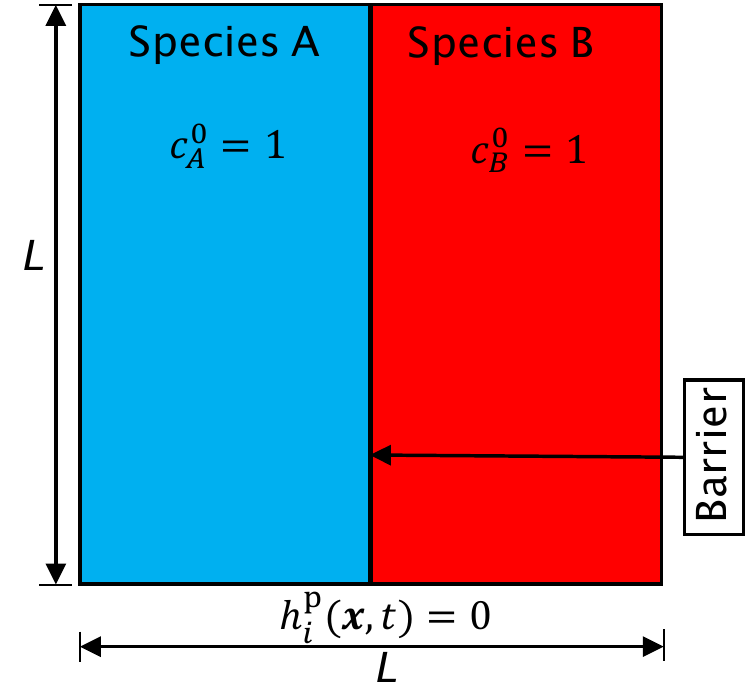}
  \caption{\textrm{\textbf{Model domain for reactive-mixing:}}~Schematic of the initial boundary value problem. 
  $L$, $h_i^\mathrm{p}\left(\mathbf{x},t\right)$, $c_A^0$, and $c_B^0$ are the length of the domain, diffusive flux on the boundary for $i^{\mathrm{th}}$ chemical species, initial concentration of species $A$, and initial concentration of species $B$, respectively. 
  Species $A$ and $B$ were initially on the left and right sides of the domain, respectively.
  Initial concentrations of $A$ and $B$ were 1.0 and mixing commenced for $t > 0$.}
  \label{Fig:reaction_tank}
\end{figure}

Figure~\ref{Fig:reaction_tank} depicts the initial boundary-value problem. 
The model domain is a square with $L = 1$. 
Zero-flux boundary conditions $h_i^\mathrm{p}\left(\mathbf{x},t\right) = 0$ are enforced on all sides of the domain.  
For all chemical species, the non-reactive volumetric source $f_i(\mathbf{x}, t)$ is equal to zero. 
Initially, species $A$ and $B$ are segregated (see Fig.~\ref{Fig:reaction_tank}) and stoichiometric coefficients are $n_A = 1$, $n_B = 1$, and $n_C = 1$. 
The total time of interest is $\mathcal{I} = 1$. 
The dispersion tensor is taken from the subsurface literature \cite{Pinder_Celia,nakshatrala2013}:
\begin{align}
  \label{Eqn:Aniso_Diff_Tensor_Lit}
  \mathbf{D}_{\mathrm{subsurface}}
  (\mathbf{x}) = D_{\mathrm{m}} \mathbf{I} + 
  \alpha_{\mathrm{T}} \|\mathbf{v}\| \mathbf{I} + 
  \frac{\alpha_\mathrm{L} - \alpha_\mathrm{T}}{\|\mathbf{v}\|} 
  \mathbf{v} \otimes \mathbf{v}.
\end{align}
The model velocity field is used to define the dispersion tensor according to stream function \cite{2002_Adrover_etal_CCE_v26_p125_p139,2009_Tsang_PRE_v80_p026305,mudunuru2016enforcing}:
\begin{align}
  \label{Eqn:Div_Free_Stream_Function}
  \psi(\mathbf{x},t) = 
  \begin{cases}
    \frac{1}{2 \pi \kappa_\mathrm{f}} \left[ \sin(2 \pi \kappa_\mathrm{f} x) 
    - \sin(2 \pi \kappa_\mathrm{f} y) + v_0 \cos(2 \pi \kappa_\mathrm{f} y)
    \right] &\; \; \mathrm{if} \; \nu T \leq t < \left( \nu 
    + \frac{1}{2} \right) T  \\
    \frac{1}{2 \pi \kappa_\mathrm{f}} \left[ \sin(2 \pi \kappa_\mathrm{f} x) 
    - \sin(2 \pi \kappa_\mathrm{f} y) - v_0 \cos(2 \pi \kappa_\mathrm{f} x)
    \right] &\; \; \mathrm{if} \; \left( \nu + \frac{1}{2} 
    \right) T \leq t < \left( \nu + 1 \right) T
  \end{cases}.
\end{align}
Using Eq.~\eqref{Eqn:Div_Free_Stream_Function}, the divergence-free velocity field components are:
\begin{align}
  \label{Eqn:Vel_x}
  \mathrm{v}_{x}(\mathbf{x},t) = -\frac{\partial 
  \psi}{\partial \mathrm{y}} = 
  \begin{cases}
    \cos(2 \pi \kappa_\mathrm{f} y) + v_0 \sin(2 \pi \kappa_\mathrm{f} y) 
    &\quad \mathrm{if} \; \nu T \leq t < \left( \nu + 
    \frac{1}{2} \right) T  \\
    \cos(2 \pi \kappa_\mathrm{f} y) &\quad \mathrm{if} \; 
    \left( \nu + \frac{1}{2} \right) T \leq t < 
    \left( \nu + 1 \right) T
  \end{cases},
\end{align}
\begin{align}
  \label{Eqn:Vel_y}
  \mathrm{v}_{y}(\mathbf{x},t) = +\frac{\partial 
  \psi}{\partial \mathrm{x}} = 
  \begin{cases}
    \cos(2 \pi \kappa_\mathrm{f} x) &\quad \mathrm{if} \; 
    \nu T \leq t < \left( \nu + \frac{1}{2} \right) T \\
    \cos(2 \pi \kappa_\mathrm{f} x) + v_0 \sin(2 \pi \kappa_\mathrm{f} x) 
    &\quad \mathrm{if} \; \left( \nu + \frac{1}{2} \right) 
    T \leq t < \left( \nu + 1 \right) T
  \end{cases}.
\end{align}

In Eqs.~\eqref{Eqn:Vel_x}--\eqref{Eqn:Vel_y}, $T$ controls the oscillation of the velocity field from clockwise to anti-clockwise.
$v_0$ is the perturbation parameter of the underlying vortex-based flow field.
Larger values of $v_0$ skew the vortices into ellipses while smaller values of $v_0$ yield circular vortex structures in the velocity field.
$\frac{\alpha_\mathrm{L}}{\alpha_\mathrm{T}}$ controls the magnitude of the anisotropic dispersion contrast.
Smaller values of $\frac{\alpha_\mathrm{L}}{\alpha_\mathrm{T}}$ indicate less anisotropy and vice versa.
The magnitude of $\kappa_\mathrm{f}L$ governs the size of the vortex structures in the flow field \cite{VESSELINOV201985ntfk,mudunuru2019mixing}.
Note that varying $v_0$ does not significantly alter vortex locations.

For reactive-mixing applications, the following QoIs are defined:
\begin{enumerate}
  \item Species production/decay, which can be analyzed by calculating normalized average concentrations, $\overline{c}_i$, and normalized average of squared concentrations, $\overline{c^2}_i$.
    Normalized average of squared concentration, $\overline{c^2}_i$, provides information on the species production/decay as a function of the eigenvalues of anisotropic dispersion.
    For example, see Theorem 2.3 in Reference \cite{mudunuru2019mixing}, which shows that $\overline{c^2}_i$ is bounded above and below by an exponential function of minimum and maximum eigenvalues of anisotropic dispersion. 
    These quantities are:
    \begin{align}
      \label{Eqn:Avg_Conc}
      \overline{c}_i &:= \frac{\langle c_i(t) \rangle}{
      \mathrm{max}\left[\langle c_i(t) \rangle \right]} 
      \quad \mathrm{where} \; \left \langle c_i(t) \right 
      \rangle = \int \limits_{\Omega} c_i(\mathbf{x},t) \, 
      \mathrm{d} \Omega, \\
      \label{Eqn:Avg_Sq_Conc}
      \overline{c^2}_i &:= \frac{\langle c^2_i \rangle}{
      \mathrm{max}\left[\langle c^2_i \rangle \right]} 
      \quad \mathrm{where} \; \left \langle c^2_i (t) 
      \right \rangle = \int \limits_{\Omega} c^2_i(
      \mathbf{x},t) \, \mathrm{d} \Omega.
    \end{align}
  \item Degree of mixing is defined as the variance of concentration:
    \begin{align}
      \label{Eqn:Degree_of_Mixing}
      \sigma^2_{c_i} := \frac{\langle c^2_i \rangle - \langle 
      c_i \rangle^2}{\mathrm{max} \left[\langle c^2_i \rangle 
      - \langle c_i \rangle^2 \right]}.
    \end{align}
\end{enumerate}
Note that the values for $\overline{c}_i$, $\overline{c^2}_i$, and $\sigma^2_{c_i}$ are non-negative and range from 0 to 1 $\forall \quad i = A, B, C$.

\subsection{Feature Generation (Numerical Model Inputs)}
\label{SubSec:Data_ML_Analyses}
First, a 2D numerical model is built using first-order finite-element structured triangular mesh, which has 81 nodes on each side.
A~total of 2,500 high-fidelity numerical simulations are run for different sets of reaction-diffusion model input parameters, of which 2,315 run to completion because certain parameter combinations do not yield to stable solution. 
Each simulation run uses 1,000 time steps ($\mathcal{I} = $ $0.0$ to $1.0$ with a uniform time step of $0.001$).
Features include:~longitudinal-to-transverse anisotropic dispersion ratio $\frac{\alpha_\mathrm{L}}{\alpha_\mathrm{T}}$, molecular diffusion $D_\mathrm{m}$, the perturbation parameter of the underlying vortex-based velocity field $v_0$, and velocity field characteristics scales $\kappa_\mathrm{f} L$ and $T$. 
Specifically, input parameters are:~$v_0 = \left[1, 10^{-1}, 10^{-2}, 10^{-3}, 10^{-4} \right]$, $\frac{\alpha_\mathrm{L}}{\alpha_\mathrm{T}} = \left[1, 10^{1}, 10^{2}, 10^{3}, 10^{4} \right]$, $D_m = \left[10^{-8}, 10^{-3}, 10^{-2}, 10^{-1} \right]$, $\kappa_\mathrm{f}L$ = $\left[1, 2, 3, 4, 5 \right]$, and $T = [1 \times 10^{-4},~2 \times 10^{-4}, 3 \times 10^{-4}, 4 \times 10^{-4}, 5 \times 10^{-4}]$.
$\alpha_\mathrm{T}$ is varied with $\alpha_\mathrm{L}$ held at 1.0.
Five features for each of the 2,315 models with 1,000 time steps formed the feature matrix with dimensions $2,315,000 \times 5$.


\section{Machine Learning Emulators}
\label{Sec:S3_MLE_Eqs}
\subsection{Labels (QoIs) and Preprocessing}
Labels are the QoIs of the 2,315 simulations at each time step yielding label vectors.
Features and labels are concatenated into training and testing data forming a $2,315,000\times 6$ matrix.
For ML classification, the degree of mixing in the system is characterized by four classes representing: Class-1 (well mixed), Class-2 (moderately mixed), Class-3 (weakly mixed), and Class-4 (ultra-weak mixing).
The corresponding $\sigma^2_i$ for these classes are 0.0--0.25, 0.25--0.5, 0.5--0.75, and 0.75--1.0, respectively.
Of course, additional classes could be defined although this would necessitate re-training of ML emulators.
These data are partitioned into training and testing data during construction of the ML emulators and Table~\ref{table:data_size} lists the different partitions.
Each emulator is trained using the three different data partitions and the performance of each assessed.
First, 0.9\% of data are used as training data to identify optimized hyperparameters and other tunable parameters.
Subsequently, emulators using the optimized hyperparameters are validated against 63\% and 81\% of data partitions.

Preprocessing is typically required for ML emulator development.
ML emulators that use the Euclidean norm (e.g., kernel-based methods) must have all features/input parameters of the same scale to make accurate predictions \cite{muller2016introduction,scikit-learn,sklearn_api}.
Common preprocessors are standardization (recasting all feature data into the standard normal distribution $N(0,1)$), normalization (independently scaling each feature between $0$ and $1$), and max-abs scaling (scale and translate individual features such that the maximal absolute value of a feature is 1).
In this study, except for Random Forests (RF), which is agnostic to feature scaling, because features are neither sparse nor skewed and do not have outliers, all data are standardized. 
For polynomial regression, we use the quadratic transformation of the data.

\subsection{Optimization of Hyperparameter and Other Tunable Parameters} 
Every ML emulator learns a function or a set of functions by comparing features and corresponding labels.
During this process, different hyperparameters for each ML emulator control the learning process. 
Some common hyperparameters are regularization, learning rate, and the cost function. 
In addition, there are additional tunable parameters for each ML emulator that also speed the learning process and make a more robust emulator, including the number of training iterations, kernel, truncation value, etc. 
Because hyperparameter optimization is an exhaustive, time-consuming process, 0.9\% of the data (23 simulations) were used with the \texttt{Gridsearch} algorithm in \textsf{Scikit-learn} \cite{scikit-learn_software}, a Python ML package.
Tables~\ref{table:linear_hyperparams} and \ref{table:other_hyperparams} list the hyperparameters for each ML emulator.
Later, 7\% and 9\% of the data were used for validation with 30\% or 10\% reserved as blind data for testing.

Because, ML emulators can introduce bias during training, overfitting is a common phenomenon.
To ameliorate this, $k$-fold cross-validation algorithm is used to avoid bias, to determine optimal computational times, and to calculate reliable variances \cite{kohavi1995study,chou2014machine}.
In this work, 10-fold cross-validation is used \cite{kohavi1995study,chou2014machine}.
First, it subdivides training data into equal ten subsets.
Then, it uses nine sets for training while one set is left for validation, and this process is repeated leaving out each subset once.
The average performance on the 10 withheld data sets are reported along with their variance.

{\small{
\begin{table}
  \centering
	\caption{Summary of training  and testing data partitions used in ML emulator development and testing.
	\label{table:data_size}}
	{\begin{tabular}{|c|c|c|c|c|} \hline
	  \multicolumn{3}{|c|}{\% of input data (No. of simulations)} & 
	  \multicolumn{2}{|c|}{Size of samples for QoIs}   \\
	  \cline{1-5}
	  Training data & Validation data & Testing data & Training & Testing  \\ \hline \hline
	  0.9\% (20)    & 0.1 (3) \% & 99\% (2,292) & 20,150   & 2,291,850  \\ 
	  63\% (1458)  & 7\% (162)   & 30\% (690)  & 1,458,500 & 694,500 \\ 
	  81\% (1875)  & 9\% (208)   & 10\% (230)  & 1,875,500 & 231,500 \\ \hline
	 \end{tabular}}
\end{table} 
}}
\subsection{ML Emulators}
This research applies 20 ML emulators to classify the state of reactive mixing and to predict the reactive-mixing QoIs. 
Among the 20 ML emulators, eight are linear, five are Bayesian, six are ensemble, and one is an MLP.
The eight linear ML emulators are ordinary least square regressor (LSQR), ridge regressor (RR), lasso regressor (LR), elastic-net regressor (ER), Huber regressor (HR), polynomial, logistic regression (LogR), and kernel ridge (KR).
Among the linear emulators, only LogR is a classifier.
The five Bayesian techniques are -- Bayesian ridge (BR), Gaussian process (GP), na\"{i}ve Bayes (NB), linear discriminant analysis (LDA), and quadratic discriminant analysis (QDA).
Among these Bayesian emulators, LDA and QDA are classifiers and remaining are regressors.
The six ensemble ML emulators are bagging, decision tree (DT), random forest (RF), AdaBoost (AdaB), DT-based AdaB, and gradient boosting method (GBM).
Among the six ensemble emulators, RF is used as both classifier and regressor. 
MLP is also used as both classifier and regressor.
\subsection{Linear ML Emulators}
Linear ML emulators tend to fit a straight line to the labels.
Each linear emulators' equation is listed in Table~\ref{table:lin_equations} along with its corresponding cost function. 
A~brief mathematical description of each linear ML emulator is explained at Appendix~A.
The equation for polynomial regression is not listed here because it applies the LSQR formula to quadratic-scaled data.
For LSQR and polynomial regressor, we optimize intercept. 
For RR, $\alpha_2$ and $\epsilon$ (tolerance/threshold) are optimized.
For LR, $\alpha_1$, $\epsilon$, and maximum iteration number are optimized.
For ER, $\alpha_1$, $\alpha_2$, $\epsilon$, $l_\mathrm{1}$ ratio, and maximum iteration number are optimized.
For HR, $\alpha_1$, $\epsilon$, and maximum iteration number are optimized.
Optimized hyperparameters and other tunable parameters (bolded) for linear ML emulators are listed in Table~\ref{table:linear_hyperparams}.
For Logistic regression, multi-class (binary or multi-class), solver, $\epsilon$, and maximum number of iterations are optimized and corresponding settings are presented in Table~\ref{table:linear_hyperparams}.
Tested solvers include Newton's method, limited memory large-scale bound constrained (LBFGS) solver, and the stochastic average gradient (SAG) solver.
For KR, $\alpha_1$, $\lambda$, and kernels are optimized (see, Table~\ref{table:linear_hyperparams}).
{\small{
\begin{table}
  \centering
	\caption{Equation and cost function of linear emulators.}
	\label{table:lin_equations}
	{\begin{tabular}{|c|c|c|} \hline
	  Emulator & Equation & Cost function  \\ \hline \hline
	  LSQR & $\hat{y}\left(\mathbb{w}, \mathbb{x}\right)  =  w_0 + w_1x_1 + \cdots + w_nx_n = \mathbb{x}\cdot \mathbb{w}$ &
	  $L_{\mathrm{lsqr}} = \underset{\mathbb{w}}{\mathrm{min}}||\mathbb{X}\mathbb{w} - \mathbb{y} ||_2^2 $ \\ \hline
	  RR & Same as above &
	  $L_{\mathrm{RR}} = \underset{\mathbb{w}}{\mathrm{min}}||\mathbb{X}\mathbb{w} - \mathbb{y} ||_2^2 + \alpha_2 || \mathbb{w} ||_2^2 $ \\ \hline
	  LR & Same as above &
	  $L_{\mathrm{LR}} = \underset{\mathbb{w}}{\mathrm{min}}||\mathbb{X}\mathbb{w} - \mathbb{y} ||_2^2 + \alpha_1 || \mathbb{w} ||_1 $ \\ \hline
	  ER & Same as above &
	  $L_{\mathrm{ER}} = \underset{\mathbb{w}}{\mathrm{min}}||\mathbb{X}\mathbb{w} - \mathbb{y} ||_2^2 + \alpha_1 || \mathbb{w} ||_1 + \alpha_2 || \mathbb{w} ||_2^2 $ \\ \hline
	  HR & Same as above &
	  $L_{\mathrm{HR}} = \underset{\mathbb{w}}{\mathrm{min}}||\mathbb{X}\mathbb{w} - \mathbb{y} ||_2^2 + \alpha_1 || \mathbb{w} ||_1 + \Sigma \sum_{i=1}^m \left[1 + H_\epsilon \left(\frac{\mathbb{x}_i \cdot \mathbb{w} - y_i}{\Sigma}\right) \right]$ \\ \hline \hline
	  LogR & $l = \log_\mathrm{b}\left( \frac{p}{1 - p}\right) = w_0 + w_1 x_1 + w_2 x_2$ &
	  $L_{\mathrm{cross-entropy}} = -\frac{1}{n} \sum_{n=1}^n \left[y_n \log\left(p_n\right) + \left(1 - y_n\right) \log\left(1 - p_n\right) \right]$ \\ \hline
	  KR & $\mathcal{K}_{\mathrm{RBF}}\left(\mathbb{x_1}, \mathbb{x_2}\right)  =  \mathrm{exp}\left(-\lambda \left|\left|\mathbb{x_1} - \mathbb{x_2}\right|\right|^{2} \right)$ &
	  $L_{\mathrm{squared}} = \left(y - \hat{y}\right)^2$ \\ \hline
	 \end{tabular}}
\end{table} 
}}
\subsection{Bayesian ML Emulators}
Bayesian ML emulators apply Bayes' rule to learn function from labels to predict equivalent label.
Equations for Bayesian ML emulators are listed in Table~\ref{table:other_equations}. 
Also, a brief mathematical description of each Bayesian ML emulator is explained in Appendix~A.
For BR, $\beta$, $\mathrm{\omega}$, maximum iterations, and $\epsilon$ are the hyperparameters and their optimized values are shown in bold in Table~\ref{table:other_hyperparams}.
For GP, kernel is optimized and its best is listed in Table~\ref{table:other_hyperparams}.
In NB, only priors and variance smoothing are hyperparameters.
For LDA, solver is optimized; solvers include singular value decomposition (SVD), LSQR, eigen value decomposition.
Among these three, SVD is fastest.
For QDA, only tolerance is optimized and best value is $10^{-4}$. 

\subsection{Ensemble Emulators}
If the relationship between features and label is nonlinear, linear ML emulators are not expected to perform well. 
Instead, nonlinear ML emulators such as an MLP and ensemble methods should work better. 
Ensemble methods bootstrap (random sampling with replacement) data to develop different tree models/predictors. 
Each label is used with replacement as input for developing individual models; therefore, tree models have different labels based on the bootstrap process. 
Because bootstrapping captures many uncorrelated base learners to develop a final model, it reduces variance; resulting in a reduced prediction error.
Also, in ensemble models, many different trees predict the same target variable; therefore, they predict better than any single tree alone.

Ensemble techniques are further classified into Bagging (bootstrapping aggregating) and Boosting (form many weak trees/learners into a strong tree).
While bagging emulators work best with strong and complex trees (e.g., fully developed decision trees), boosting emulators work best with weak models (e.g., shallow decision trees).
In this study, several averaging/bagging and boosting ensemble emulators are explored to classify and predict reactive mixing.
The averaging emulators include bagging and RF while boosting emulators include AdaBoost (AdaB), DT-based AdaB, and gradient boosting method (GBM).
%
{\small{
\begin{table}
  \centering
	\caption{Equation and cost function of Bayesian, ensemble, and MLP emulators where $L_{\mathrm{lml}} = -\frac{1}{2}\left(\log\|\omega^{-1}\mathbb{I} + \mathbf{\Phi} \mathbf{A}^{-1} \mathbf{\Phi}^\intercal \| + \mathbf{\Psi}^\intercal\left(\mathrm{\omega}^{-1}\mathbb{I} + \mathbf{\Phi} \mathbf{A}^{-1} \mathbf{\Phi}^\intercal\right)^{-1}\mathbf{\Psi}\right) + \sum_{i=0}^N \left(o \log\beta_i - \mathrm{r}\beta_i\right) + u\log\omega - w\omega$.}
	\label{table:other_equations}
	{\begin{tabular}{|c|c|c|} \hline
	  Emulator & Equation & Cost function  \\ \hline \hline
	  BR & $p\left(\mathbb{w}|\mathrm{\omega} \right)  =  N \left(\mathbb{w}|0, \mathrm{\omega}^{-1} \mathbf{\mathbb{I}}\right)$ & $L_{\mathrm{lml}}$ \\ \hline
	  GP & $p\left(\mathbb{y}|\mathbb{X}, \mathbb{w}, \beta \right)  =  N\left(\mathbb{y}|\mathbb{X}\mathbb{w}, \beta \right)$ & $L_{\mathrm{lml}} $ \\ \hline
	  NB & $p\left(x_i\,|\,y\right)  =  \frac{1}{\sqrt{2 \pi \sigma_y^2}} \exp{\left(-\frac{\left(x_i - \mu_y\right)^2}{2 \sigma_y^2}\right)} $ &
	  Maximum $p(x)$ \\ \hline
	  LDA & $p\left(y = k|\mathbb{x}\right)  = \frac{1}{\left(2 \pi \right)^{j/2} \left| \mathrm{det}[\sum_k] \right|^{1/2}} \exp{\left(-\frac{1}{2} \left(\mathbb{x} - \mu_k \mathbb{1} \right) \cdot ({\scriptstyle\sum}_k)^{-1} \left(\mathbb{x} - \mu_k \mathbb{1} \right)\right)}$ &
	  $L_{\mathrm{cross-entropy}} $ \\ \hline
	  QDA & $p\left(y = k|\mathbb{x}\right)$ & $L_{\mathrm{cross-entropy}} $ \\ \hline  \hline
	  DT & $G\left(Q, s\right) = \frac{n_{\mathrm{left}}}{\mathfrak{T}_m} H\left(Q_{\mathrm{left}}(s)\right) + \frac{n_{\mathrm{right}}}{\mathfrak{T}_m} H\left(Q_{\mathrm{right}}(s)\right)$ &
	  $L_{\mathrm{MSE}} = \frac{1}{n} \sum_{i=1}^n \left( y_i - \hat{y_i}\right)^2$ \\ \hline
	  Bagging & $\hat{f}  =  \sum_{i=1}^M f_i (\mathbb{x}_i)$ & $L_{\mathrm{MSE}}$ \\ \hline 
	  RF & $\hat{f}_{rf}^M = \frac{1}{M} \sum_{m=1}^M \mathfrak{T} \left( \mathbb{x} \right)$ & $L_{\mathrm{MSE}}$ \\ \hline
	  AdaB & $f = \mathrm{inf}\left[y \in \mathbb{y}: \sum_{m:h_m \leq y} \mathrm{log}\left(\frac{1}{\theta_m}\right) \geq \frac{1}{2} \sum_m \mathrm{log}\left(\frac{1}{\theta_m}\right) \right]$ & 
	  $L_\mathrm{square-loss} = \frac{\left|y_i\left(\mathbb{x}\right) - y_i\right|^2}{J^2}$ \\ \hline
	  DT-based AdaB & Use DT regression with AdaB procedure & $L_\mathrm{square-loss}$ \\ \hline
	  GBM & $f = \sum_{m=1}^M \gamma_m h_m \left(\mathbb{x}_m\right)$ & $L_{\mathrm{lsqr}} = \underset{w}{\mathrm{min}} \sum_{i=1}^n \left(\mathbb{x}_i w - y_i \right)^2$ \\ \hline  \hline
	  MLP & $a_n^{\left(l\right)} = F\left(\sum_{\mathscr{K}=1}^{\mathscr{N}_{l-1}}w_{\mathscr{K},n}^{\left(l\right)} a_\mathscr{K}^{\left(l-1\right)} + b_n^{\left(l\right)}\right)$ & ReLU \\ \hline
	\end{tabular}}
\end{table} 
}}

For DT, maximum tree depth, maximum number of features, and minimum sample splitting are optimized and best settings are listed in bold in Table~\ref{table:ensemble_hyperparams}.
In Bagging, tree number, bootstrapping, and maximum number of features are optimized and their best settings are prescribed in Table~\ref{table:ensemble_hyperparams}.
In RF, maximum depth of tree, tree number in forest, minimum sample splitting number, bootstrapping, and maximum feature number are optimized and their best settings are listed in bold in Table~\ref{table:ensemble_hyperparams}.
For AdaB and DT-based AdaB, number of trees, loss function, and $\gamma$ are optimized and their best settings are in bold in Table~\ref{table:ensemble_hyperparams}. 
In GBM, number of trees, sub-sampling, and $\gamma$ are optimized and their best settings are prescribed in bold in Table~\ref{table:ensemble_hyperparams}.
For MLP, number of hidden layers, activation function, $\alpha$, $\gamma$, solver, and maximum number of iteration are optimized and their best values are bold in Table~\ref{table:ANN_hyperparams}.
Solvers in MLP are adaptive momentum (Adam), LBFGS, and SGD.
\subsection{Performance Metrics}
Training time and ${R}^2$ score are performance metrics for each emulator. 
Training time should be fast while ${R}^2$ measures the correlation between $\mathbb{y}$ and $\hat{\mathbb{y}}$.
For $n$ pairs of data points, the ${R}^2$ score is: 
\begin{equation}
    R^2 = \frac{\sum \limits_{i=1}^n \left(y_i - y_{\mathrm{mean}}\right)^2 - \sum \limits_{i=1}^n \left(y_i -\hat{y}_i\right)^2}{\sum \limits_{i=1}^n \left(y_i - y_{\mathrm{mean}}\right)^2},
\end{equation}
which ranges from 0 to 1 for the worst and best predictions, respectively.
For classification, the performance metrics is defined as:
\begin{equation}
    \mathrm{Accuracy} = \frac{1}{n_\mathrm{samples}} \sum_{i=1}^{n_\mathrm{samples}} \mathbb{1}(y) \left(\hat{y_i} = y_i \right),
\end{equation}
where $\mathbb{1}(y)$ is the indicator function \cite{hastie2009elements}.
%
 {\small{
\begin{table}[!ht]
    \centering
    \caption{Hyperparameters and tunable parameters for generalized linear ML emulators, logistic regression, and KR with the best parameters in bold.} 
    \begin{tabular}{|cll|}\hline
      \textbf{Emulator} & \textbf{Hyperparameter and tunable parameter} & \textbf{Sought range} \\ \hline \hline 
      \multirow{1}*{\minitab[c]{LSQR}} & Fit intercept    & \textbf{True}, False \\\hline
      \multirow{2}*{\minitab[c]{RR}} & $\alpha_2$  & \textbf{1.0}, 100, 1,000  \\
        & Max. no. of iterations      & $\boldsymbol{50}$, 300, 1,000 \\ \hline
      \multirow{3}{*}{\minitab[c]{LR}}  & $\alpha_1$  & $10^{-1}$, $10^{-2}$, $10^{-3}$, $\boldsymbol{10^{-4}}$ \\
        & $\epsilon$             & $\boldsymbol{10^{-3}}$, $10^{-4}$ \\
        & Max. no. of iterations   & 50, 100, 300, $\boldsymbol{1,000}$ \\ \hline
      \multirow{5}{*}{\minitab[c]{ER}} & $\alpha_1$ and $\alpha_2$  & $10^{-1}$, $10^{-2}$, $10^{-3}$, $\boldsymbol{10^{-4}}$ \\
        & $\epsilon$         & $10^{-2}$, $\boldsymbol{10^{-3}}$, $10^{-4}$ \\ 
        & $l_1$ ratio        & 0.1, \textbf{0.5}, 1.0 \\
        & Max. no. of iterations  & $10^2$, $10^3$, $\boldsymbol{10^4}$ \\ 
        & Tolerance          & $10^{-2}$, $10^{-3}$, $\boldsymbol{10^{-4}}$ \\\hline
      \multirow{3}*{\minitab[c]{HR}} & $\alpha_1$  & $10^{-1}$, $10^{-2}$, $10^{-3}$,$\boldsymbol{10^{-4}}$ \\
        & $\epsilon$                  & $10^{-3}$, $\boldsymbol{10^{-4}}$, $10^{-5}$ \\ 
        & Max. no. of iterations      & 10, \textbf{50}, 100 \\ \hline
      \multirow{4}*{\minitab[c]{LogR}}   & Multi-class           & OVR, \textbf{Multinomial} \\
        & Solver                & Newton-cg, lbfgs, \textbf{SAG}   \\ 
        & $\epsilon$             & $10^{-3}$, $\boldsymbol{10^{-4}}$, $10^{-5}$, \\ 
        & Max. no. of iterations     & 10, \textbf{50}, 100, 200, 300 \\ \hline
      \multirow{3}*{\minitab[c]{KR}} & $\alpha$  & $10^{-2}$, $10^{-3}$, $\boldsymbol{10^{-4}}$ \\
        & $\lambda$              & $\boldsymbol{1}$, 2, 3 \\
        & Kernel                 & linear, polynomial, \textbf{RBF} \\ \hline 
    \end{tabular}
    \label{table:linear_hyperparams}
\end{table}
 }}
%
{\small{
\begin{table}[!ht]
    \centering
    \caption{Hyperparameters and tunable parameters for Bayesian emulators where bold parameters are best suited parameters. Exponential sine squared $\left(\mathscr{K}(x, x^{'}) = \sigma^2 \mathrm{exp}\left(-2 \mathrm{sin}^2 \left(\pi |x - x^{'}|/\mathrm{p}\right)/\mathrm{l}^2\right)\right)$ is parameterized by a length-scale parameter ($\mathrm{l}$) >0 and a periodicity ($\mathrm{p}$) >0.}
    \begin{tabular}{|cll|}\hline
      \textbf{Emulator} & \textbf{Hyperparameter and tunable parameter} & \textbf{Sought range} \\ \hline \hline 
      \multirow{2}*{\minitab[c]{BR}} & No. of iterations    & \textbf{100}, 200, 300 \\
        & $\epsilon$           & $10^{-2}$, $10^{-3}$, $\boldsymbol{10^{-4}}$ \\\hline
      \multirow{1}*{\minitab[c]{GP}} & Kernel & Exponential sine squared, \textbf{RBF} \\ \hline
      \multirow{2}*{\minitab[c]{NB}} & Priors  & True, \textbf{None} \\
        & Variance smoothing                & $10^{-7}$, $10^{-8}$, $\boldsymbol{10^{-9}}$ \\ \hline 
      \multirow{1}*{\minitab[c]{LDA}} & Solver  & \textbf{SVD}, LSQR, Eigen \\ \hline
      \multirow{1}*{\minitab[c]{QDA}} & Tolerance   & $10^{-3}$, $\boldsymbol{10^{-4}}$, $10^{-5}$ \\ \hline
    \end{tabular}
    \label{table:other_hyperparams}
\end{table}
}}
%
{\small{
\begin{table}[!ht]
  \caption{Hyperparameters and tunable parameters for ensemble ML emulators with the best parameters in bold.}
    \begin{tabular}{|cll|}\hline
      \textbf{Emulator} & \textbf{Hyperparameter and tunable parameter} & \textbf{Sought range} \\ \hline \hline 
      \multirow{3}*{\minitab[c]{DT}} & Maximum depth     & 2, 3, \textbf{None} \\
        & Max. no. of features      & 3, 4, $\boldsymbol{5}$ \\
        & Min. sample splits  & $\boldsymbol{5}$ 3, 4 \\ \hline 
      \multirow{3}{*}{Bagging} & No. of trees  & $\boldsymbol{100}$, 200, 500 \\
        & Bootstrap              & \textbf{True}, False \\
        & Max. no. of features       & 3, 4, $\boldsymbol{5}$ \\\hline
      \multirow{5}*{\minitab[c]{RF}} & Maximum depth & 2, 3, \textbf{None} \\
        & No. of trees in the forest & 250, $\boldsymbol{500}$, 1,000 \\
        & Bootstrap              & True, \textbf{False} \\
        & Max. no. of features in a tree   & 3, \textbf{4}, 5 \\
        & Min. sample splits               & \textbf{2}, 3, 4  \\ \hline
      \multirow{3}{*}{AdaB} & No. of trees  & \textbf{100}, 200, 300 \\
        & Loss function type         & linear, \textbf{square}, exponential \\
        & $\gamma$         & 0.1, 0.5, 0.75,\textbf{1.0} \\ \hline 
      \multirow{3}*{\minitab[c]{DT-based \\ AdaB}} & No. of trees  &  \textbf{100}, 200, 500 \\
        & Loss function type          & linear, \textbf{square}, exponential \\
        & $\gamma$          & 0.1, 0.5, \textbf{1.0} \\ \hline
      \multirow{3}*{\minitab[c]{GBM}} & No. of trees  & \textbf{100}, 200, 500 \\
        & Sub-sample             & \textbf{0.5}, 0.7, 0.8   \\
        & $\gamma$         & \textbf{0.1}, 0.25, 0.5    \\ \hline
    \end{tabular}
    \label{table:ensemble_hyperparams}
\end{table}
 }}
%
 {\small{
\begin{table}[!ht]
  \caption{Hyperparameters and tunable parameters for MLP emulator with the best parameters in bold.}
    \begin{tabular}{|cll|}\hline
      \textbf{Emulator} & \textbf{Hyperparameter and tunable parameter} & \textbf{Sought range} \\ \hline \hline 
      \multirow{6}*{\minitab[c]{MLP}} & No. of hidden layers  & 5, 25, 50, 100, \textbf{200} \\
        & Activation function     & \textbf{ReLU}, tanh, logistic  \\
        & $\alpha$                & $10^{-1}$, $10^{-2}$, $\boldsymbol{10^{-4}}$         \\
        & $\gamma$           & $10^{-1}$, $10^{-1}$,  $\boldsymbol{10^{-3}}$      \\
        & Solver                  & \textbf{Adam}, lbfgs, sgd  \\
        & Max. no. of iterations  & 1--5,000, \textbf{200}    \\ \hline
    \end{tabular}
    \label{table:ANN_hyperparams}
\end{table}
 }}

\section{Results}
\label{Sec:S4_Results}
\begin{figure}
  \centering
  \subfigure[ $\kappa_\mathrm{f}L = 2$ and $t = 0.1$]
    {\includegraphics[clip=true,width = 0.3\textwidth]
    {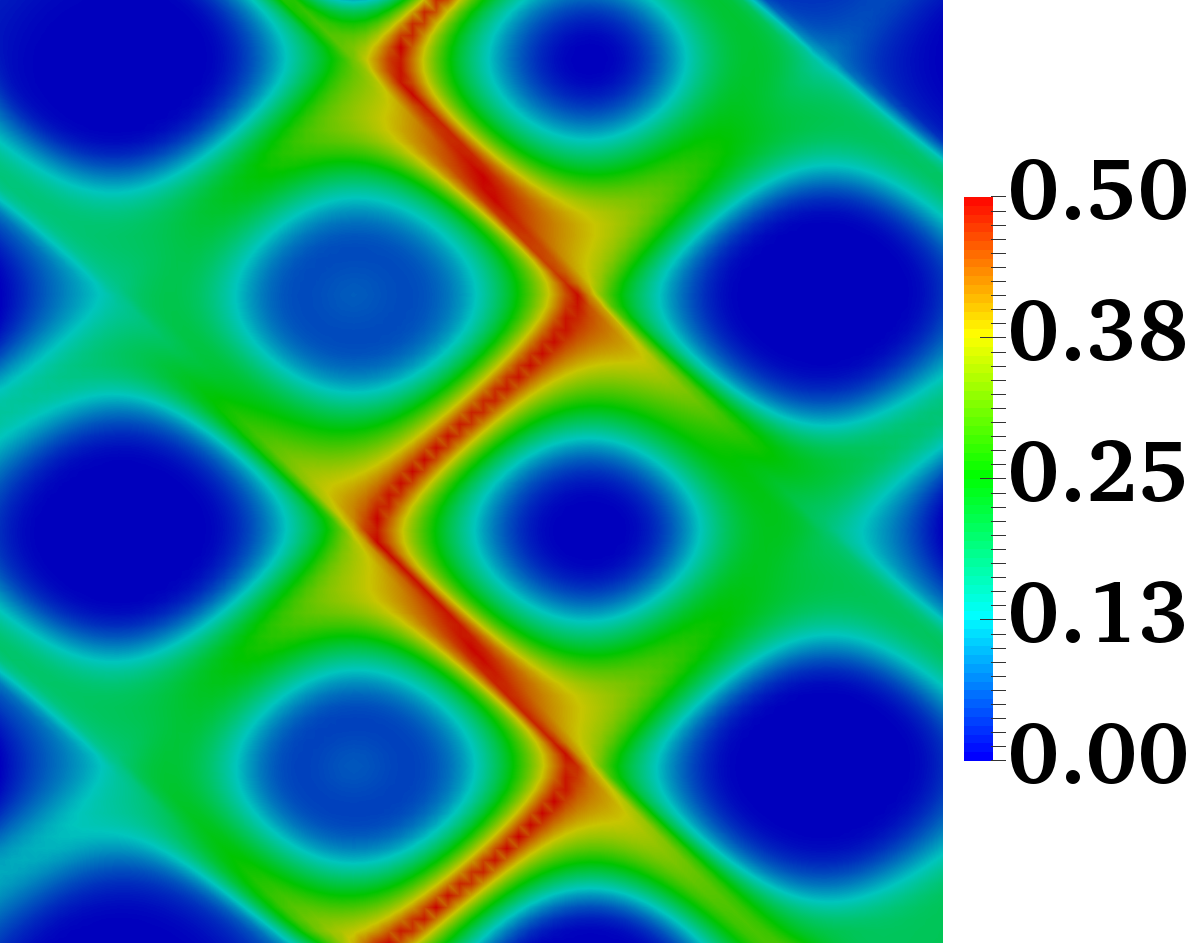}}
  \hspace{-0.5in}
  \subfigure[$\kappa_\mathrm{f}L = 2$ and $t = 0.5$]
    {\includegraphics[clip=true,width = 0.3\textwidth]
    {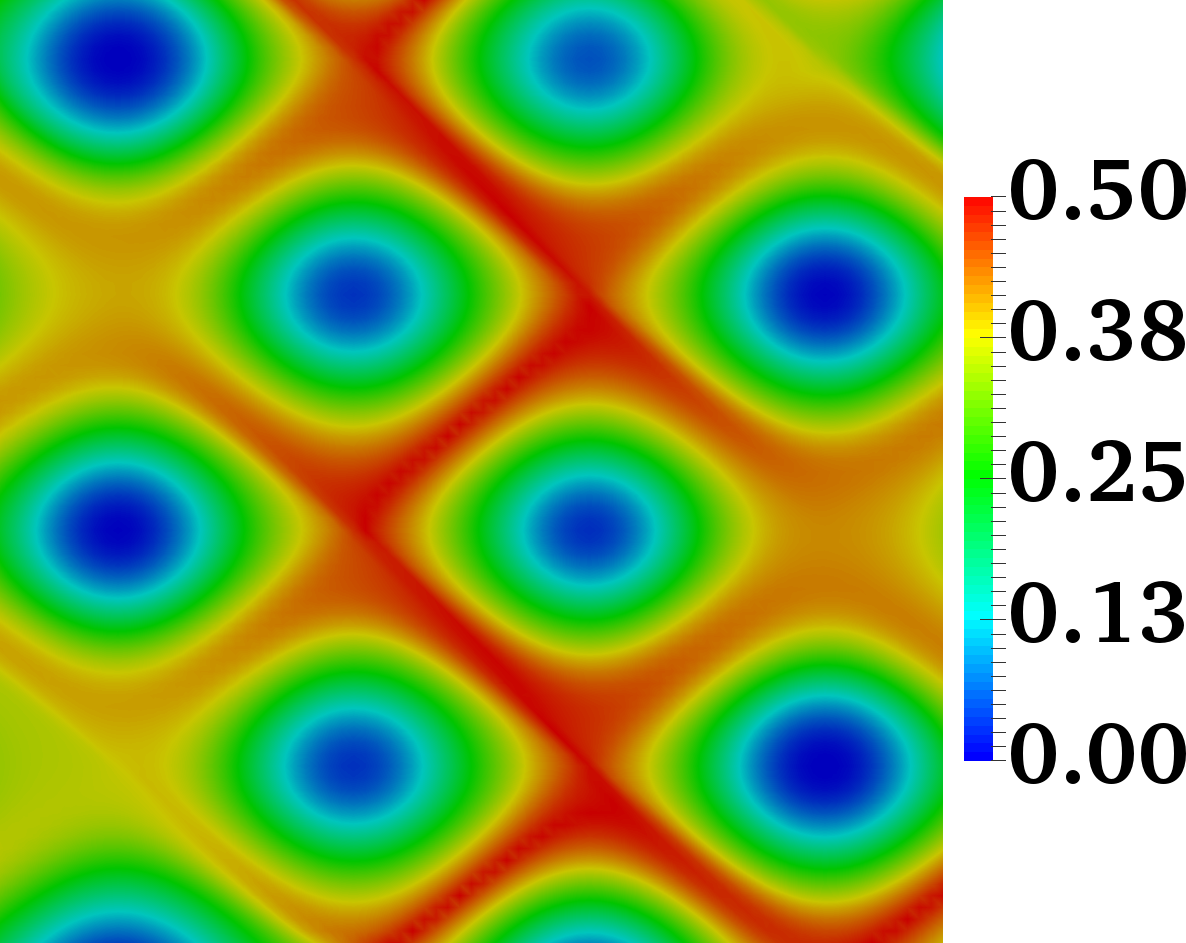}}
  \hspace{-0.5in}
  \subfigure[$\kappa_\mathrm{f}L = 2$ and $t = 1.0$]
    {\includegraphics[clip=true,width = 0.3\textwidth]
    {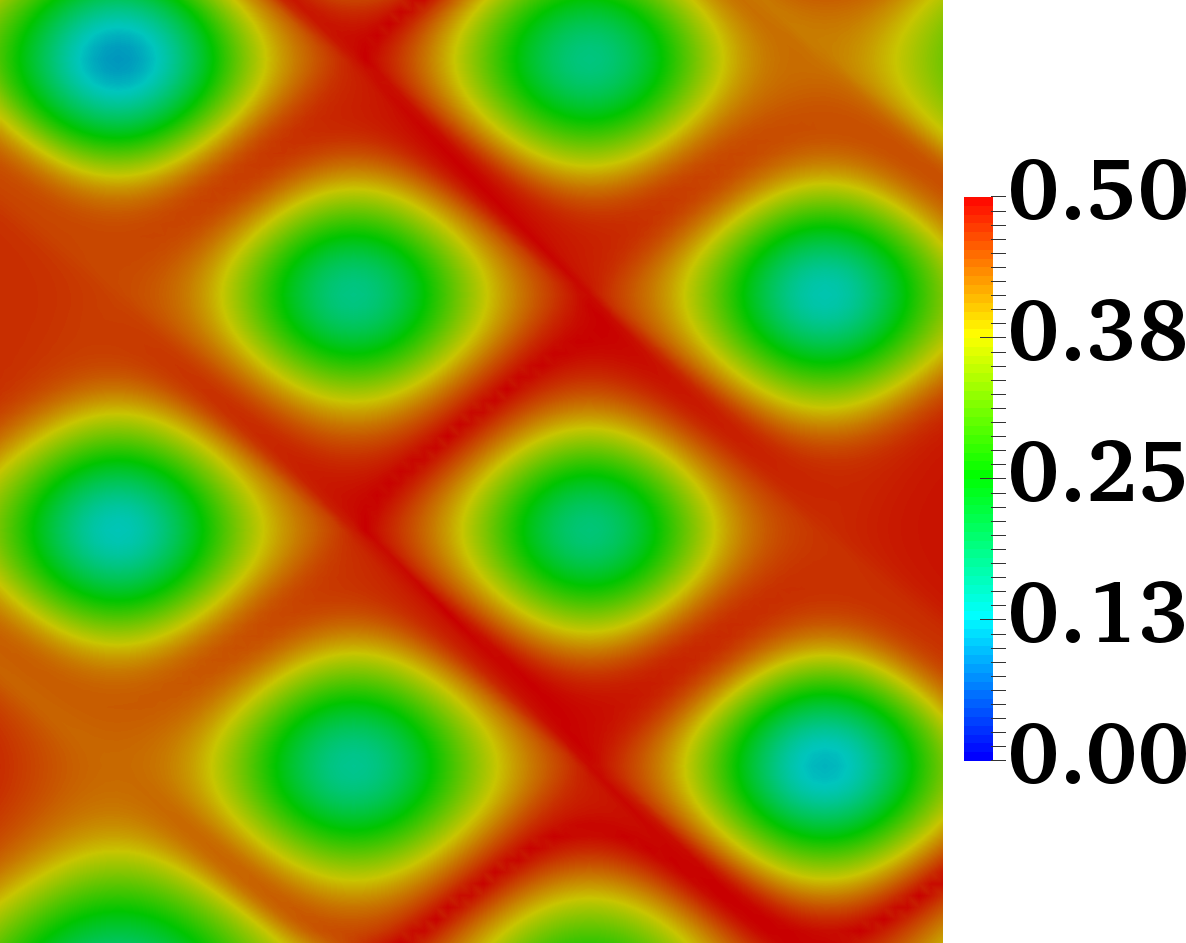}}
  \subfigure[$\kappa_\mathrm{f}L = 3$ and $t = 0.1$]
    {\includegraphics[clip=true,width = 0.3\textwidth]
    {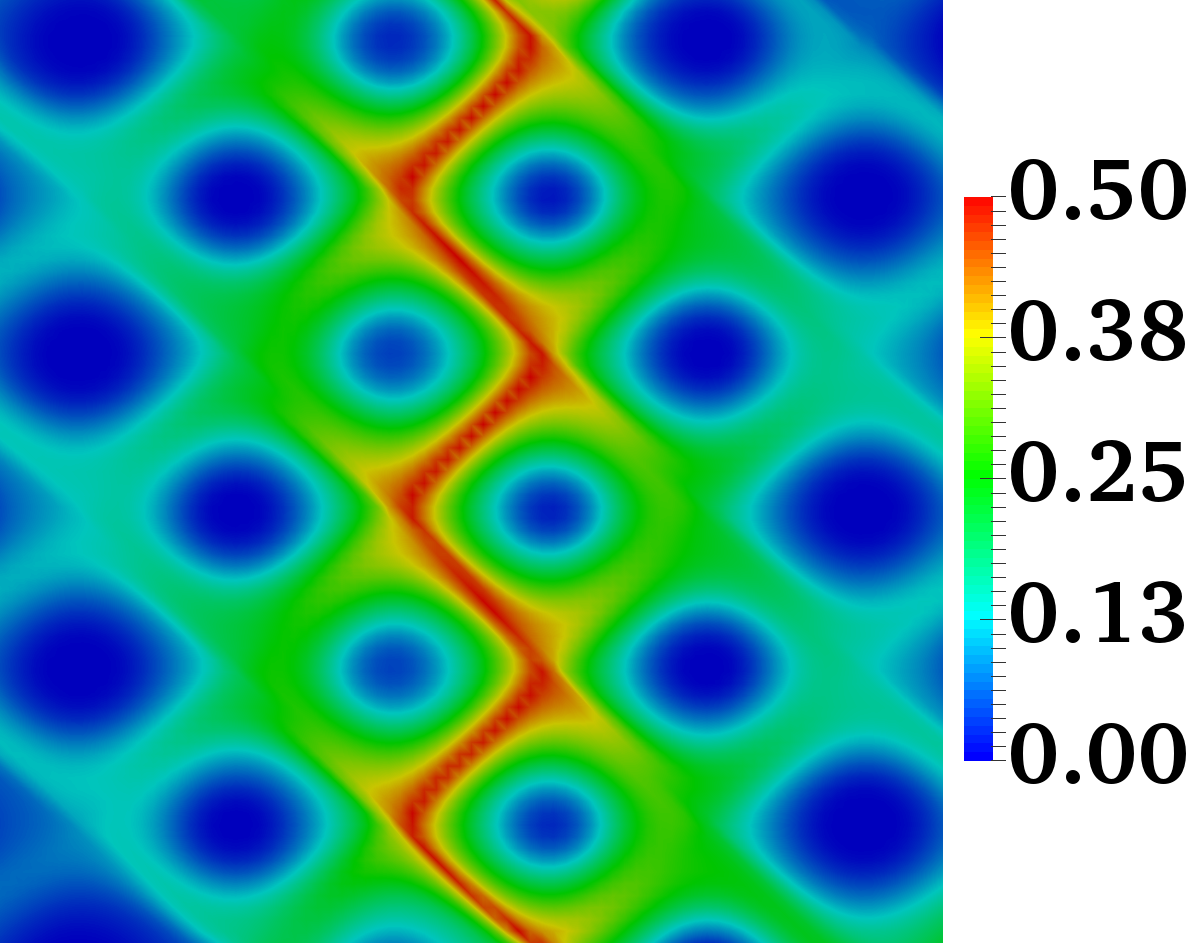}}
  \hspace{-0.5in}
  \subfigure[$\kappa_\mathrm{f}L = 3$ and $t = 0.5$]
    {\includegraphics[clip=true,width = 0.3\textwidth]
    {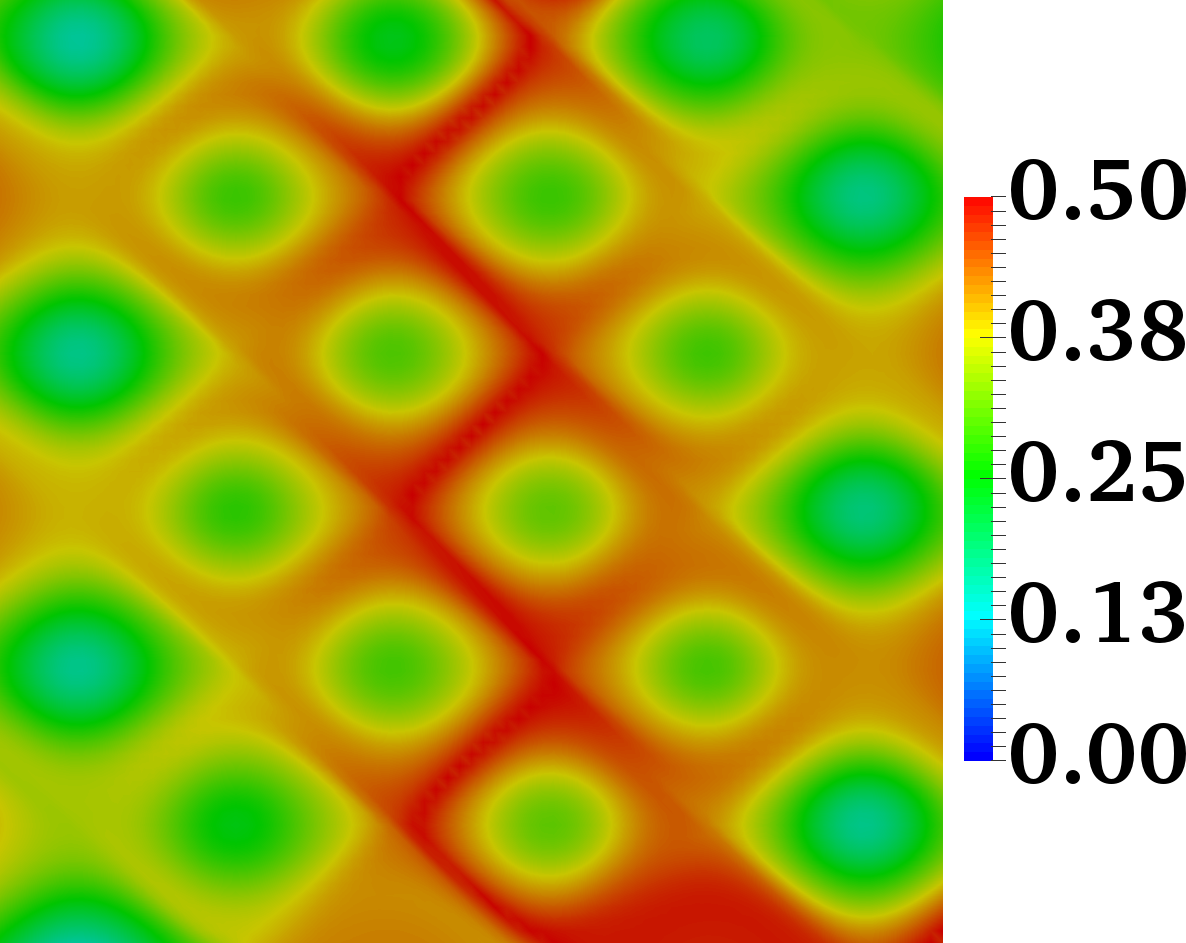}}
  \hspace{-0.5in}
  \subfigure[$\kappa_\mathrm{f}L = 3$ and $t = 1.0$]
    {\includegraphics[clip=true,width = 0.3\textwidth]
    {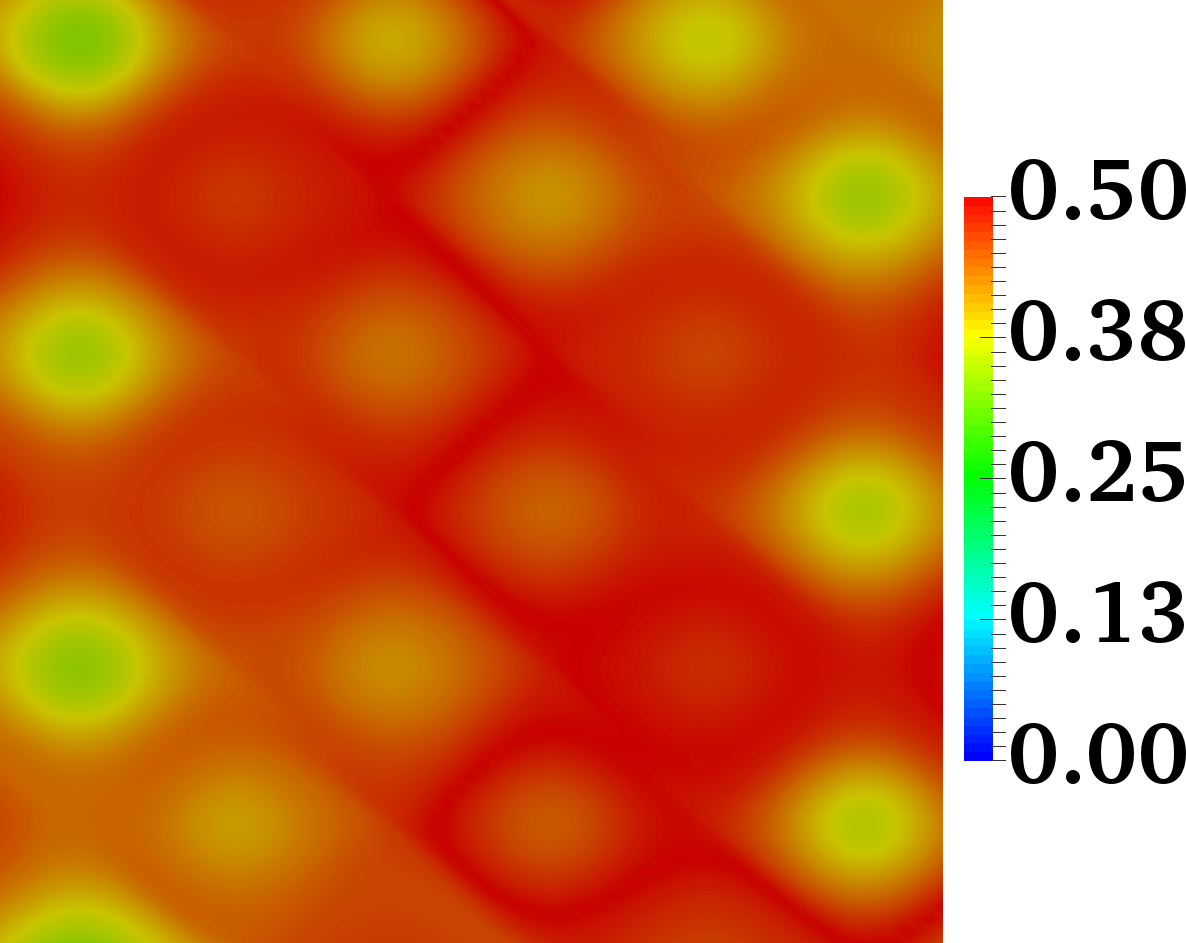}}
  \subfigure[$\kappa_\mathrm{f}L = 4$ and $t = 0.1$]
    {\includegraphics[clip=true,width = 0.3\textwidth]
    {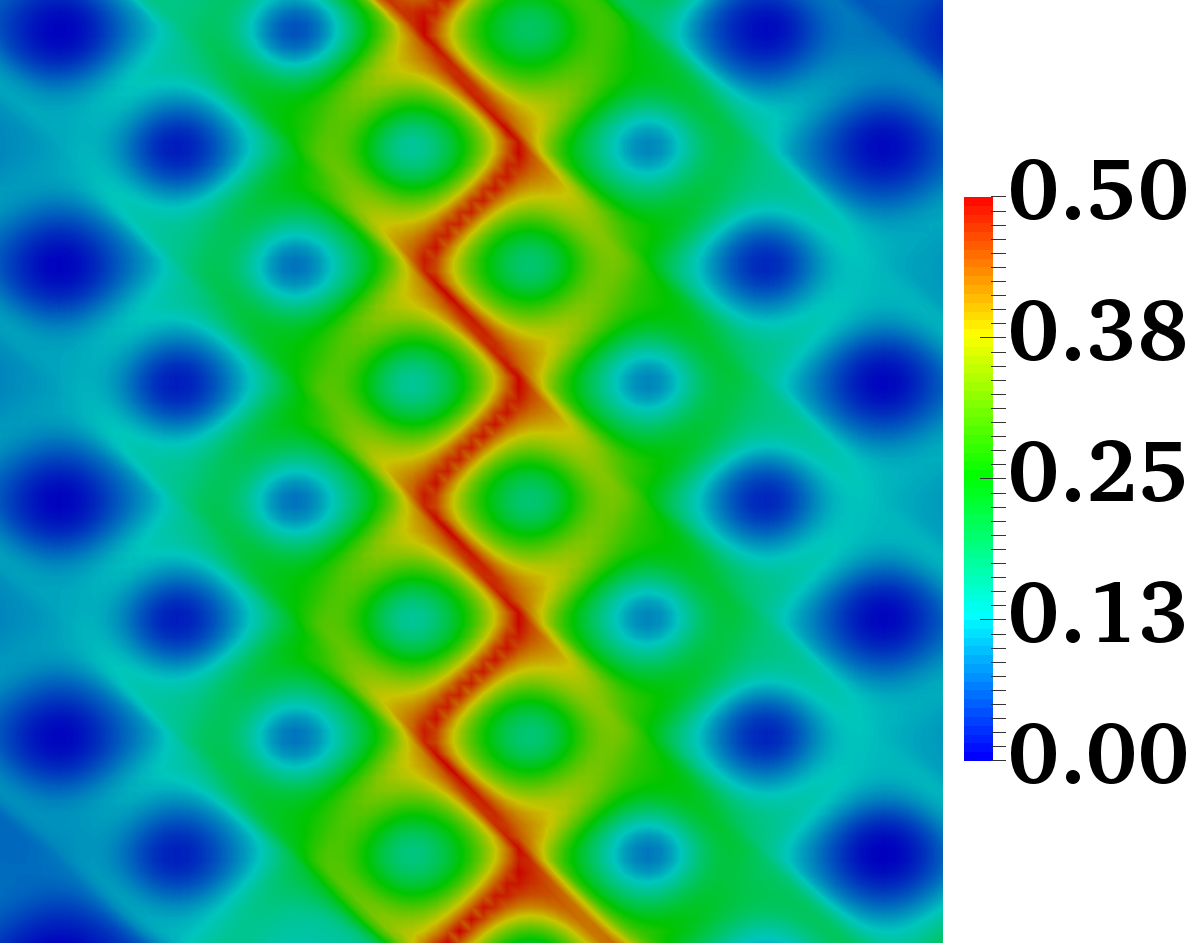}}
  \hspace{-0.5in}
  \subfigure[$\kappa_\mathrm{f}L = 4$ and $t = 0.5$]
    {\includegraphics[clip=true,width = 0.3\textwidth]
    {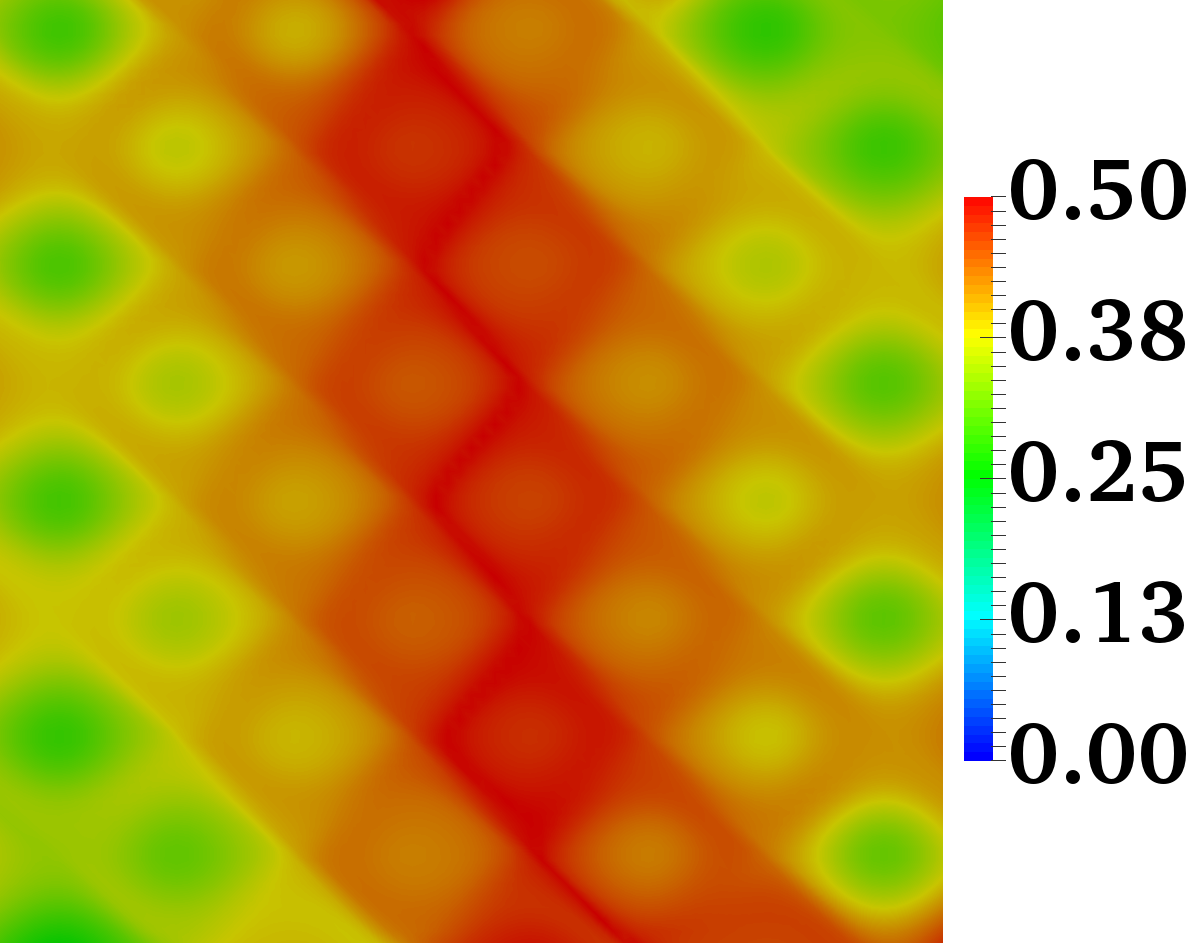}}
  \hspace{-0.5in}
  \subfigure[$\kappa_\mathrm{f}L = 4$ and $t = 1.0$]
    {\includegraphics[clip=true,width = 0.3\textwidth]
    {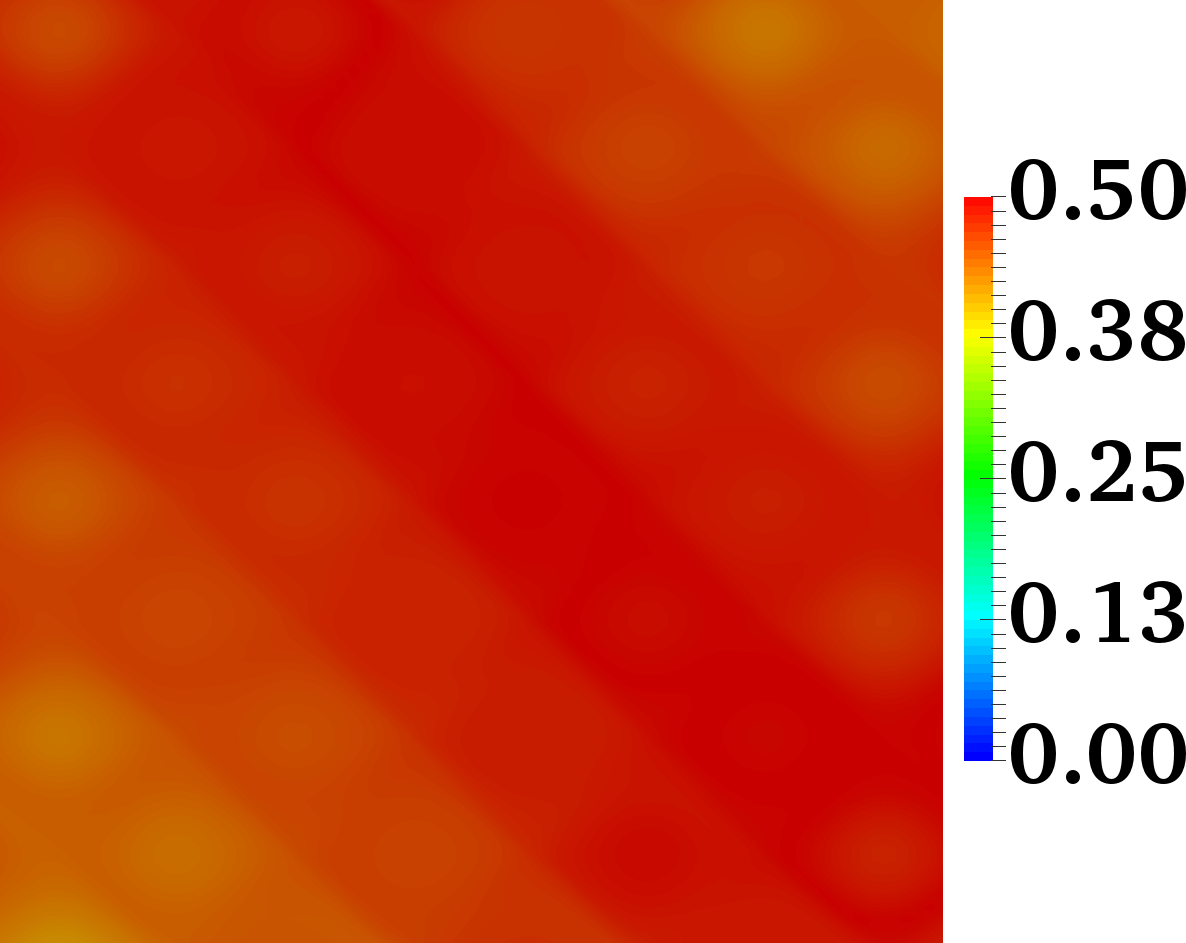}}
  \subfigure[$\kappa_\mathrm{f}L = 5$ and $t = 0.1$]
    {\includegraphics[clip=true,width = 0.3\textwidth]
    {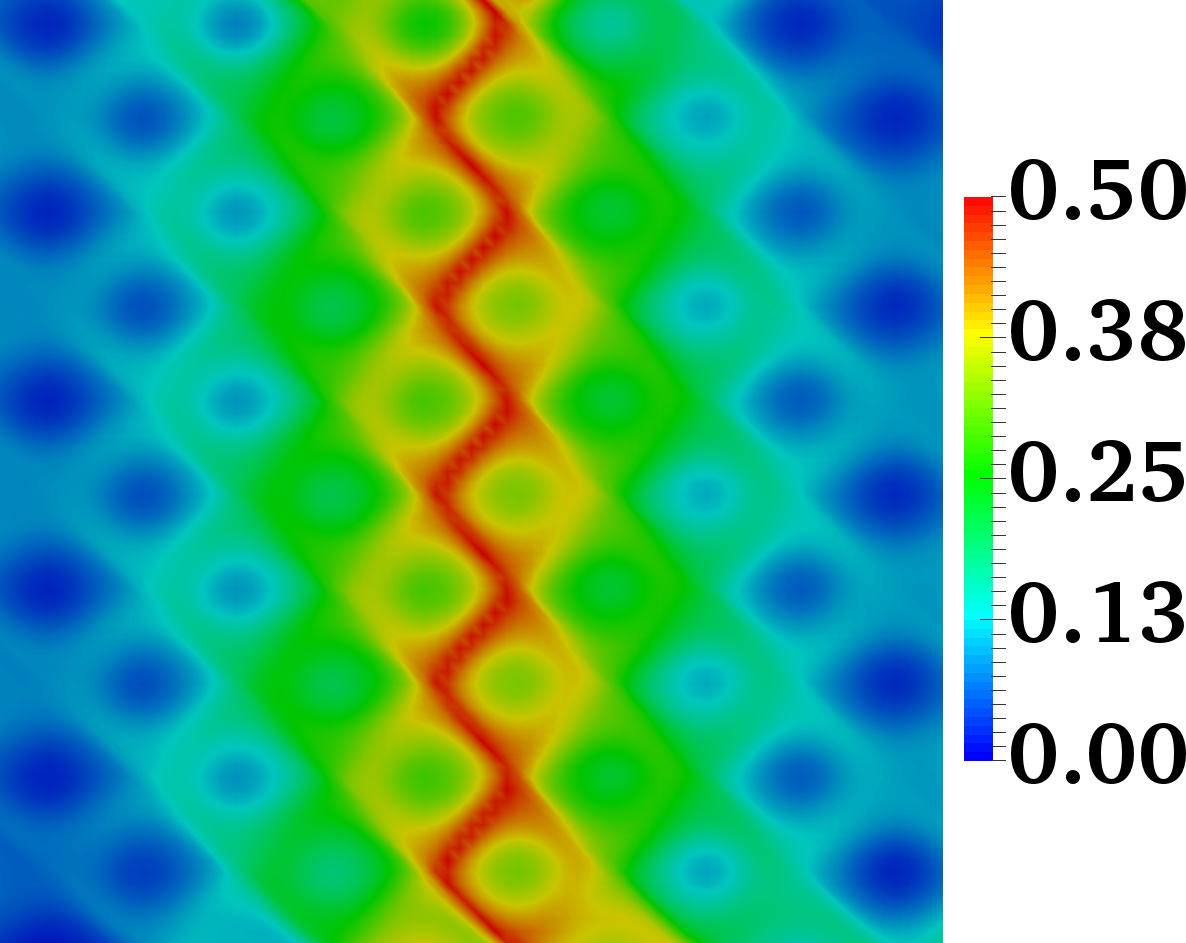}}
  \hspace{-0.5in}
  \subfigure[$\kappa_\mathrm{f}L = 5$ and $t = 0.5$]
    {\includegraphics[clip=true,width = 0.3\textwidth]
    {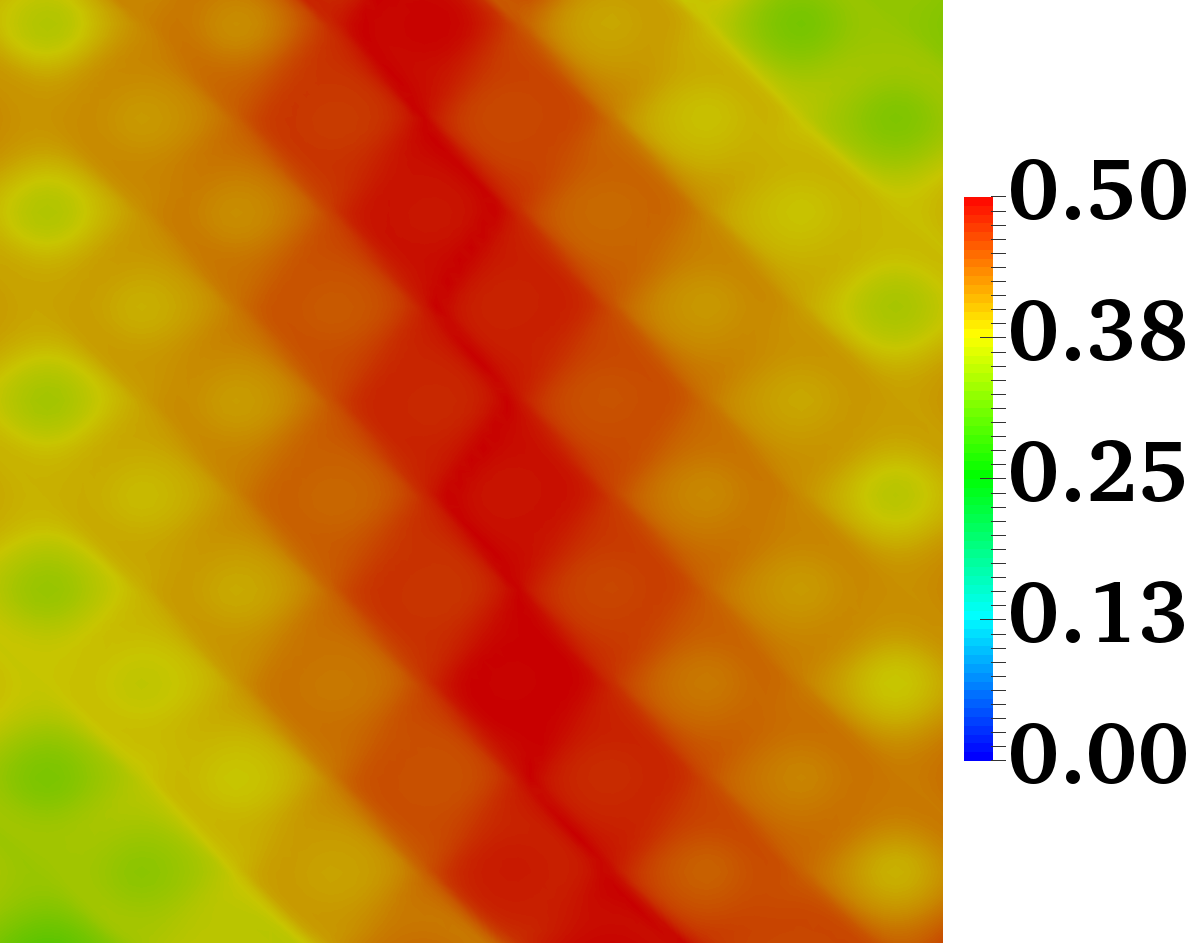}}
  \hspace{-0.5in}
  \subfigure[$\kappa_\mathrm{f}L = 5$ and $t = 1.0$]
    {\includegraphics[clip=true,width = 0.3\textwidth]
    {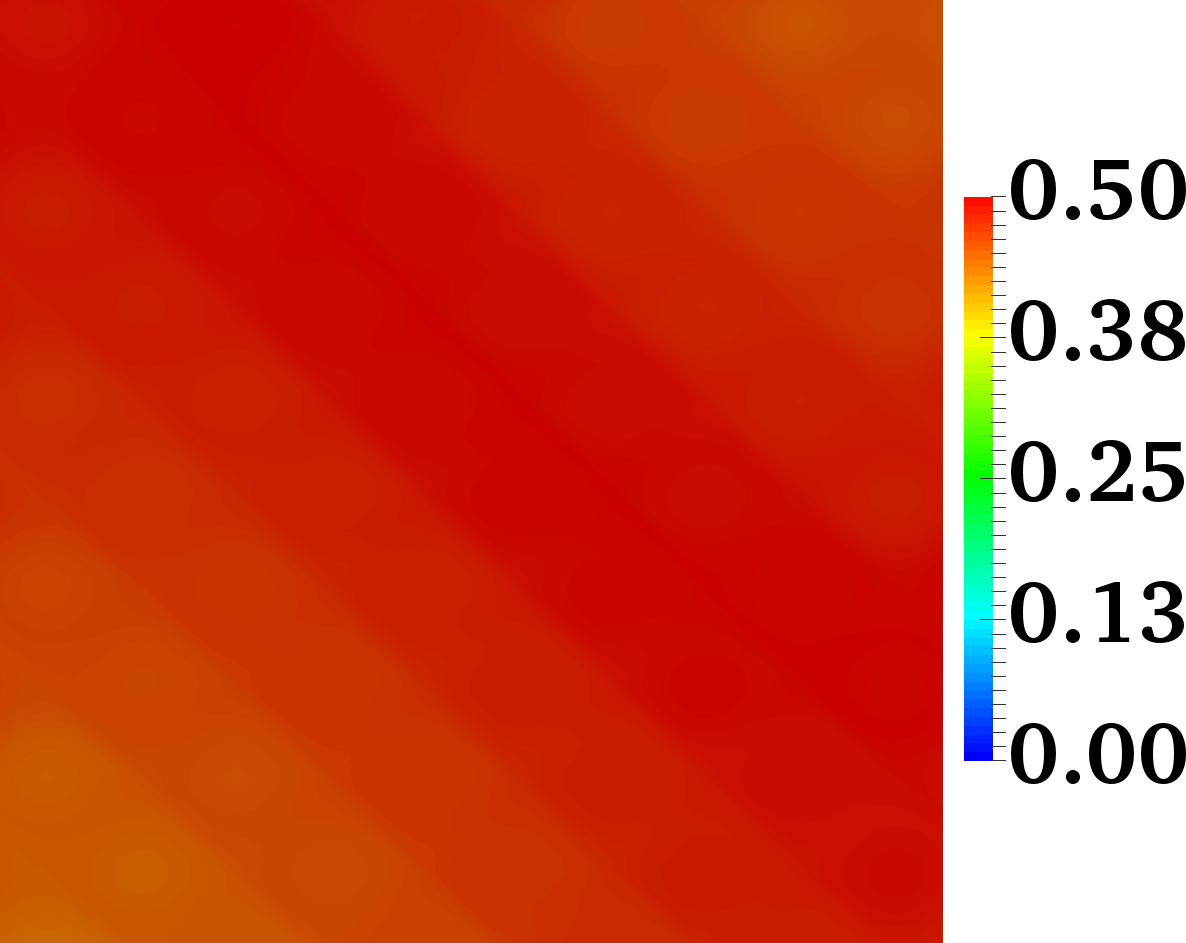}}
  \caption{\textrm{\textbf{Concentration contours of species $C$ under high anisotropy:}}~Concentration of product $C$ at times $t = 0.1, \, 0.5,$ and $1.0$. 
  Other input parameters were $\frac{\alpha_\mathrm{L}}{\alpha_\mathrm{T}} = 10^{3}$ (high anisotropy), $v_0 = 1$, $T = 0.1$, and $D_\mathrm{m} = 10^{-3}$. 
  Increased $\kappa_\mathrm{f}L$ increases $C$ production, especially at later times.
  \label{Fig:Contours_C_Difftimes_1}}
\end{figure}

\begin{figure}
  \centering
  \subfigure[$\kappa_\mathrm{f}L = 2$ and $t = 0.1$]
    {\includegraphics[clip=true,width = 0.3\textwidth]
    {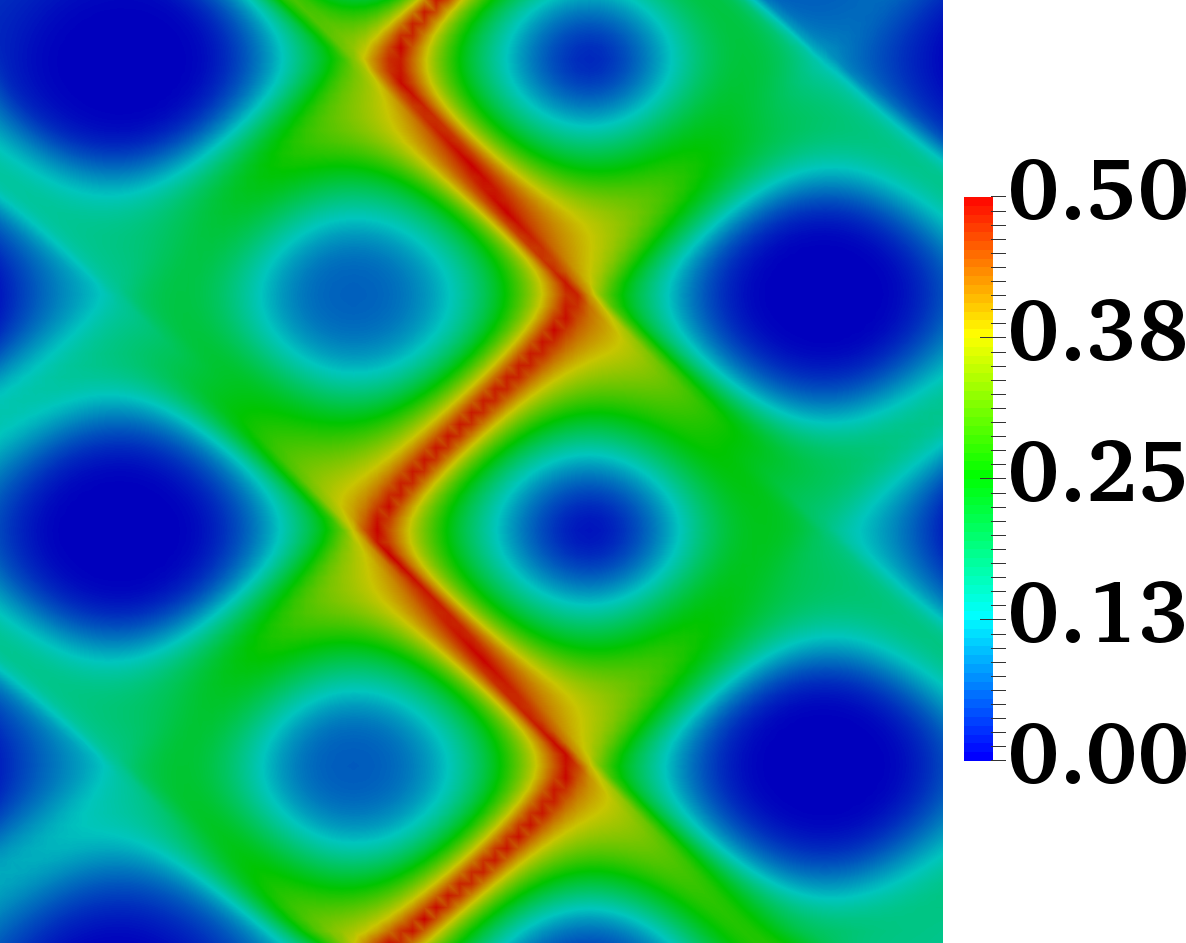}}
  \hspace{-0.5in}
  \subfigure[$\kappa_\mathrm{f}L = 2$ and $t = 0.5$]
    {\includegraphics[clip=true,width = 0.3\textwidth]
    {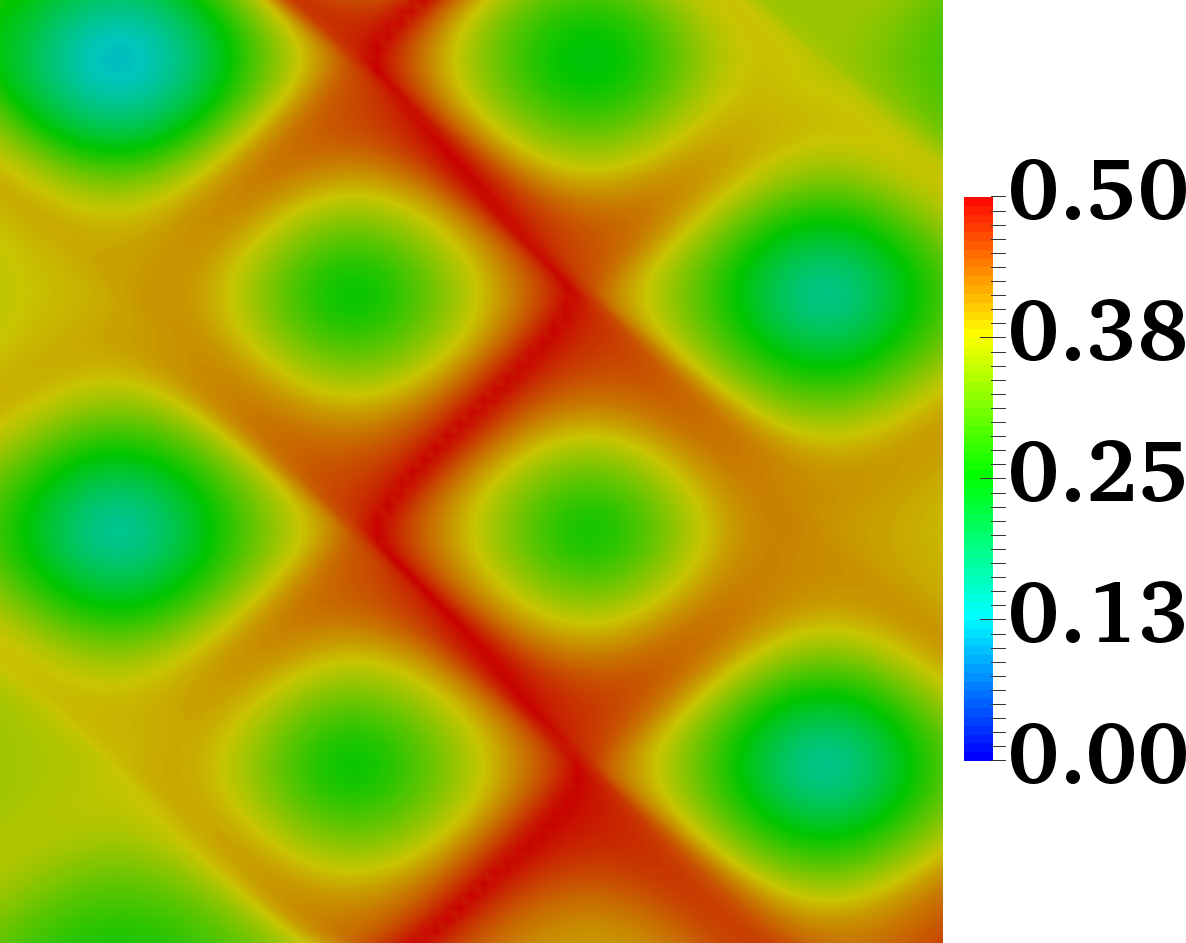}}
  \hspace{-0.5in}
  \subfigure[$\kappa_\mathrm{f}L = 2$ and $t = 1.0$]
    {\includegraphics[clip=true,width = 0.3\textwidth]
    {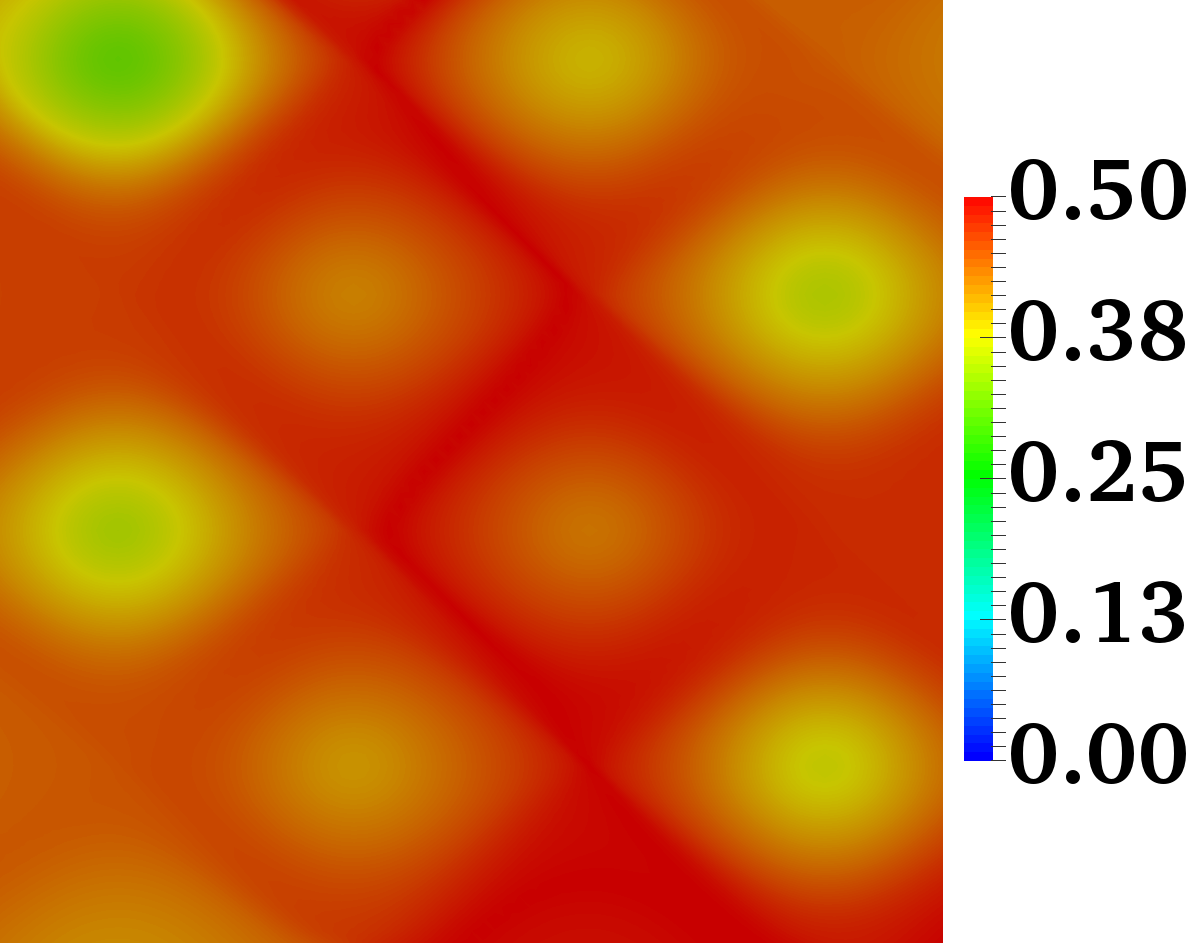}}
  \subfigure[$\kappa_\mathrm{f}L = 3$ and $t = 0.1$]
    {\includegraphics[clip=true,width = 0.3\textwidth]
    {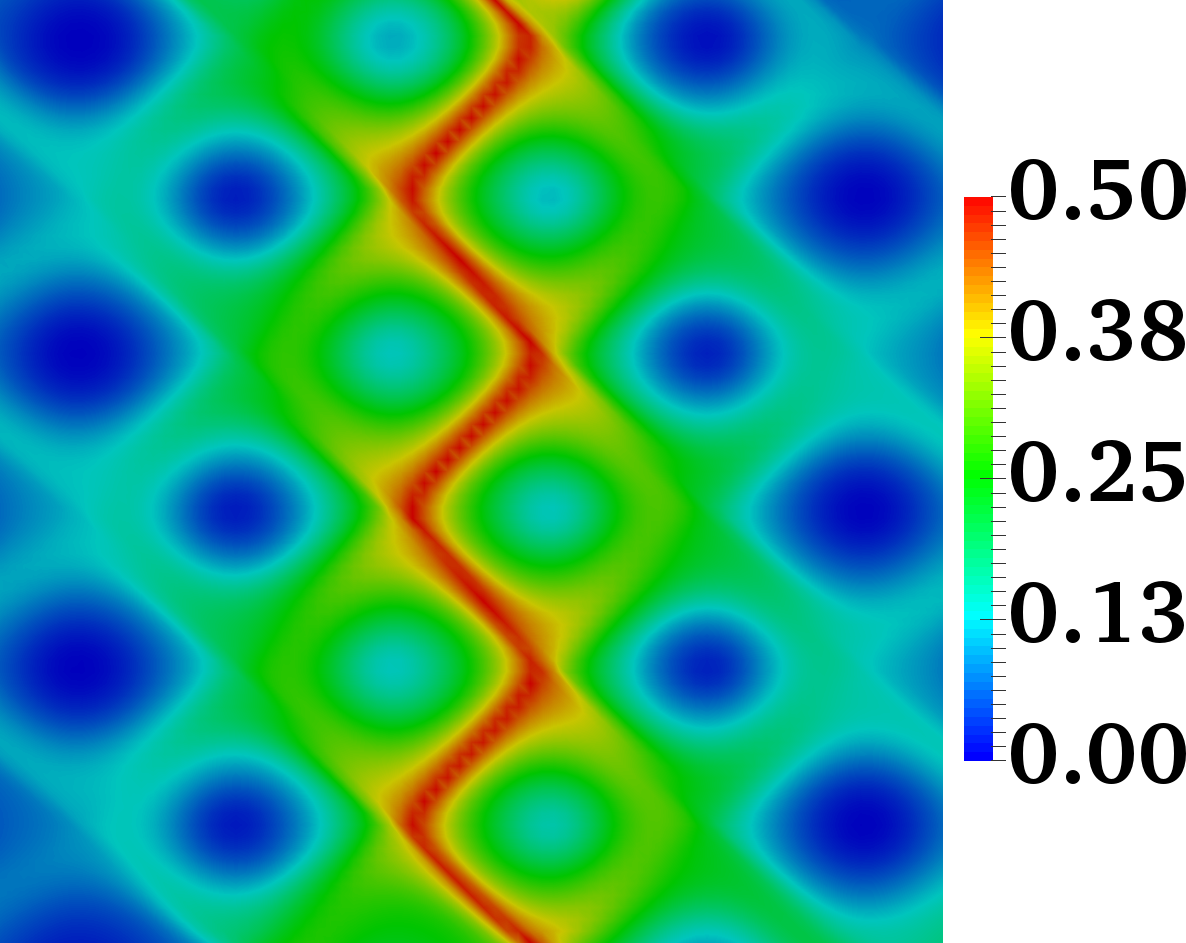}}
  \hspace{-0.5in}
  \subfigure[$\kappa_\mathrm{f}L = 3$ and $t = 0.5$]
    {\includegraphics[clip=true,width = 0.3\textwidth]
    {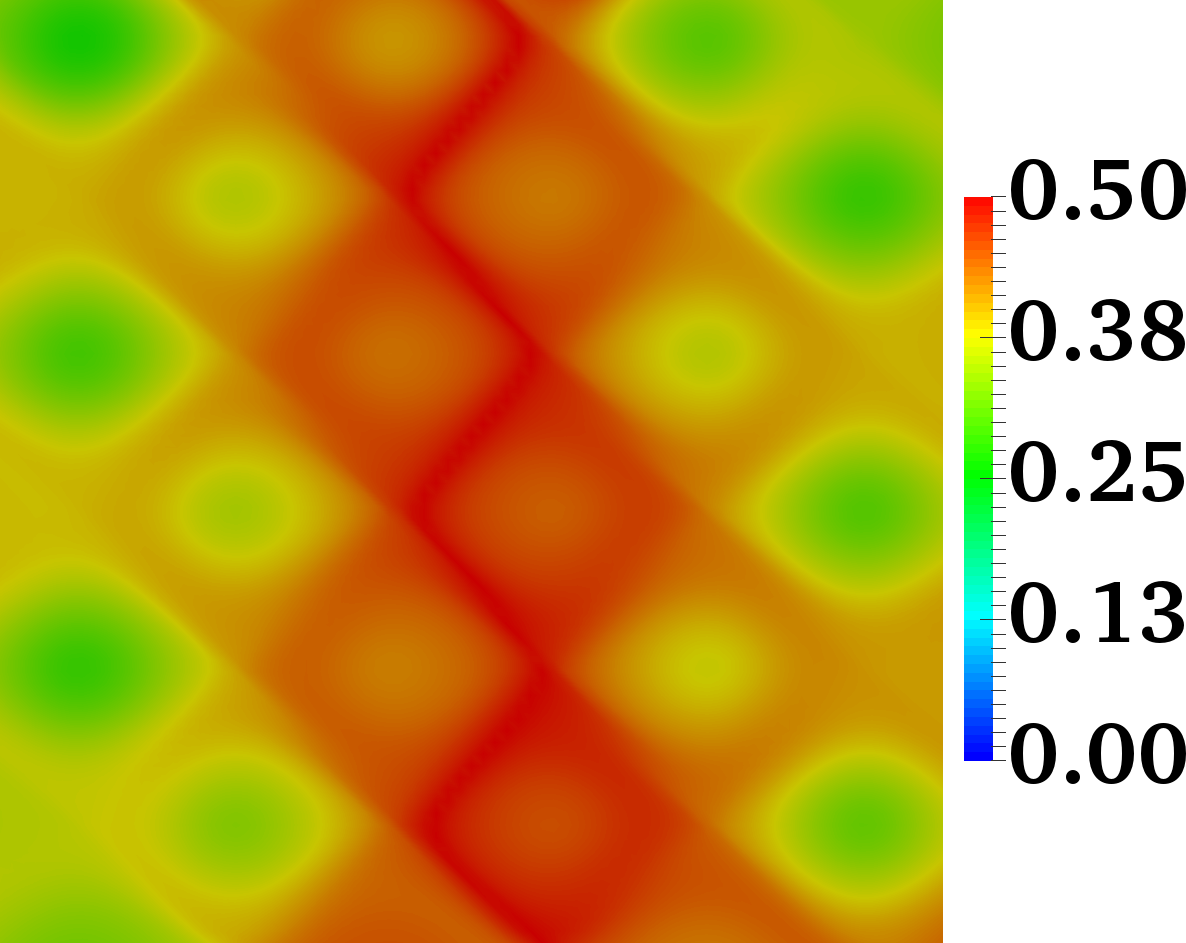}}
  \hspace{-0.5in}
  \subfigure[$\kappa_\mathrm{f}L = 3$ and $t = 1.0$]
    {\includegraphics[clip=true,width = 0.3\textwidth]
    {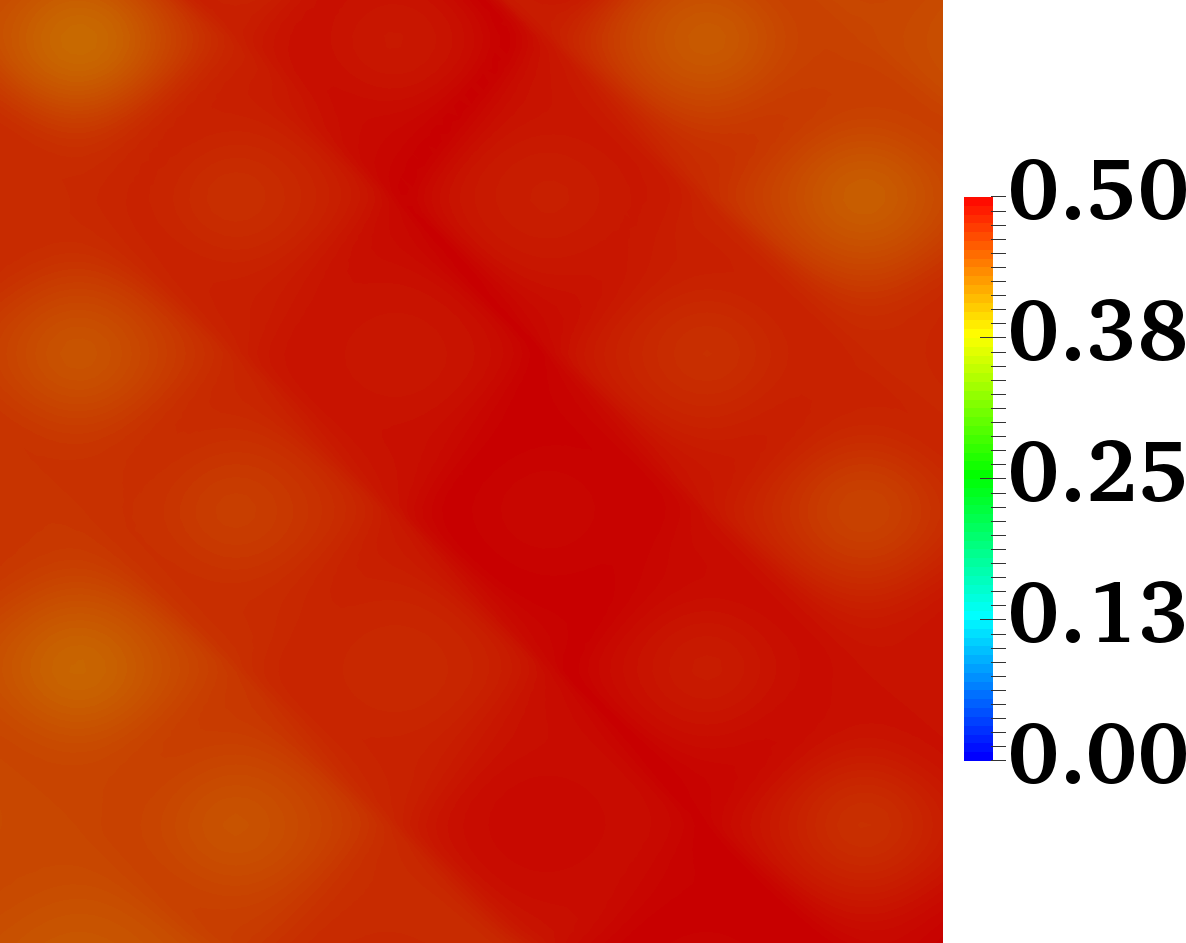}}
  \subfigure[$\kappa_\mathrm{f}L = 4$ and $t = 0.1$]
    {\includegraphics[clip=true,width = 0.3\textwidth]
    {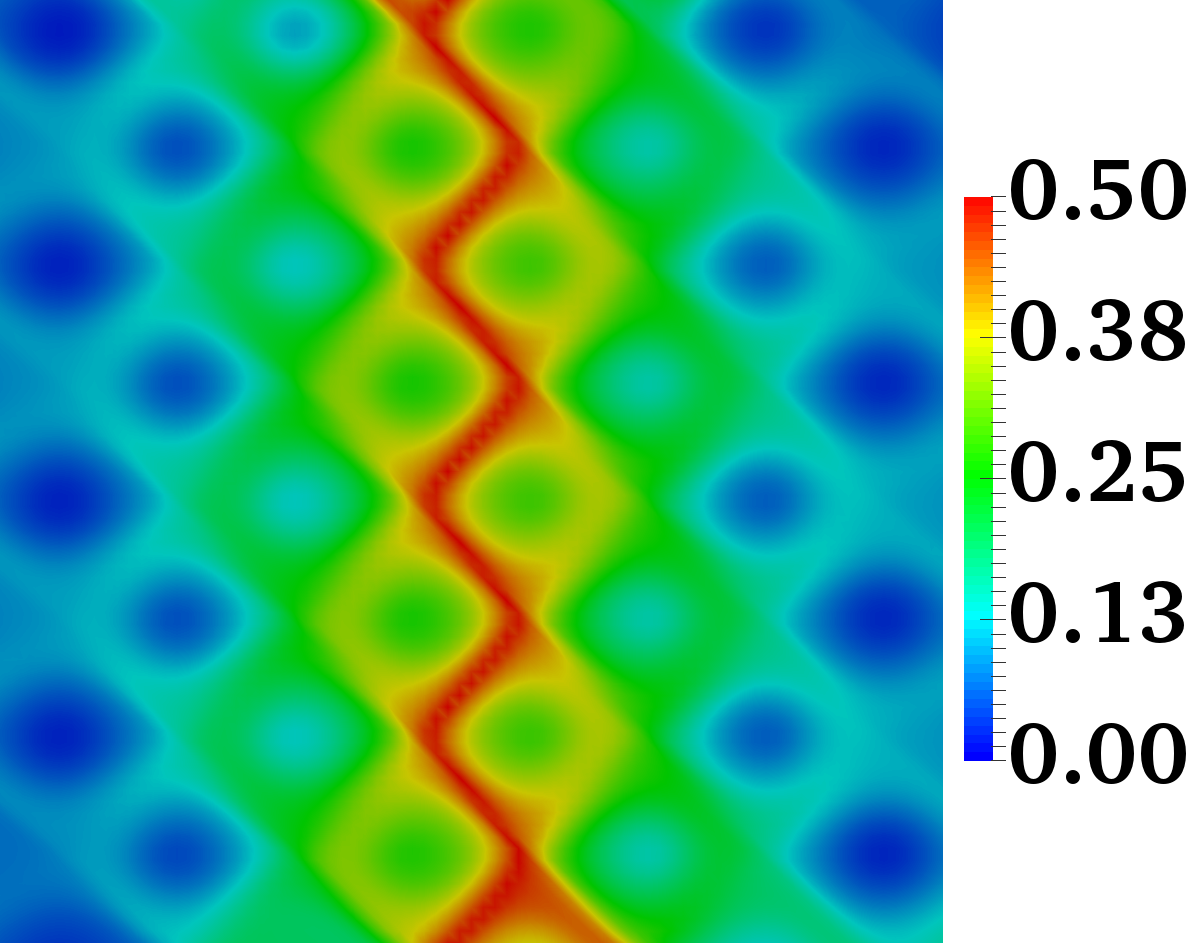}}
  \hspace{-0.5in}
  \subfigure[$\kappa_\mathrm{f}L = 4$ and $t = 0.5$]
    {\includegraphics[clip=true,width = 0.3\textwidth]
    {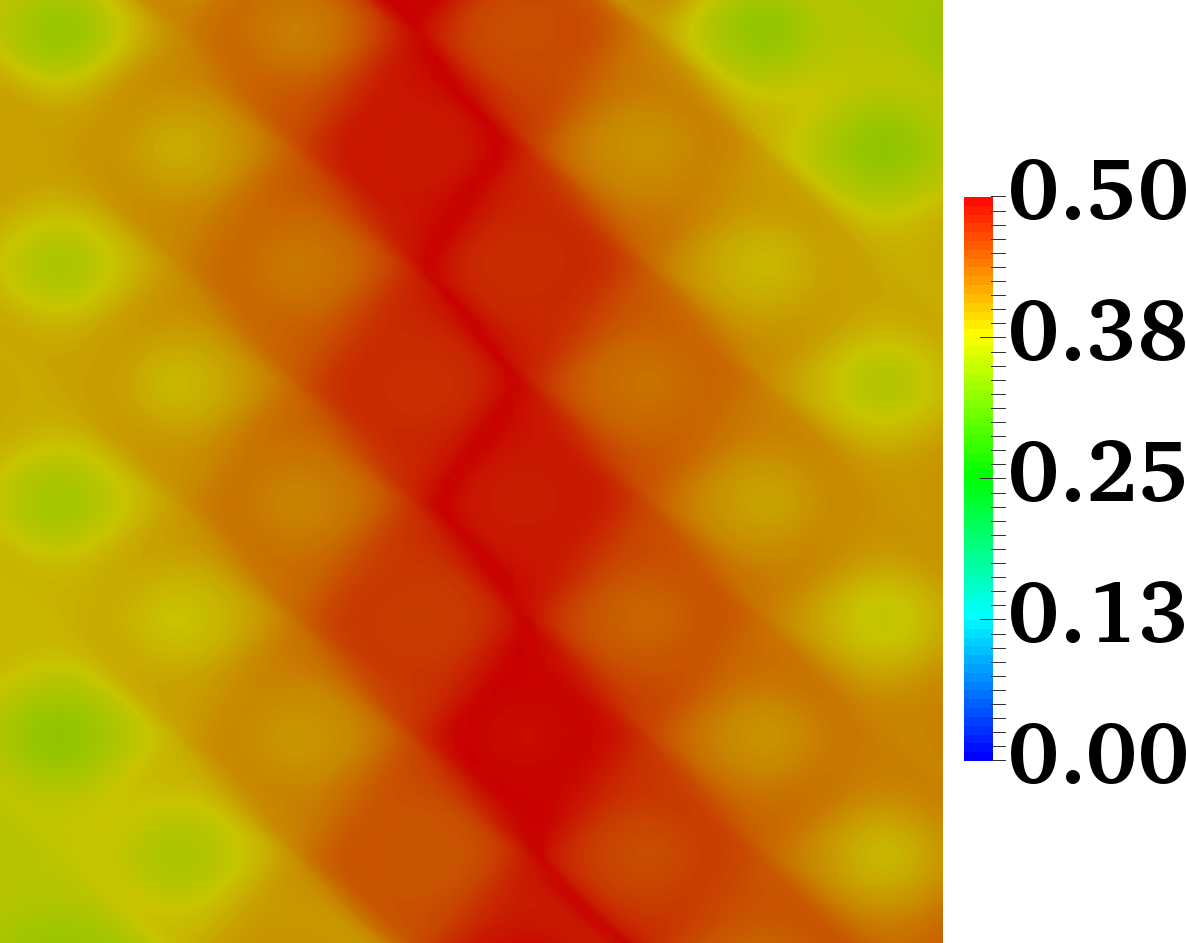}}
  \hspace{-0.5in}
  \subfigure[$\kappa_\mathrm{f}L = 4$ and $t = 1.0$]
    {\includegraphics[clip=true,width = 0.3\textwidth]
    {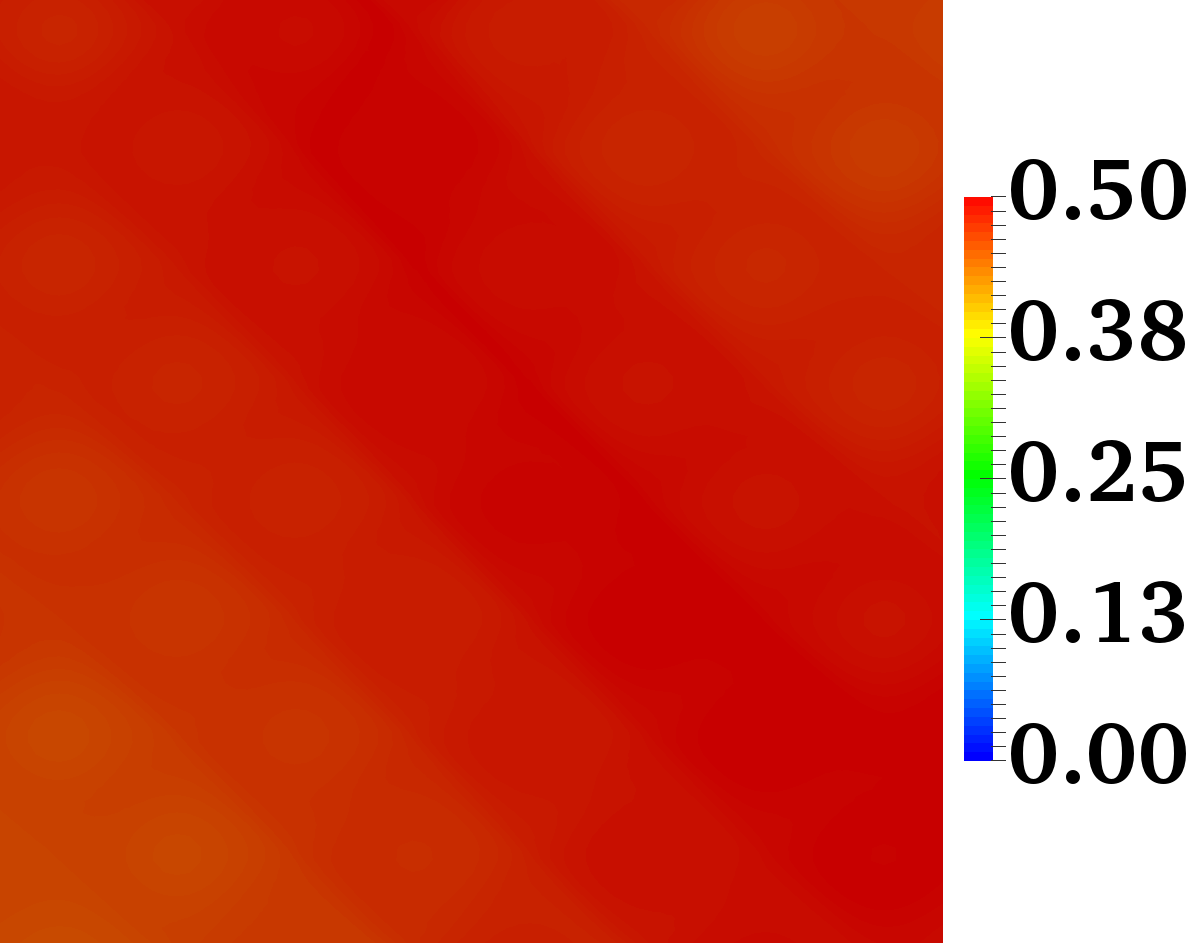}}
  \subfigure[$\kappa_\mathrm{f}L = 5$ and $t = 0.1$]
    {\includegraphics[clip=true,width = 0.3\textwidth]
    {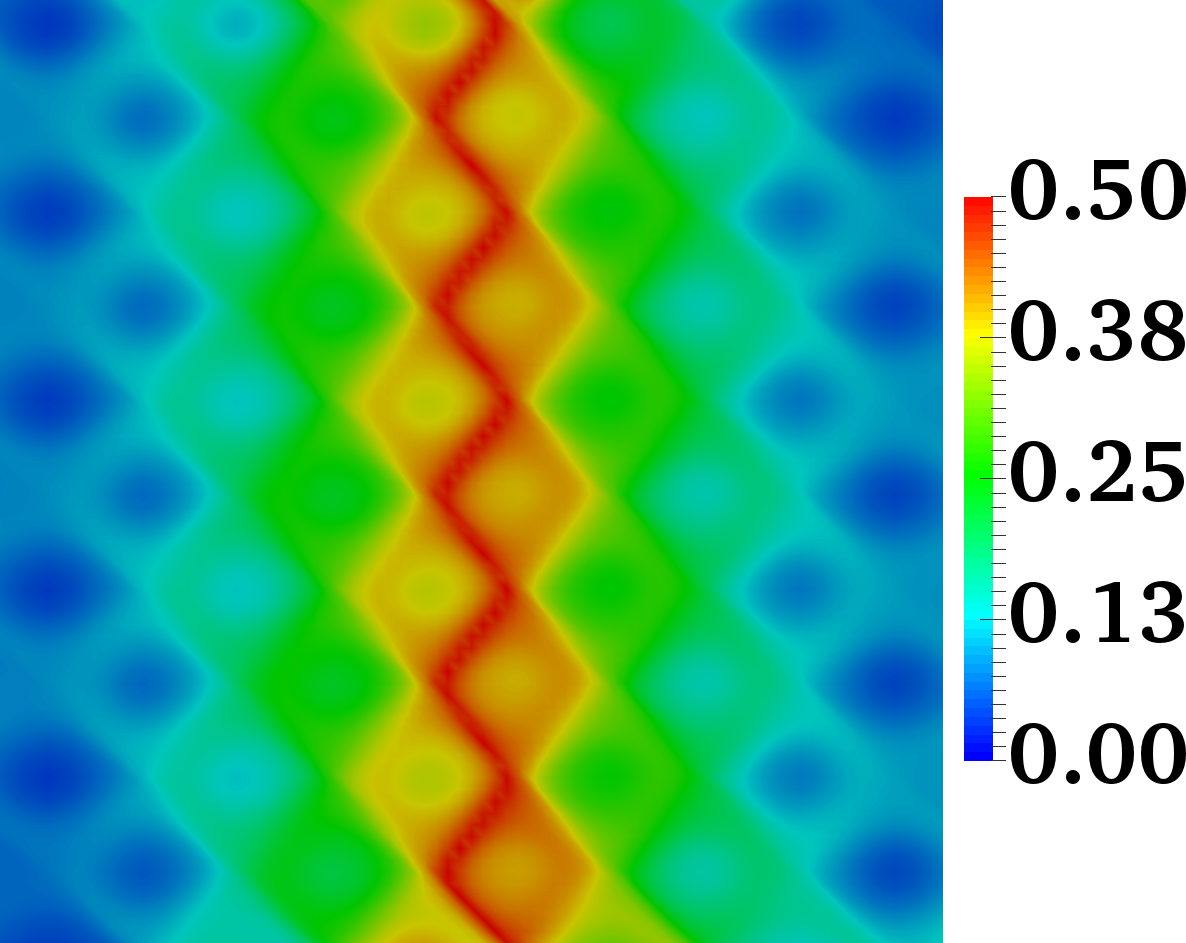}}
  \hspace{-0.5in}
  \subfigure[$\kappa_\mathrm{f}L = 5$ and $t = 0.5$]
    {\includegraphics[clip=true,width = 0.3\textwidth]
    {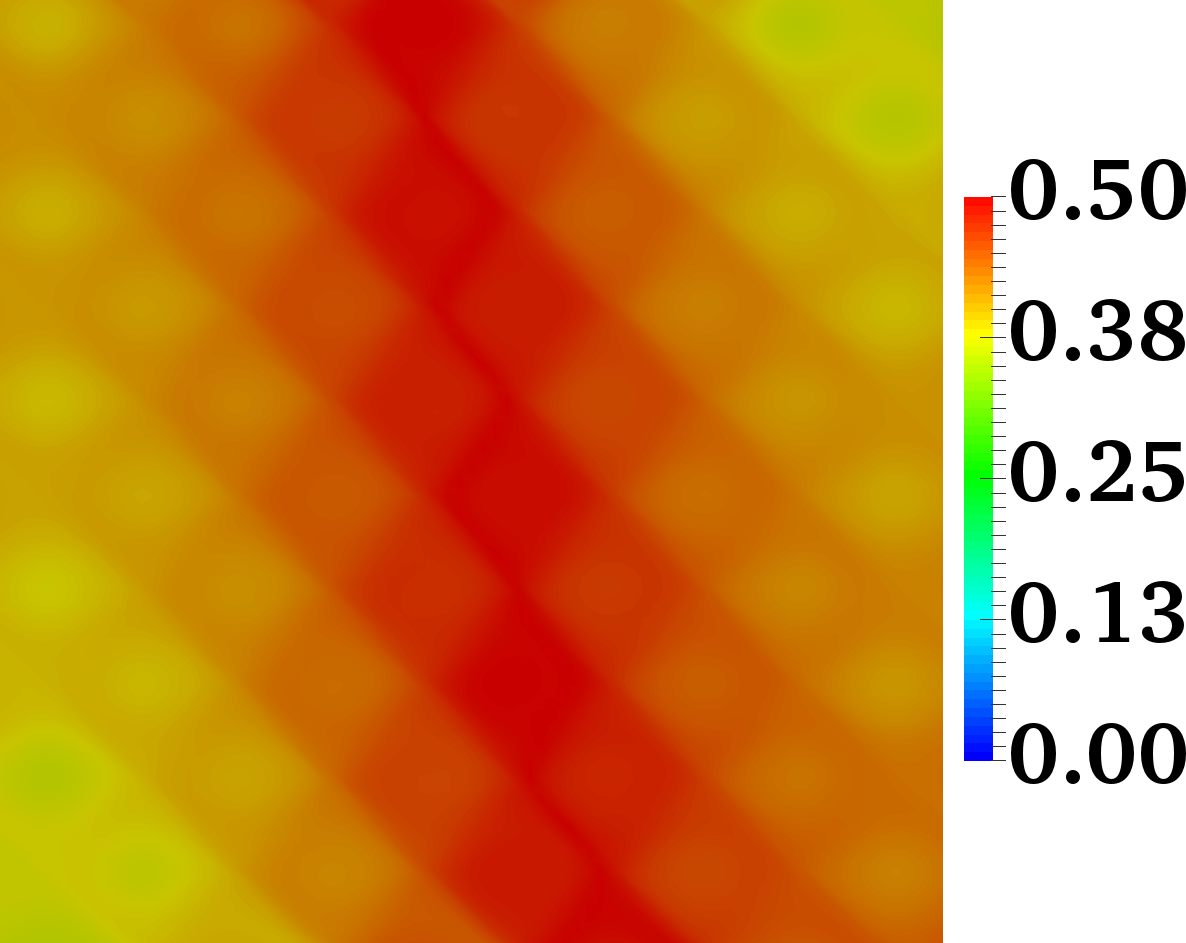}}
  \hspace{-0.5in}
  \subfigure[$\kappa_\mathrm{f}L = 5$ and $t = 1.0$]
    {\includegraphics[clip=true,width = 0.3\textwidth]
    {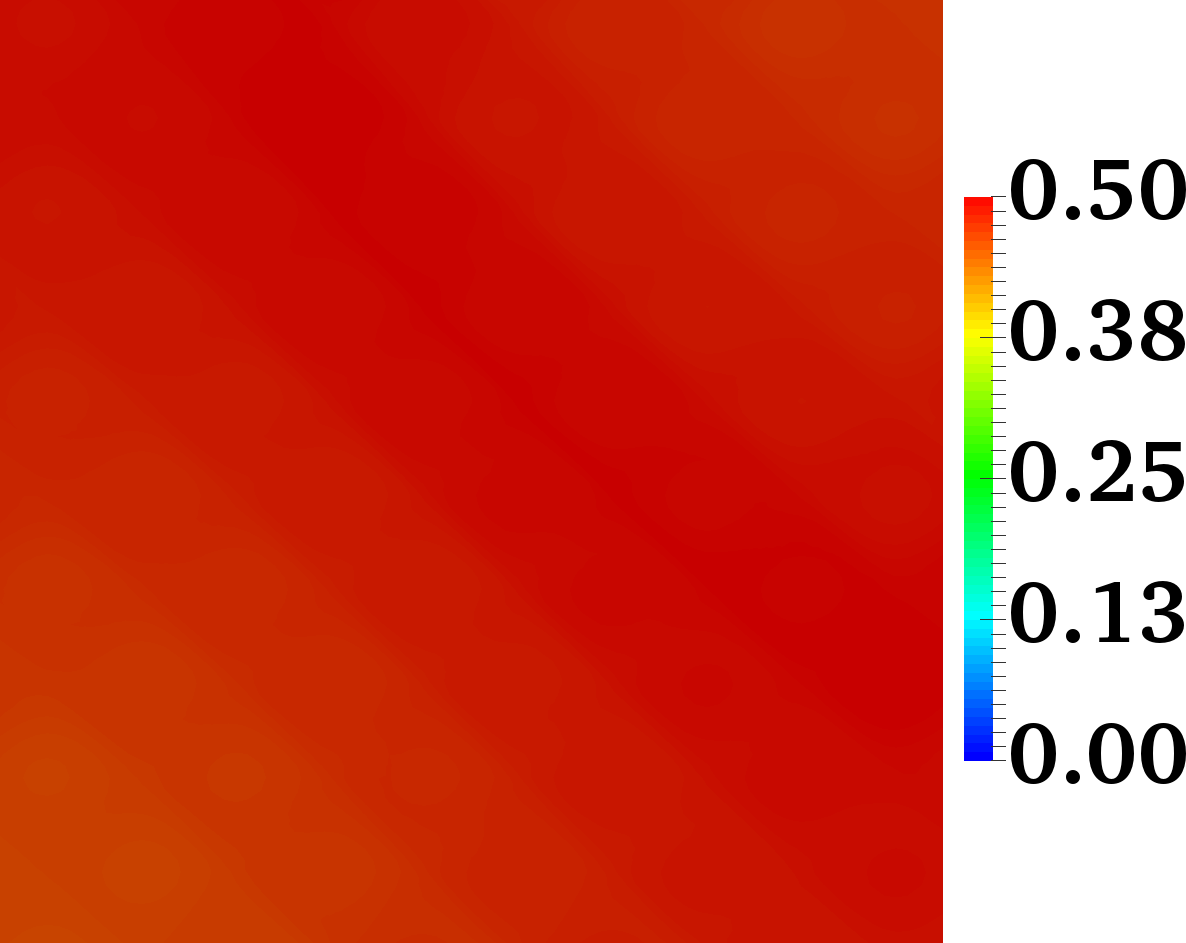}}
  \caption{\textrm{\textbf{Concentration contours of species $C$ under medium anisotropy:}}~Concentration of product $C$ at times $t = 0.1, \, 0.5,$ 
    and $1.0$. 
    Other input parameters were $\frac{\alpha_\mathrm{L}}{\alpha_\mathrm{T}} = 100$ (medium anisotropy), $v_0 = 1$, $T = 0.1$, and $D_\mathrm{m} = 10^{-3}$. 
    Lower anisotropy increased $C$ production than higher anisotropy in Fig.~\ref{Fig:Contours_C_Difftimes_1}.
  \label{Fig:Contours_C_Difftimes_2}}
\end{figure}

\begin{figure}
  \centering
  \subfigure[$\kappa_\mathrm{f}L = 2$ and $t = 0.1$]
    {\includegraphics[clip=true,width = 0.3\textwidth]
    {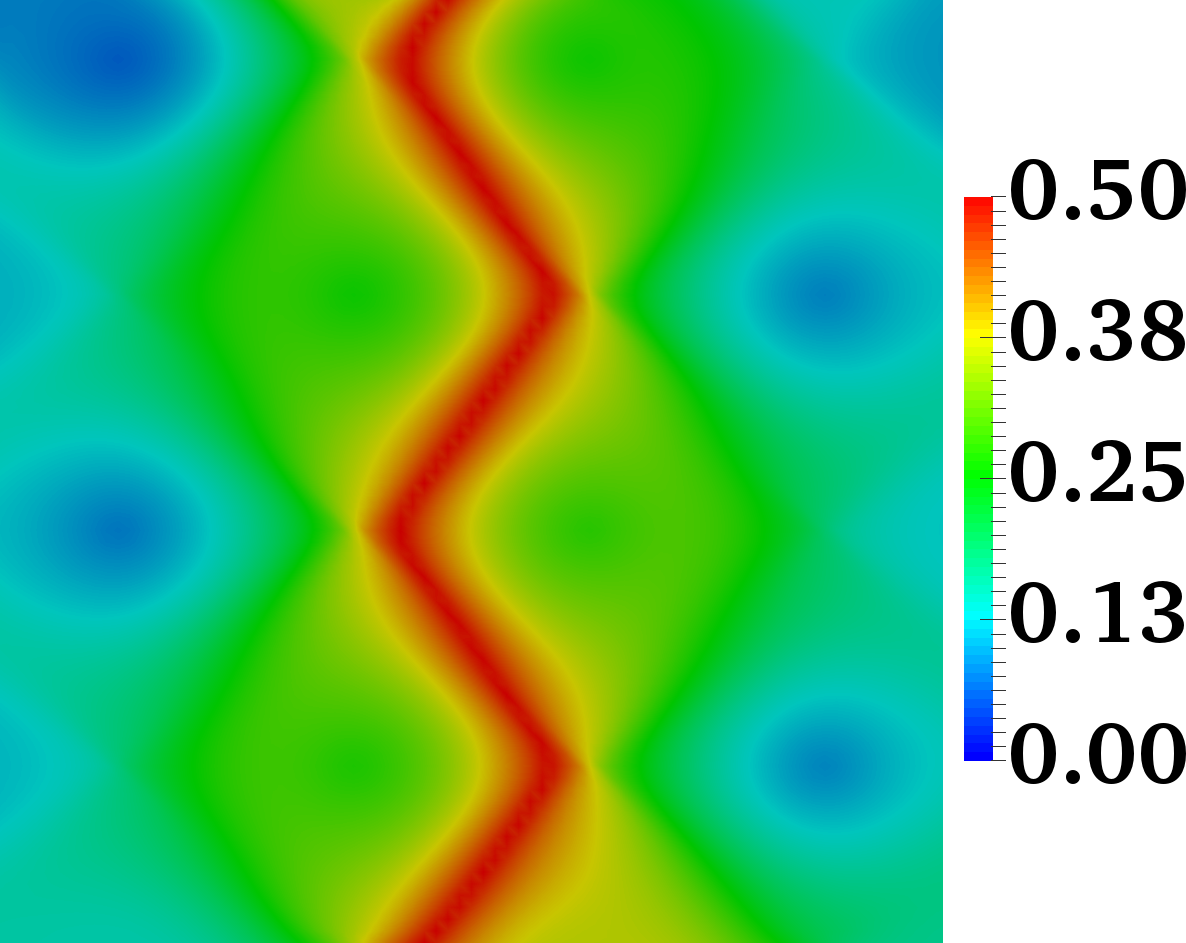}}
  \hspace{-0.5in}
  \subfigure[$\kappa_\mathrm{f}L = 2$ and $t = 0.5$]
    {\includegraphics[clip=true,width = 0.3\textwidth]
    {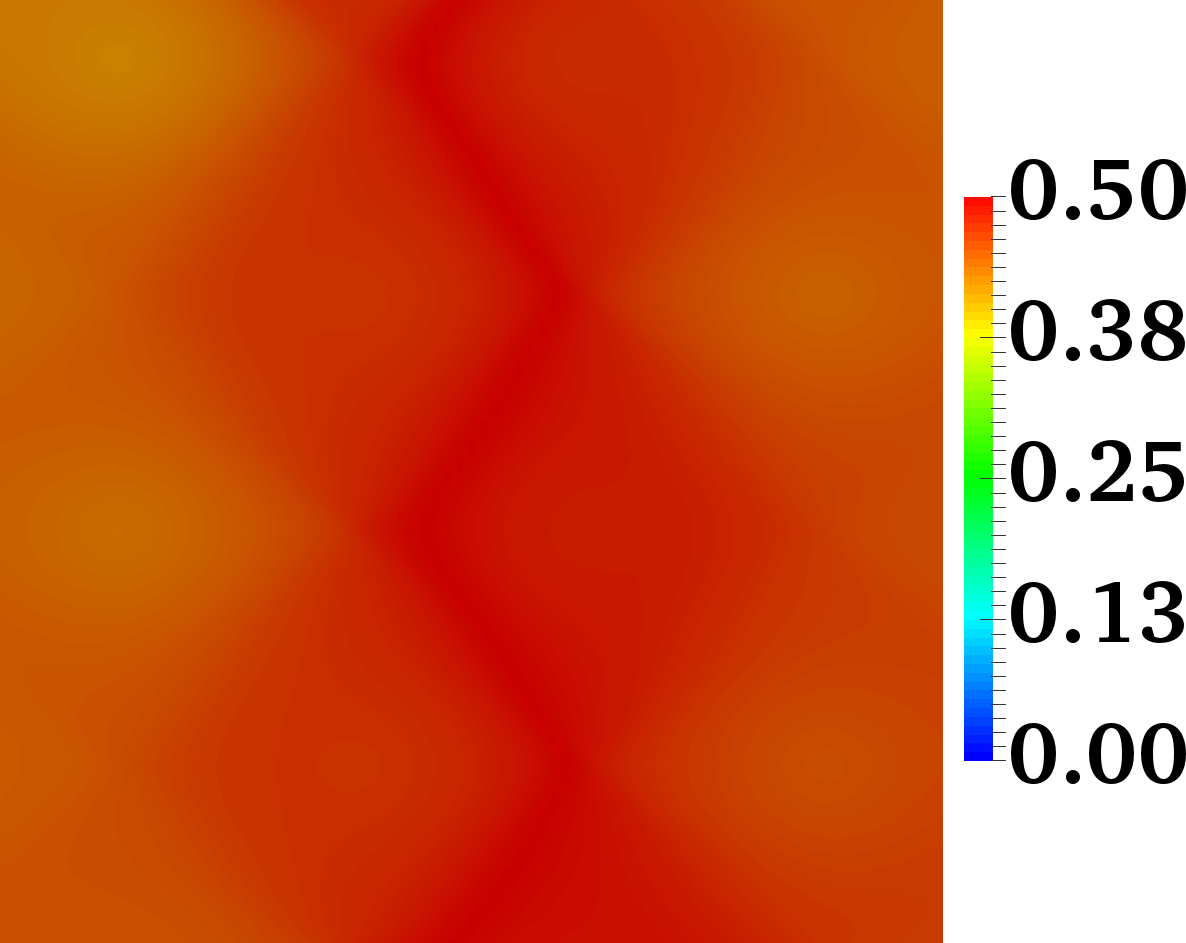}}
  \hspace{-0.5in}
  \subfigure[$\kappa_\mathrm{f}L = 2$ and $t = 1.0$]
    {\includegraphics[clip=true,width = 0.3\textwidth]
    {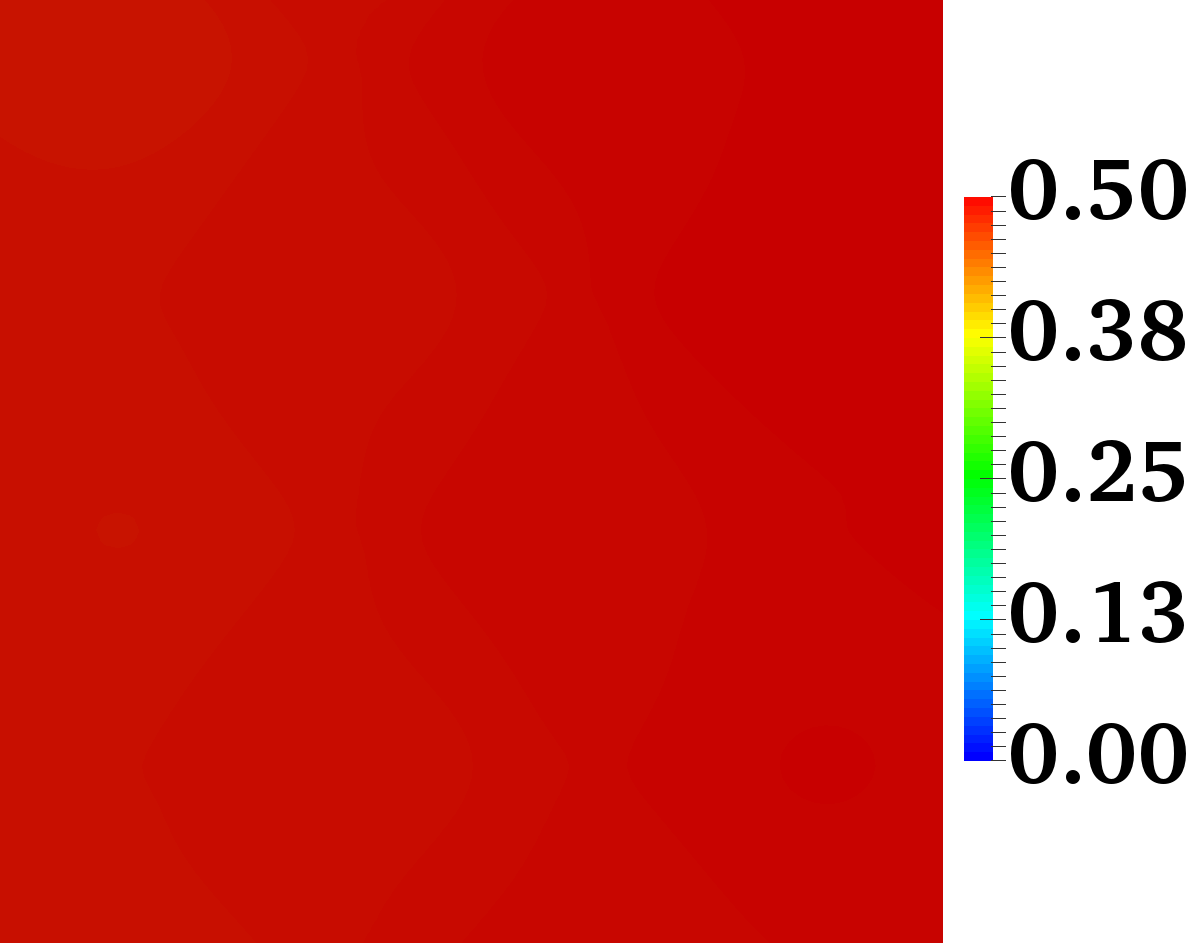}}
  \subfigure[$\kappa_\mathrm{f}L = 3$ and $t = 0.1$]
    {\includegraphics[clip=true,width = 0.3\textwidth]
    {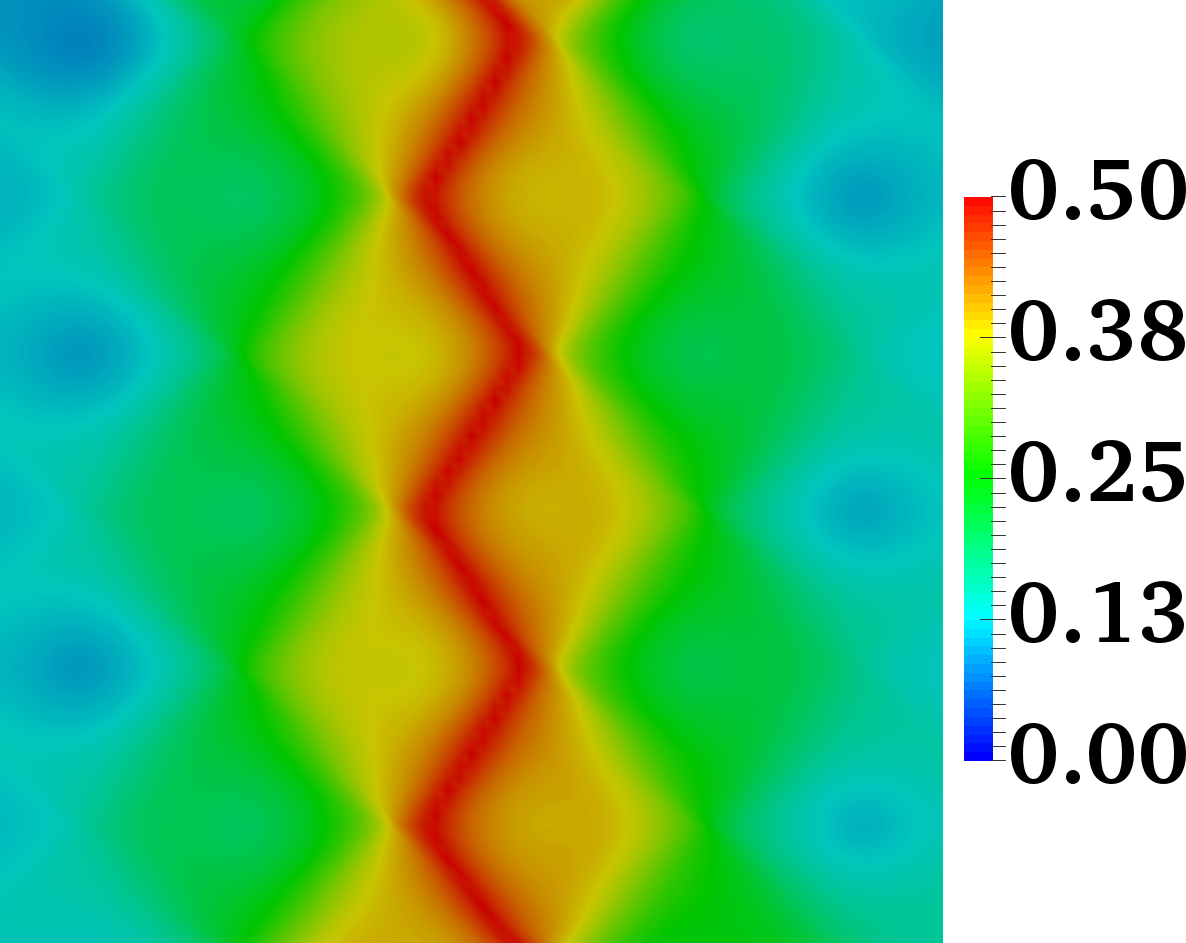}}
  \hspace{-0.5in}
  \subfigure[$\kappa_\mathrm{f}L = 3$ and $t = 0.5$]
    {\includegraphics[clip=true,width = 0.3\textwidth]
    {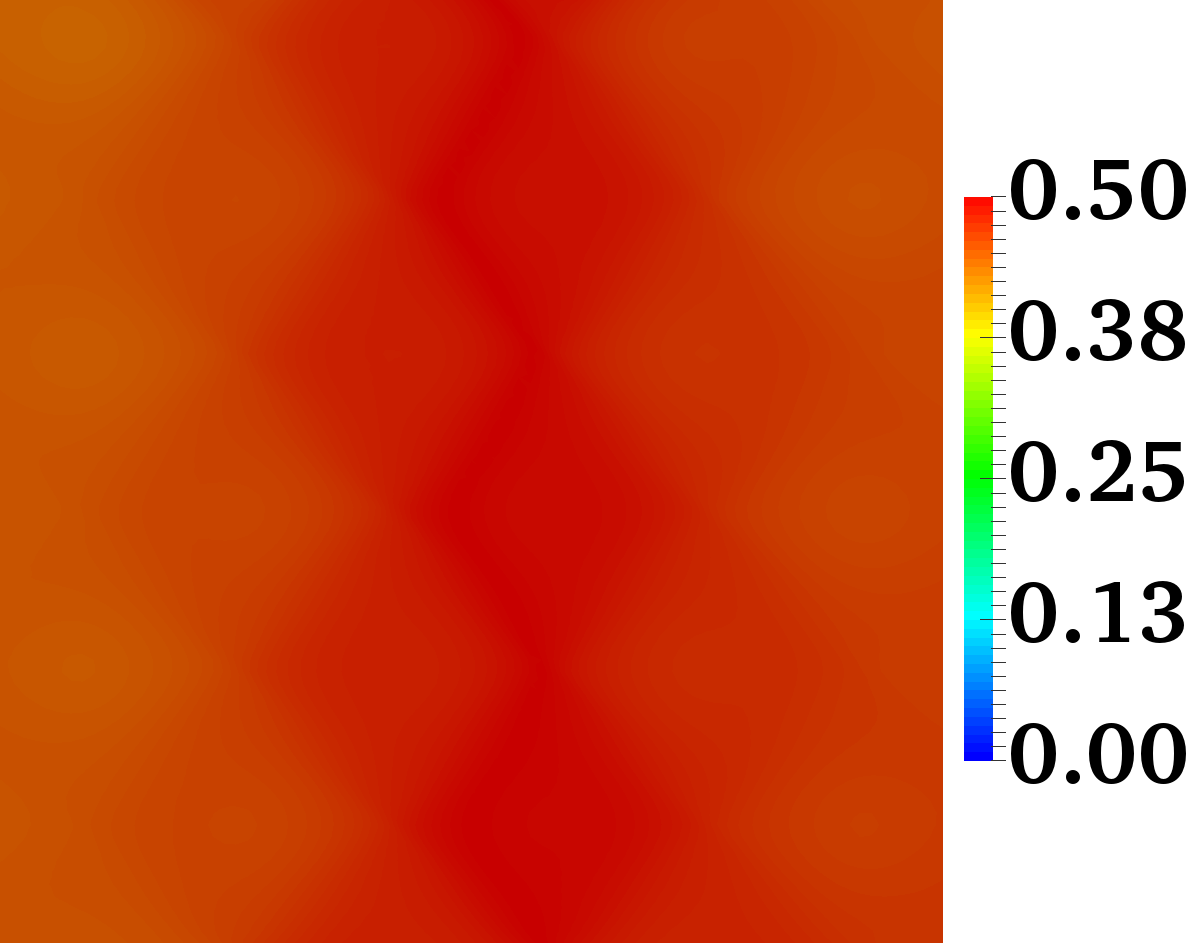}}
  \hspace{-0.5in}
  \subfigure[$\kappa_\mathrm{f}L = 3$ and $t = 1.0$]
    {\includegraphics[clip=true,width = 0.3\textwidth]
    {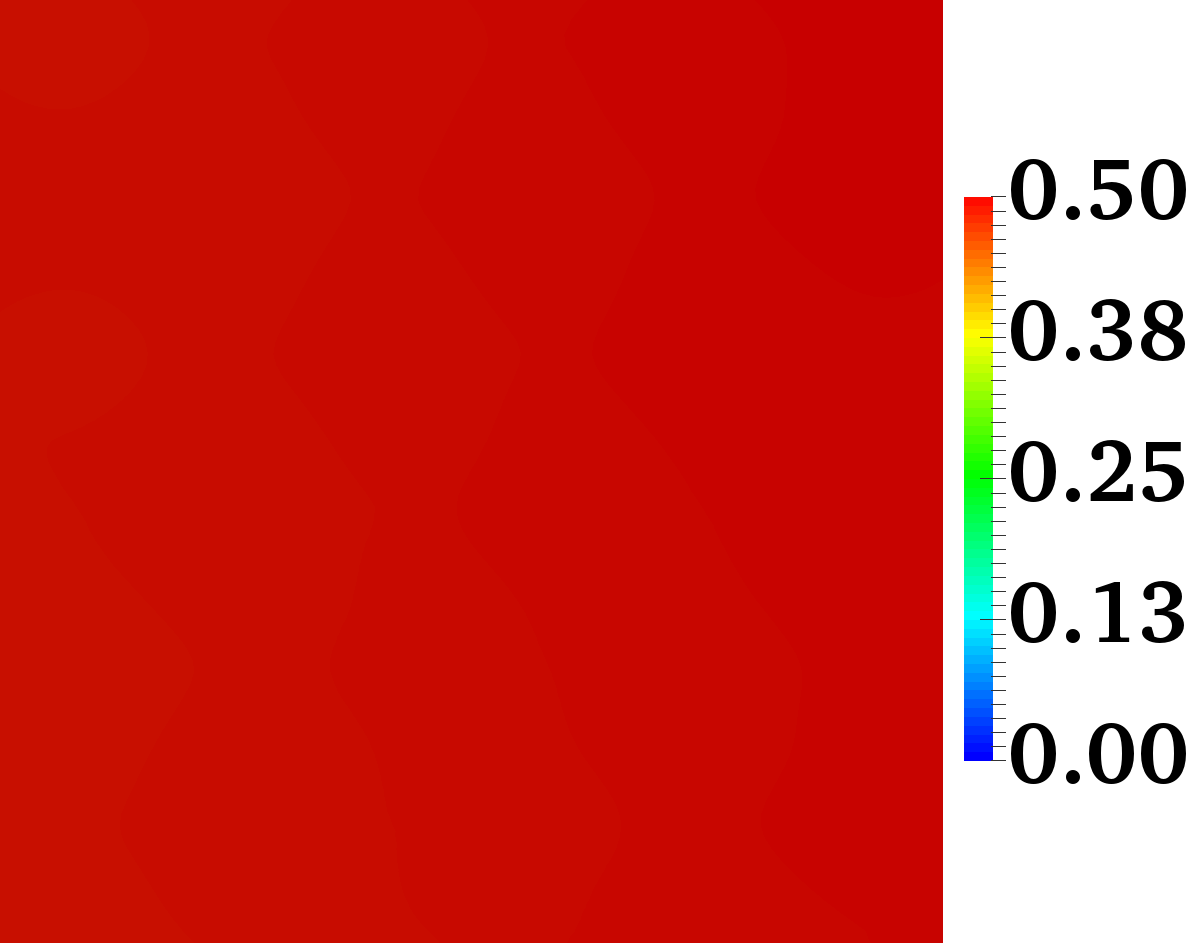}}
  \subfigure[$\kappa_\mathrm{f}L = 4$ and $t = 0.1$]
    {\includegraphics[clip=true,width = 0.3\textwidth]
    {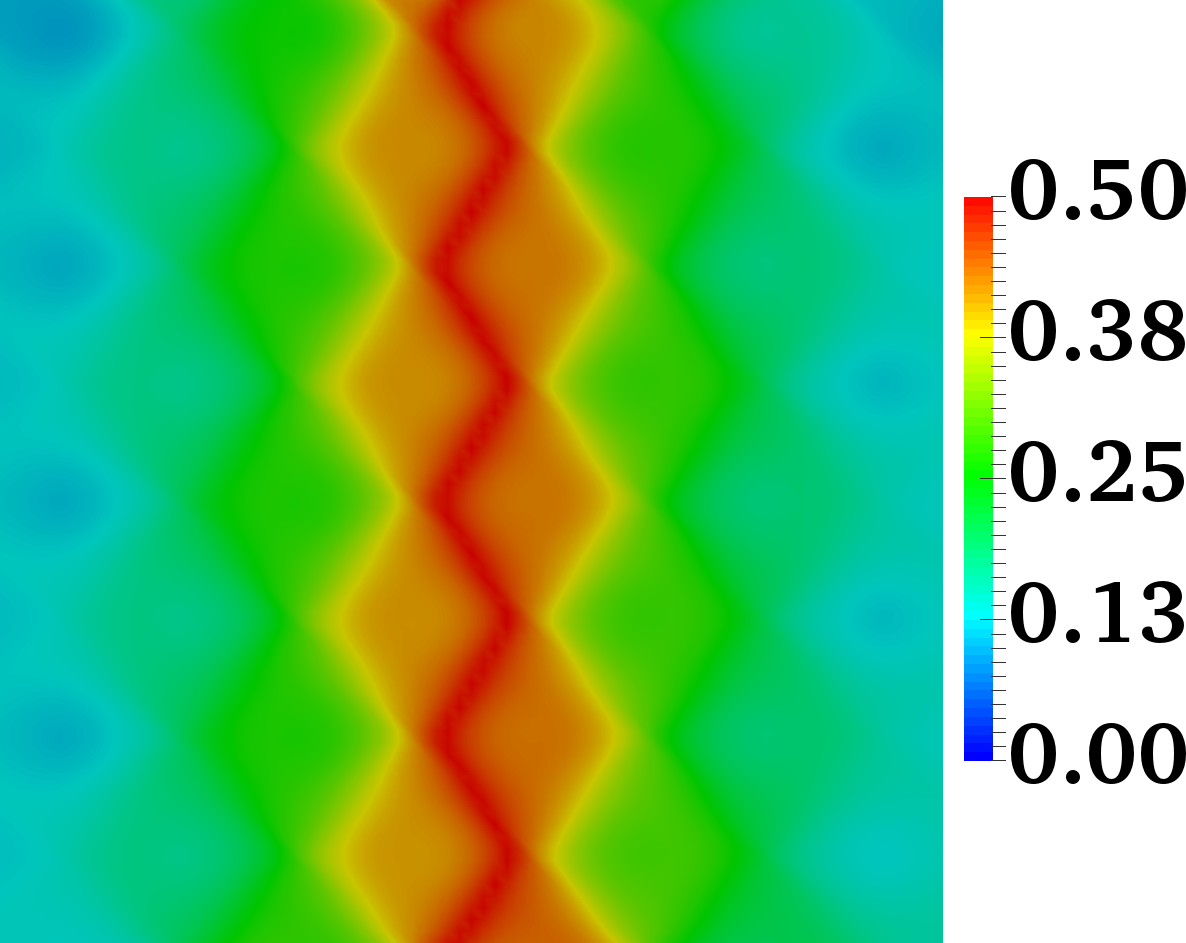}}
  \hspace{-0.5in}
  \subfigure[$\kappa_\mathrm{f}L = 4$ and $t = 0.5$]
    {\includegraphics[clip=true,width = 0.3\textwidth]
    {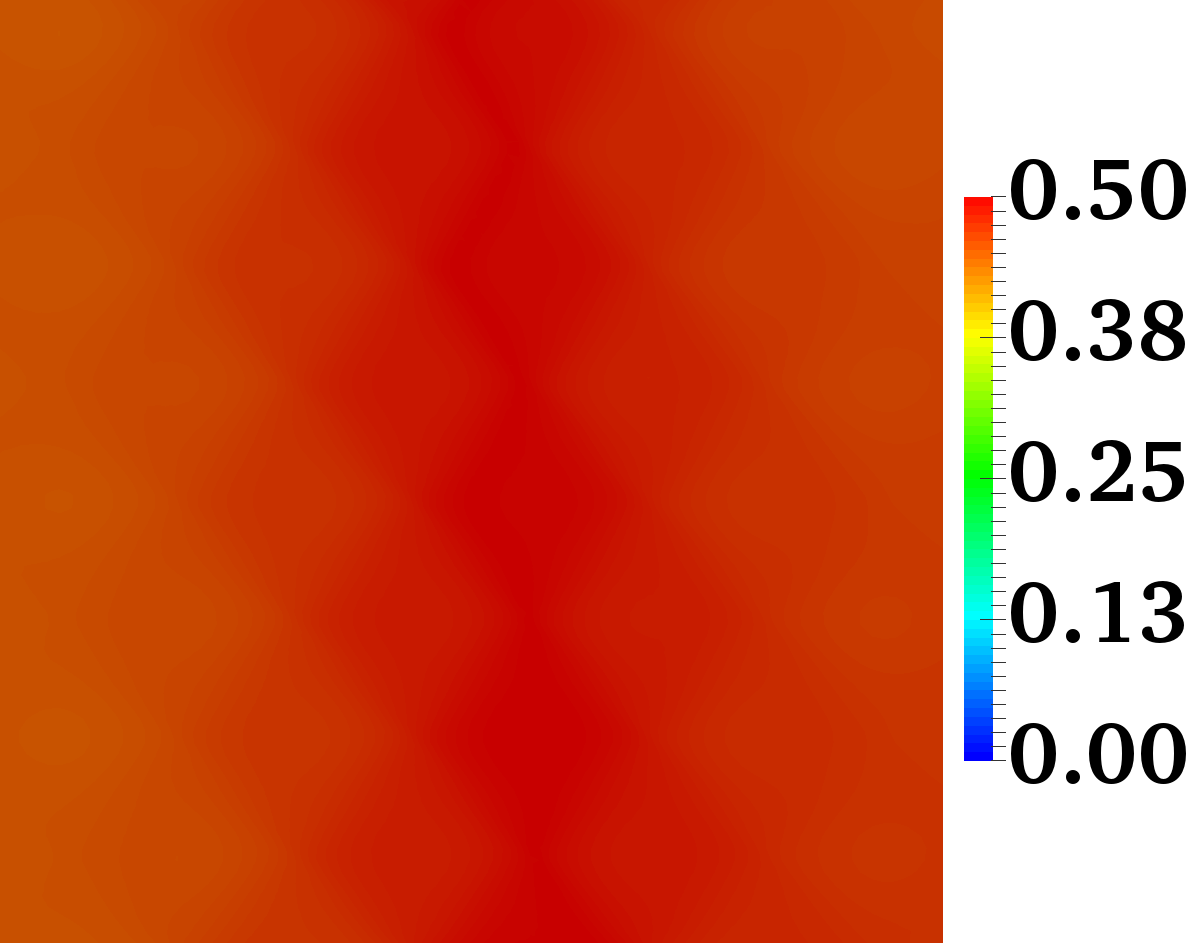}}
  \hspace{-0.5in}
  \subfigure[$\kappa_\mathrm{f}L = 4$ and $t = 1.0$]
    {\includegraphics[clip=true,width = 0.3\textwidth]
    {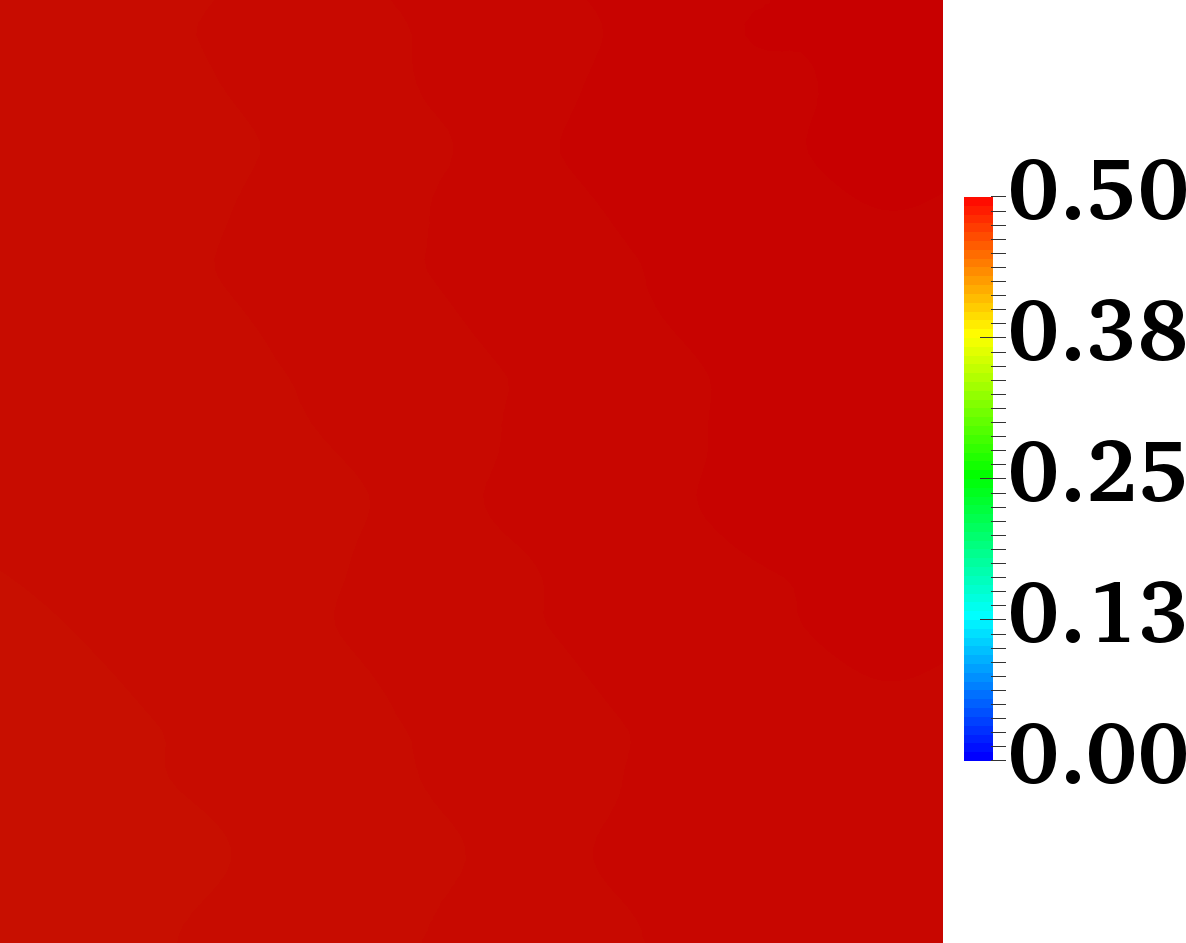}}
  \subfigure[$\kappa_\mathrm{f}L = 5$ and $t = 0.1$]
    {\includegraphics[clip=true,width = 0.3\textwidth]
    {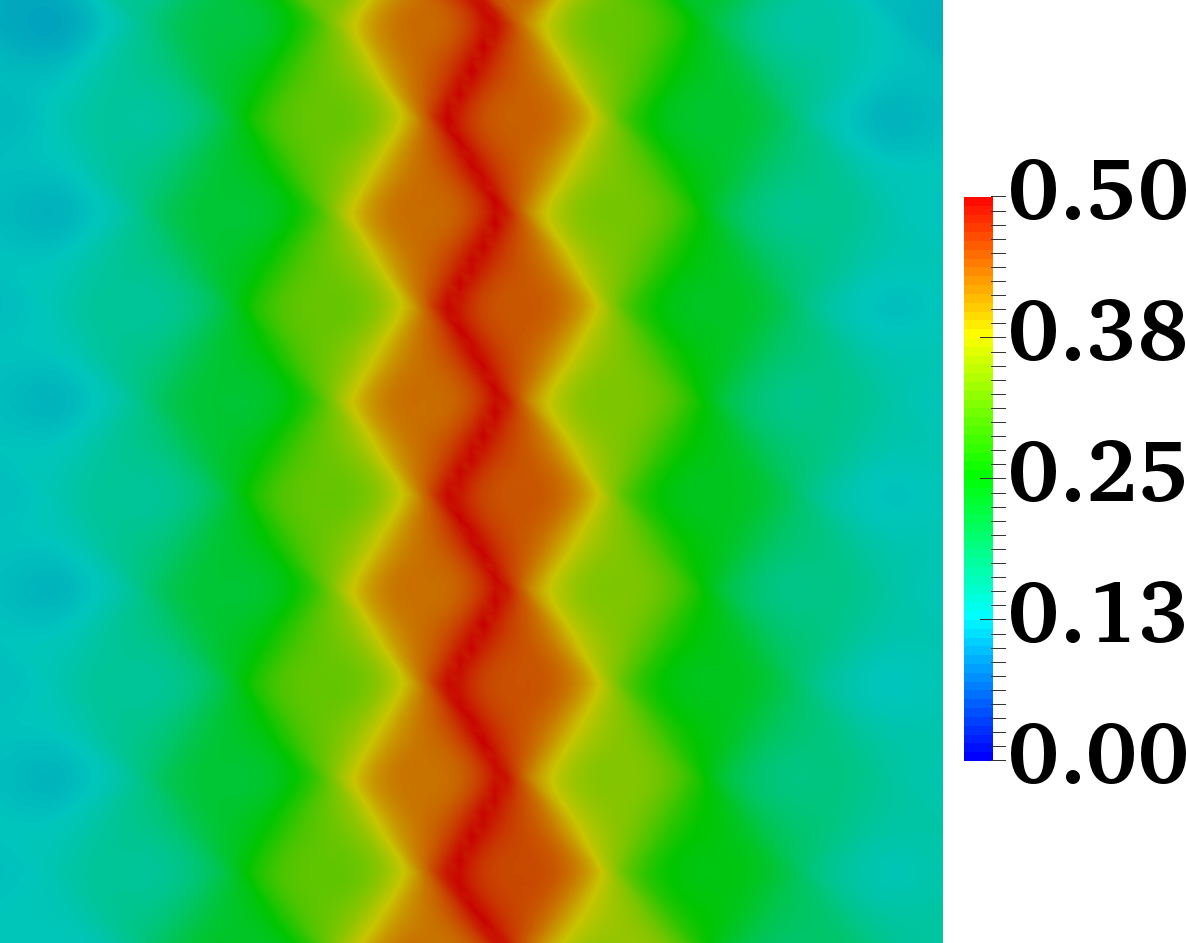}}
  \hspace{-0.5in}
  \subfigure[$\kappa_\mathrm{f}L = 5$ and $t = 0.5$]
    {\includegraphics[clip=true,width = 0.3\textwidth]
    {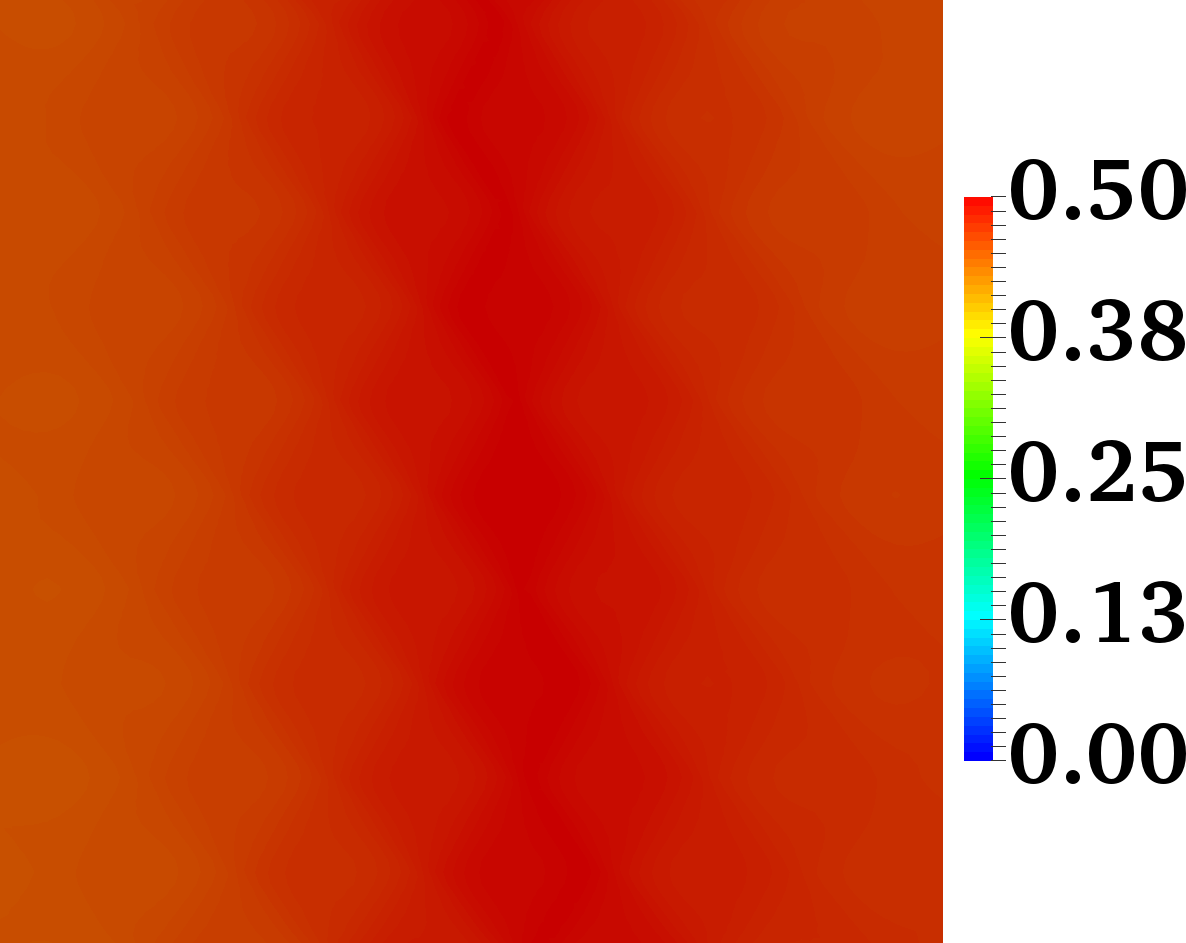}}
  \hspace{-0.5in}
  \subfigure[$\kappa_\mathrm{f}L = 5$ and $t = 1.0$]
    {\includegraphics[clip=true,width = 0.3\textwidth]
    {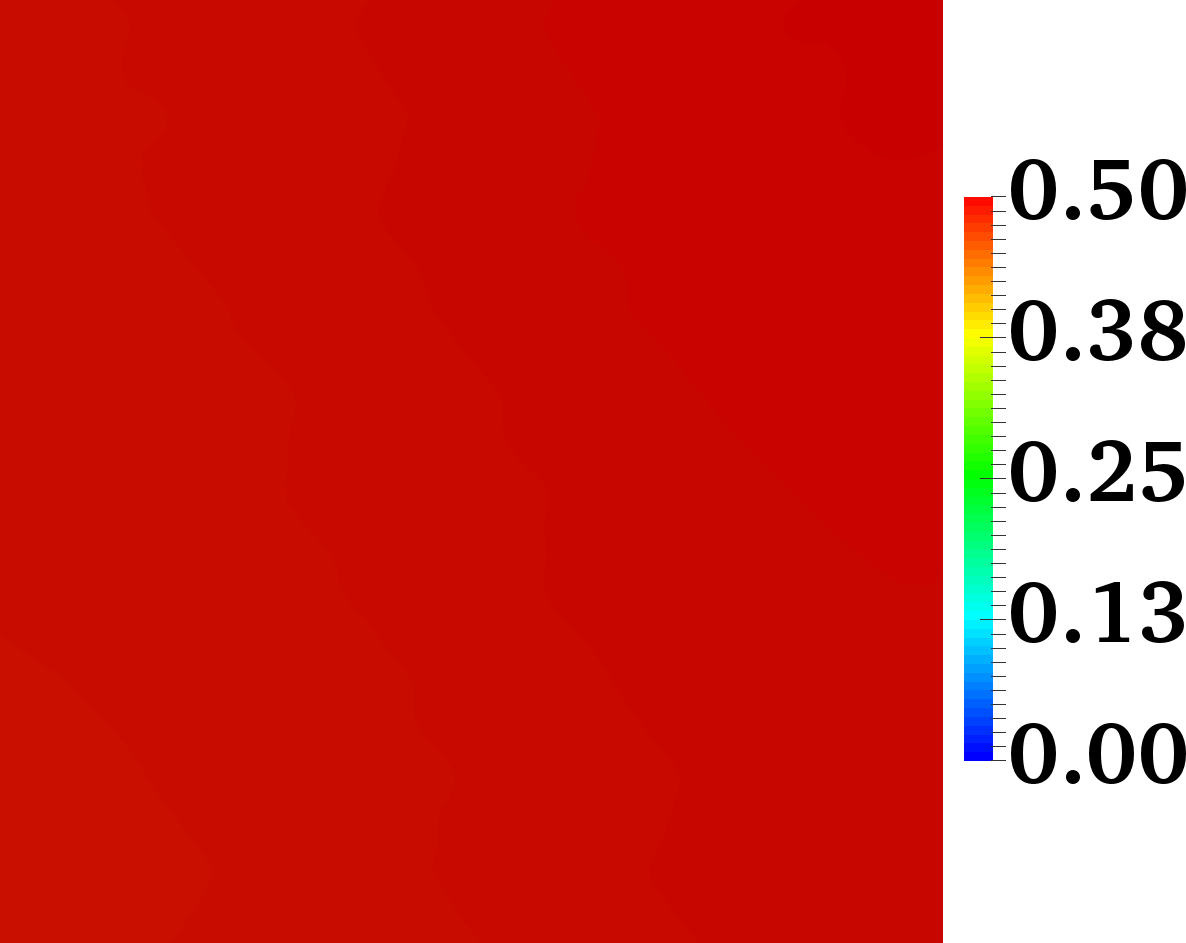}}
  \caption{\textrm{\textbf{Concentration contours of species $C$ under low anisotropy:}}~Concentration of product $C$ at times $t = 0.1, \, 0.5,$  and $1.0$. 
  Other input parameters were $\frac{\alpha_\mathrm{L}}{\alpha_\mathrm{T}} = 10$ (low anisotropy), $v_0 = 1$, $T = 0.1$, and $D_\mathrm{m} = 10^{-3}$. 
  At low anisotropy, production of $C$ increased.
  During late times (e.g., $t = 0.5$  and $1.0$), diffusion dominates $C$ production while $\kappa_\mathrm{f}L$ and $\frac{\alpha_\mathrm{L}}{\alpha_\mathrm{T}}$ minimally affect $C$ production.
  \label{Fig:Contours_C_Difftimes_3}}
\end{figure}

After time $t = 0$, reactants $A$ and $B$ are allowed to mix and form product $C$.
The extent of mixing depended upon the reaction-diffusion inputs (features). 
Increased degree of mixing increases the yield of product $C$.
Product $C$ yield at normalized simulation times $t = 0$, 0.5, and 1.0 are shown in Figs.~\ref{Fig:Contours_C_Difftimes_1}-\ref{Fig:Contours_C_Difftimes_3} revealing the significance of $k_\mathrm{f}L$ on product formation at different times.
The importance of $\frac{\alpha_\mathrm{L}}{\alpha_\mathrm{T}}$ on product formation at various times was also evident.
For $k_\mathrm{f}L = 2$ and $\frac{\alpha_\mathrm{L}}{\alpha_\mathrm{T}} = 10^3$ (see Fig.~\ref{Fig:Contours_C_Difftimes_1} (a-c)) at $t = 0.1$, there is little reaction at the center of the vortices. 
However, regions with zero concentration decrease as $k_\mathrm{f}L$ increases.
For example, at $k_\mathrm{f}L = 3$ and $t = 1.0$, more product is formed and negligible zero concentration of $C$ is present in the model domain.
At $k_\mathrm{f}L = 5$ and $t = 1.0$, the system is nearly well-mixed even at high anisotropy.
Because high $k_\mathrm{f}L$ creates a higher number of vortices that enhance reactant mixing, it increases product yield.
Figure~\ref{Fig:Contours_C_Difftimes_2} shows the product $C$ yield under medium anisotropy.
Reducing anisotropy ($\frac{\alpha_\mathrm{L}}{\alpha_\mathrm{T}}$) from 1,000 to 100 improve product yield even under low $\kappa_\mathrm{f}L$ (see Fig.~\ref{Fig:Contours_C_Difftimes_2}(c)).
Among $\frac{\alpha_\mathrm{L}}{\alpha_\mathrm{T}}$, $k_\mathrm{f}L$, and $D_\mathrm{m}$, $\frac{\alpha_\mathrm{L}}{\alpha_\mathrm{T}}$ controls the reaction at early times while $k_\mathrm{f}L$ and $D_\mathrm{m}$ controls reaction at late times.   
Higher values of $\frac{\alpha_\mathrm{L}}{\alpha_\mathrm{T}}$ decreases product yield but higher values of  $k_\mathrm{f}L$ and $D_\mathrm{m}$ increases the product yield. 

\begin{figure}
  \centering
  \subfigure[Species $C$:~RF]
    {\includegraphics[width=0.5\textwidth]{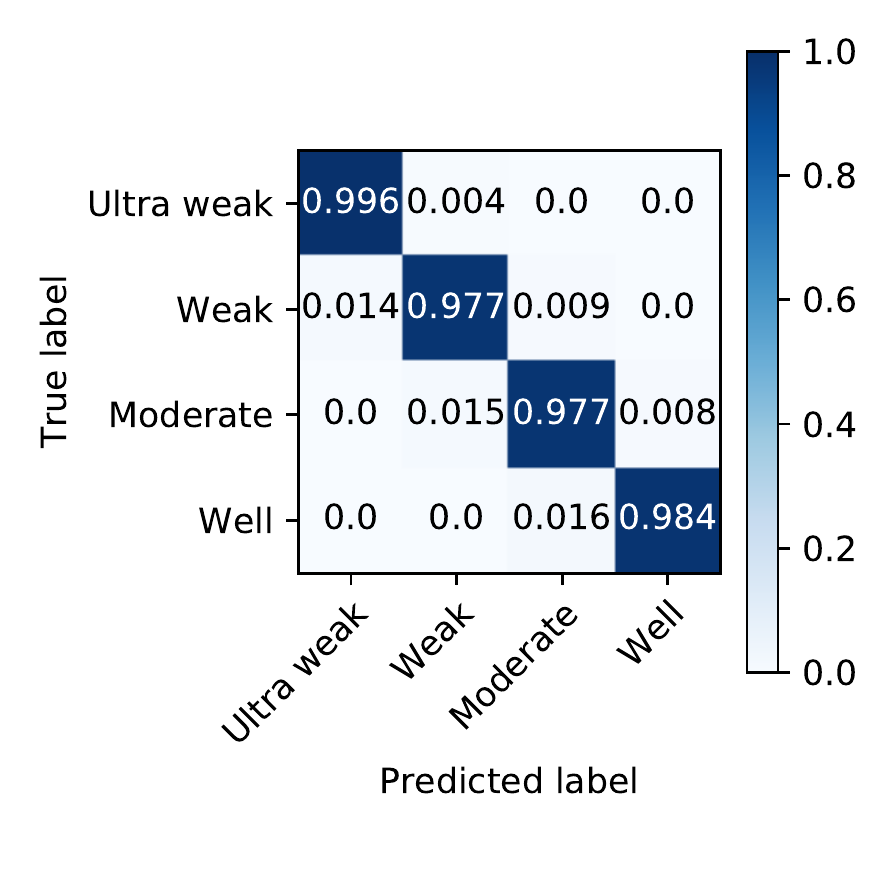}}
  \hspace{-0.6in}
  \subfigure[Species $C$:~MLP]
    {\includegraphics[width=0.5\textwidth]{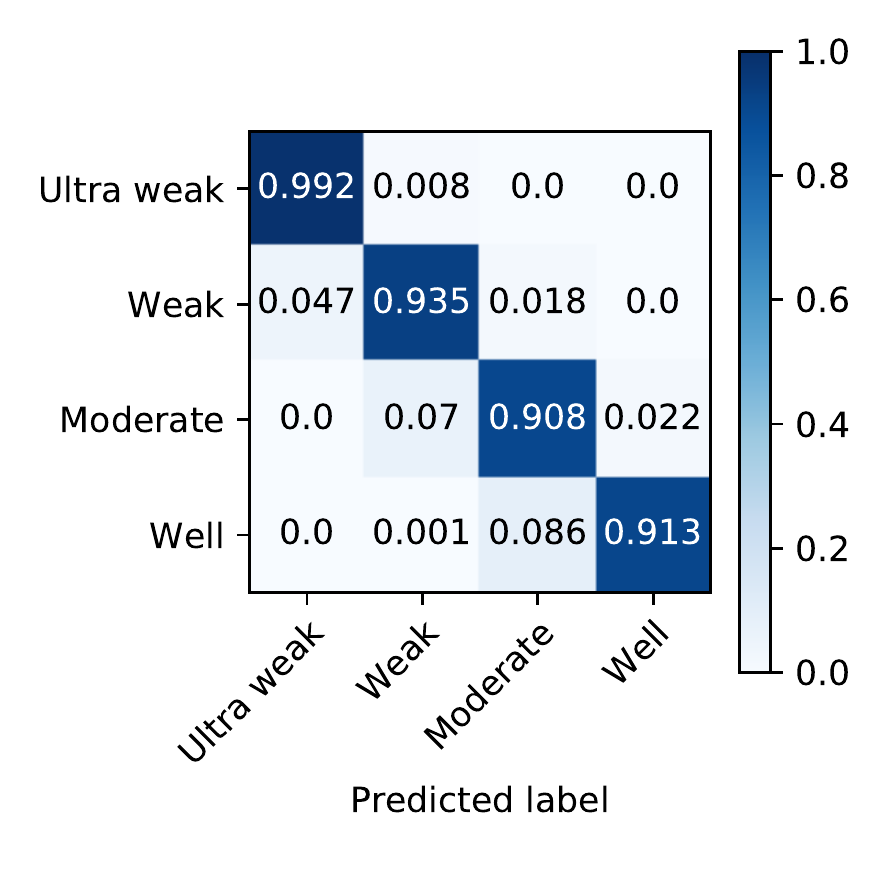}}
\caption{Confusion matrices classifying the degree of mixing for the RF (left) and MLP (right) emulators.}
\label{fig:confusion_matrix}
\end{figure}

ML emulators are also used to classify the mixing state of the system.
Out of 20 ML emulators, only LogR, LDA, QDA, RF, and MLP are used for classification.
Table~\ref{table:Lin_classifiers_performance} shows the training score, testing score, sample sizes, and training time for each linear ML emulator.
Because the progress of reactive-mixing is nonlinear, linear ML emulators (e.g.,~LogR, LDA, QDA) fail to learn an accurate function for the state of mixing.
Mixing state classification by linear classifiers on training and testing data have accuracies \textless80\%. 
Nonlinear classifiers such as RF and MLP learn better functions are quite accurate, \textgreater95\%. 
Results from RF and MLP are used to plot the confusion matrix of Figure~\ref{fig:confusion_matrix} to show true and false predictions.
Confusion matrix for RF and MLP are constructed using approximately 1\% of data (23 simulations as training data) while the remaining 99\% (2,292 simulations) data are used as testing data.
In the confusion matrix, diagonal and off-diagonal elements show true and false predictions, respectively.
The RF and MLP emulators false prediction scores are less than 2\% and 10\%, respectively.
Similar trends are observed for species $A$ and $B$, hence the confusion matrices for them are not shown here.

Table~\ref{table:Lin_regressor_performance} shows the training and testing scores for the six linear ML emulators.
Although training times are short (always \textless20\,minutes), training and testing $R^2$ scores never exceed 73\%.
Also, we apply three Bayesian ML emulators (e.g., BR, GP, NB) to predict QoIs that show similar performance as linear emulators.
Among them, training and testing scores of BR and NB are \textless75\%.
GPs fail to converge for large datasets because of lack of sparsity and due to large training sample size ($\approx \mathcal{O}(10^4) - \mathcal{O}(10^6)$); however, GP trained on a smaller sample size scores >99\%.
This increased prediction capability of GP compared to other Bayesian ML emulators can be attributed to the RBF kernels.
As species $A$ and $B$ decay or product $C$ increases in an exponential fashion, RBF kernels used by GP emulators are better suited to model such a reactive-mixing system.
Hence, GP emulators trained on small (0.25\% of data) data perform best and show promise to predict QoIs.    

\begin{figure}
  \centering
  \subfigure[Species $A$:~$\mathfrak{c}_A$]
    {\includegraphics[clip=true,width = 0.3\textwidth]
    {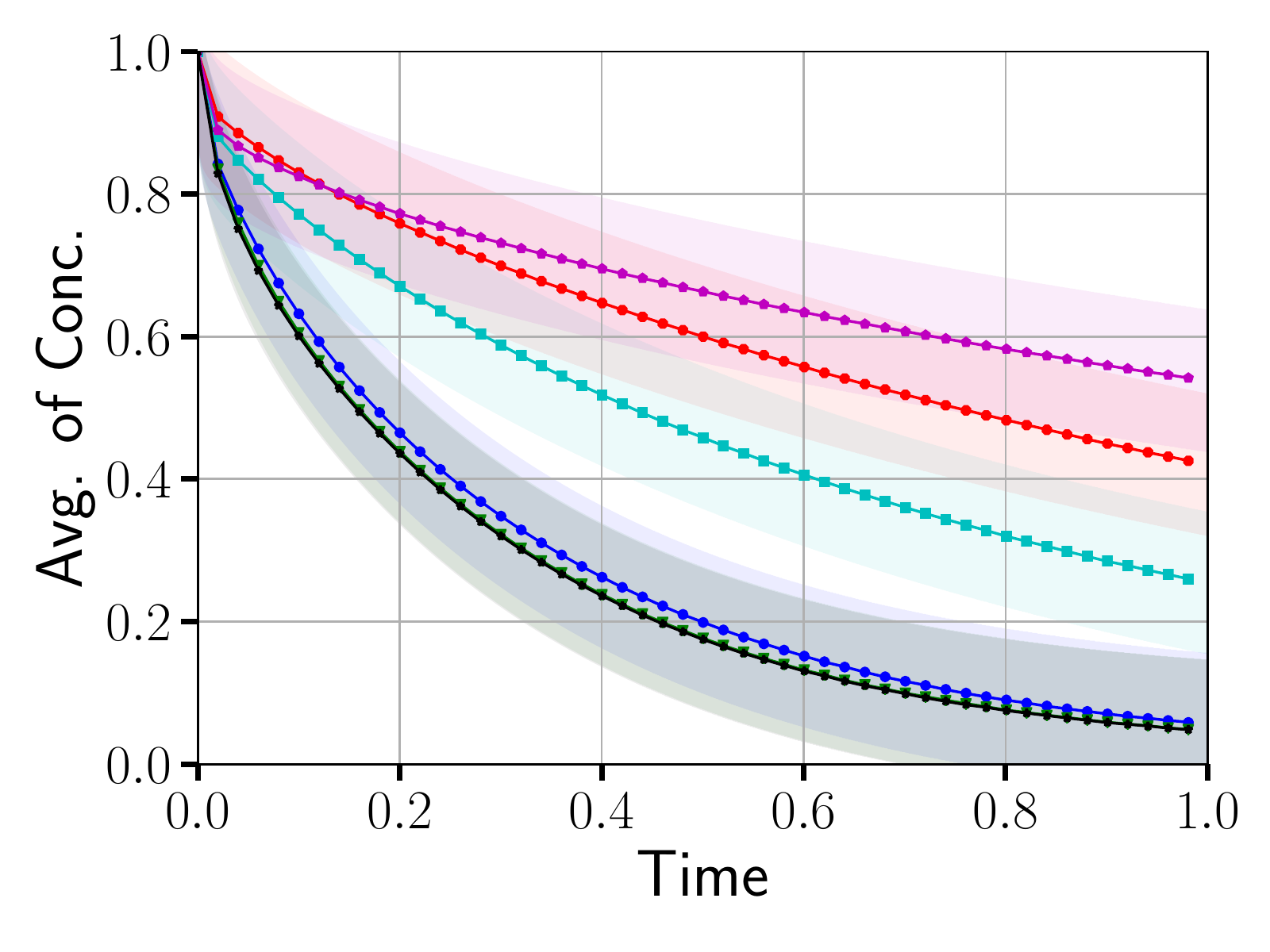}}
  \hspace{-0.1in}
  \subfigure[Species $A$:~$\mathbb{c}_A$]
    {\includegraphics[clip=true,width = 0.3\textwidth]
    {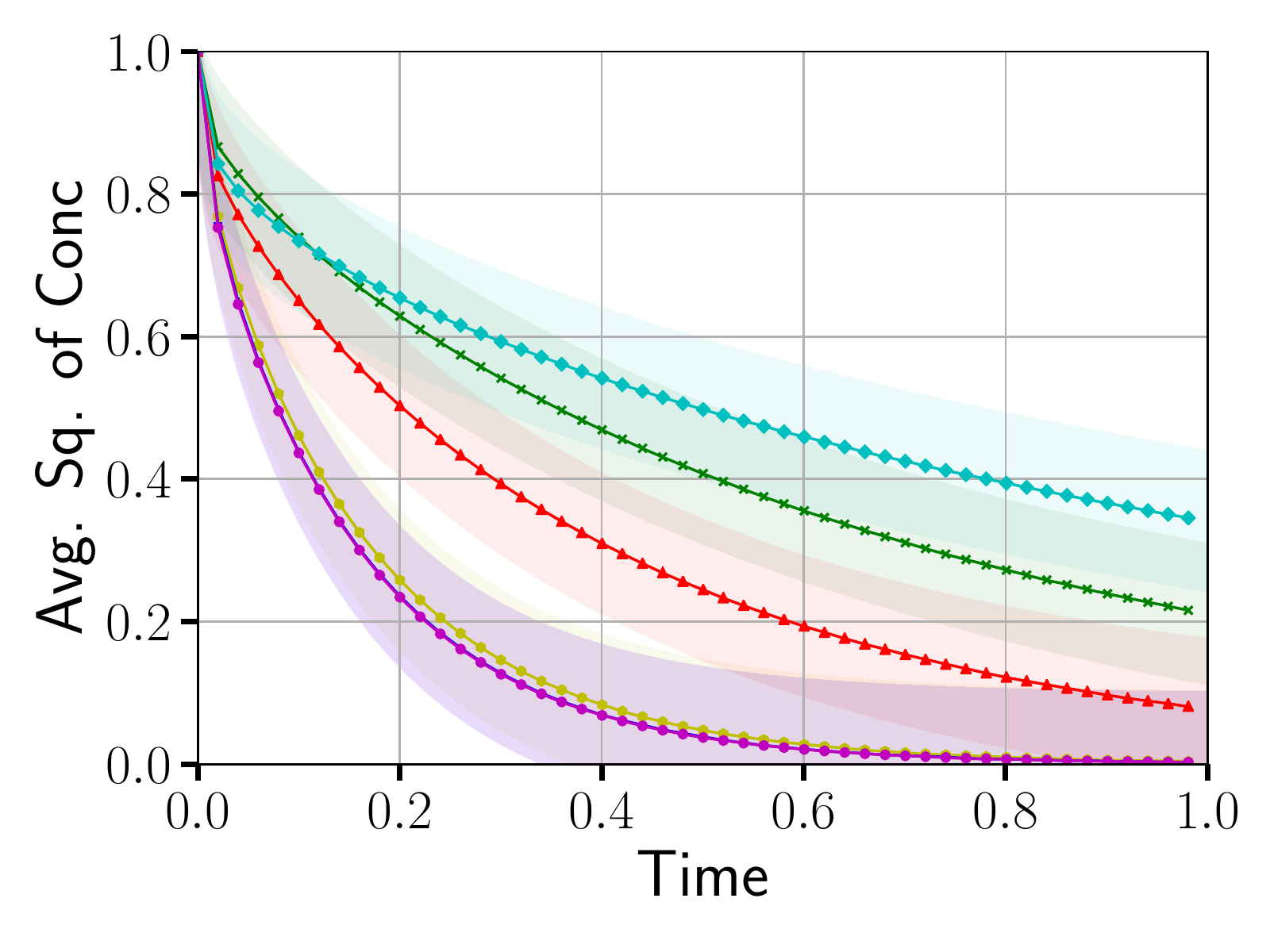}}
  \hspace{-0.1in}
  \subfigure[Species $A$:~$\sigma^2_A$]
    {\includegraphics[clip=true,width = 0.3\textwidth]
    {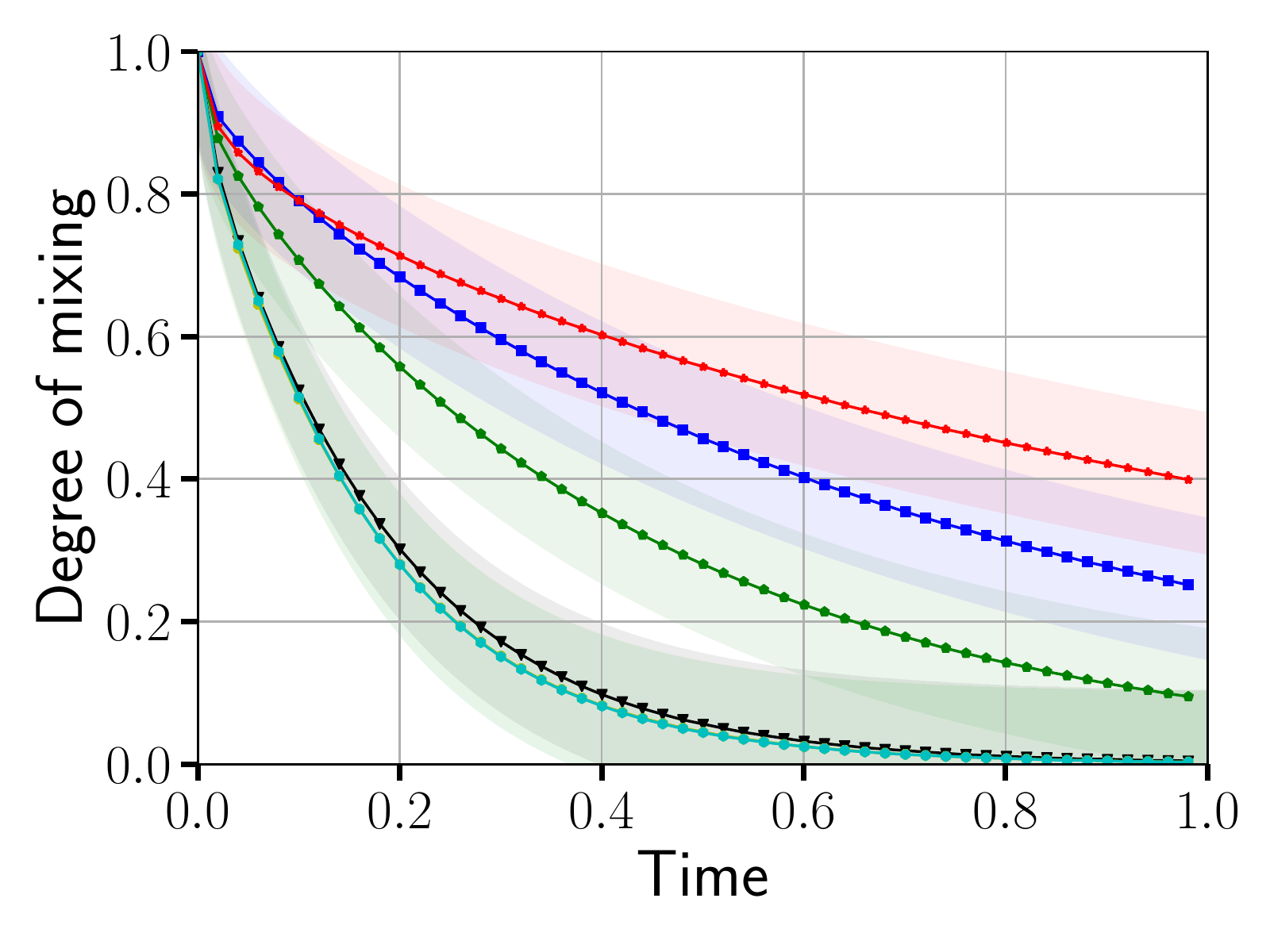}}
  \subfigure[Species $B$:~$\mathfrak{c}_B$]
    {\includegraphics[clip=true,width = 0.3\textwidth]
    {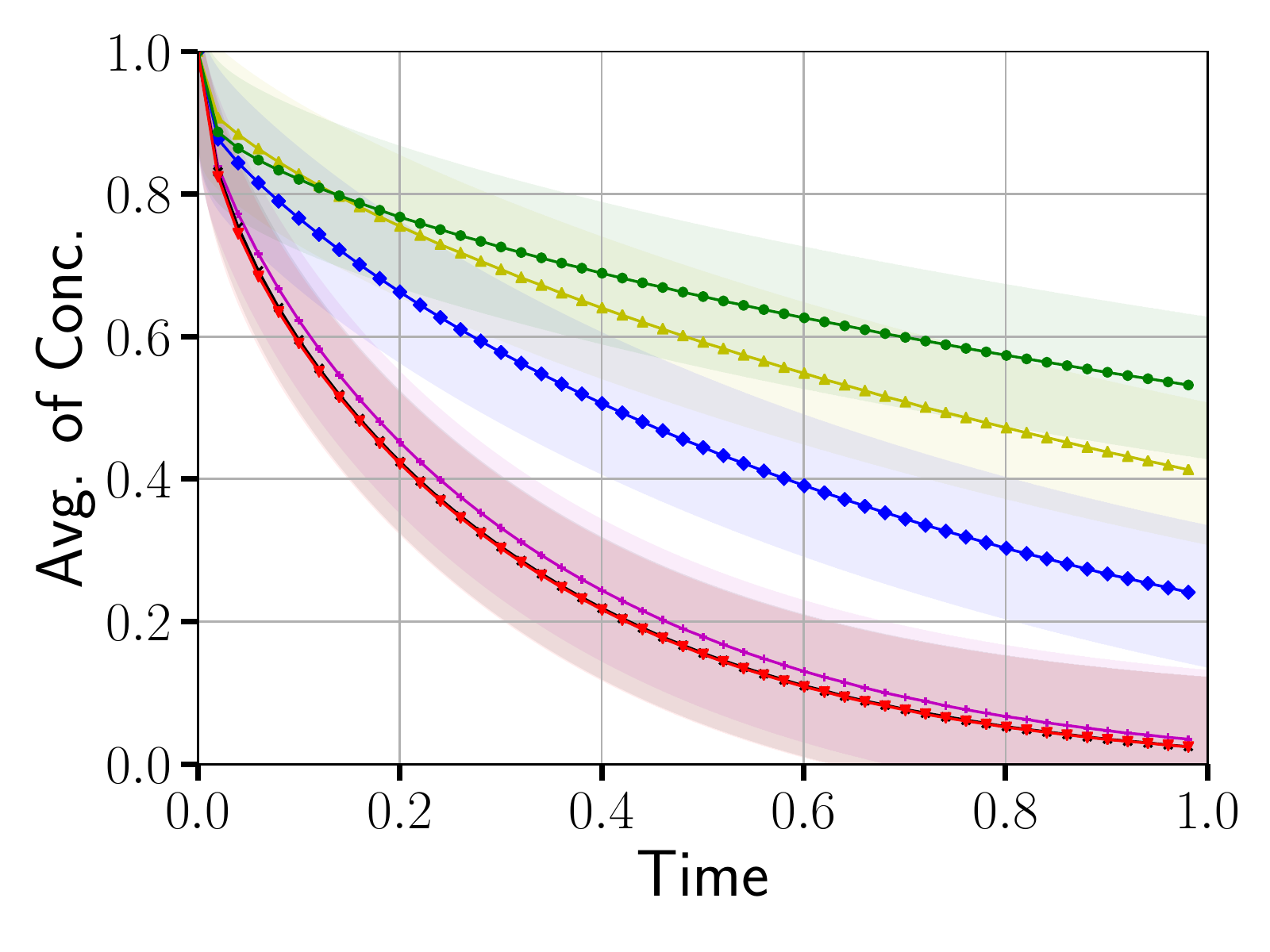}}
  \hspace{-0.1in}
  \subfigure[Species $B$:~$\mathbb{c}_B$]
    {\includegraphics[clip=true,width = 0.3\textwidth]
    {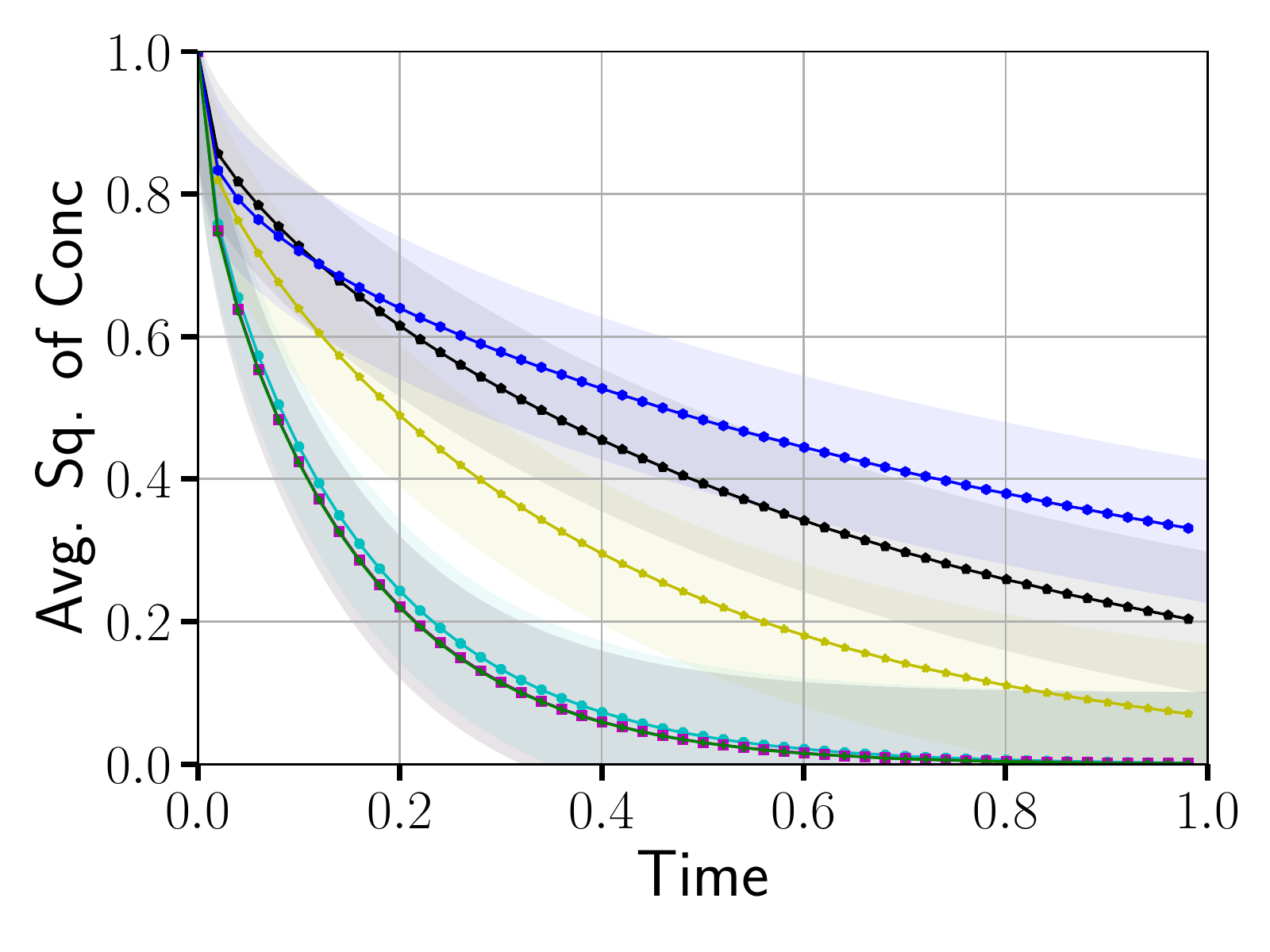}}
  \hspace{-0.1in}
  \subfigure[Species $B$:~$\sigma^2_B$]
    {\includegraphics[clip=true,width = 0.3\textwidth]
    {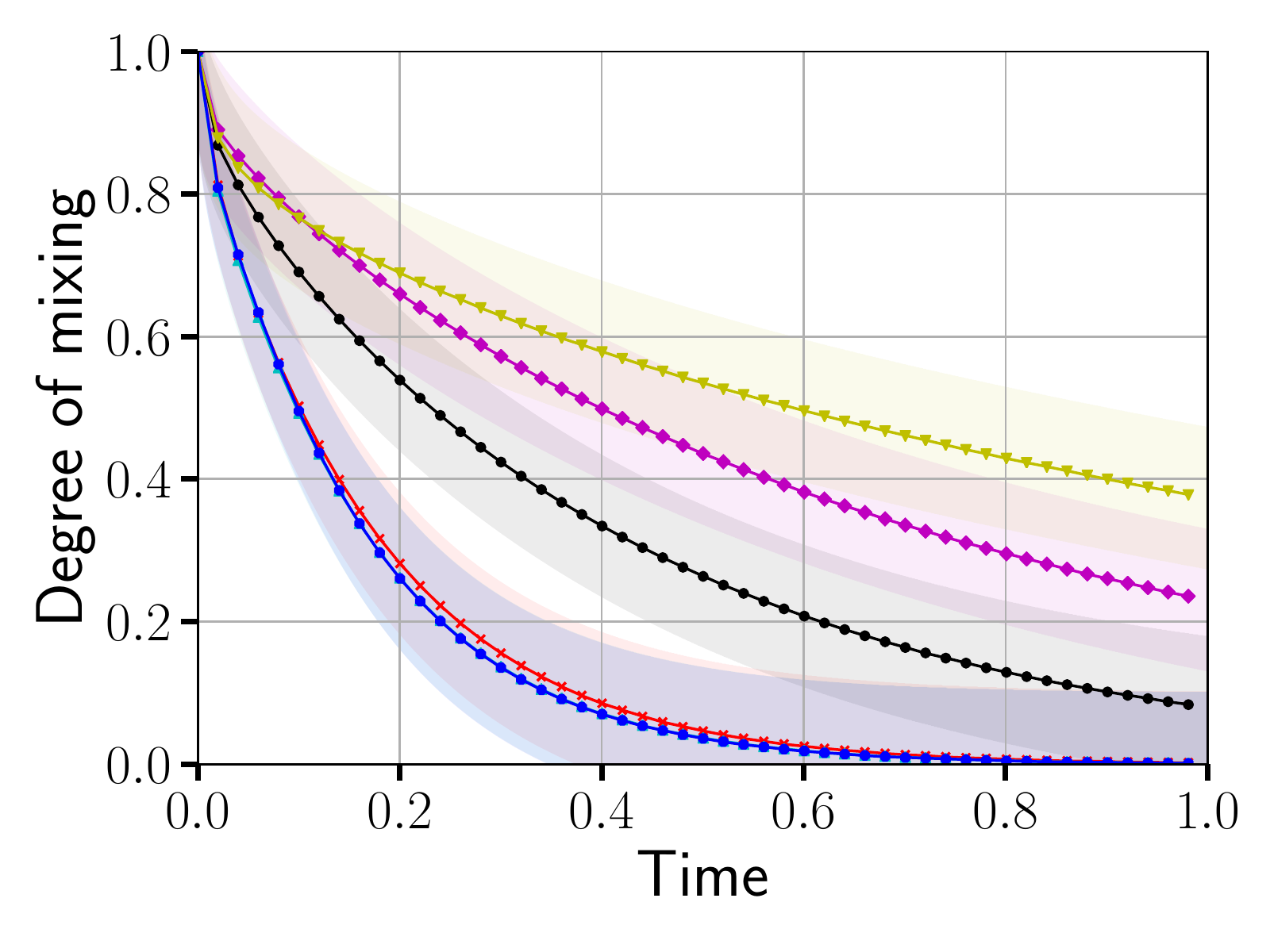}}
  \subfigure[Species $C$:~$\mathfrak{c}_C$]
    {\includegraphics[clip=true,width = 0.3\textwidth]
    {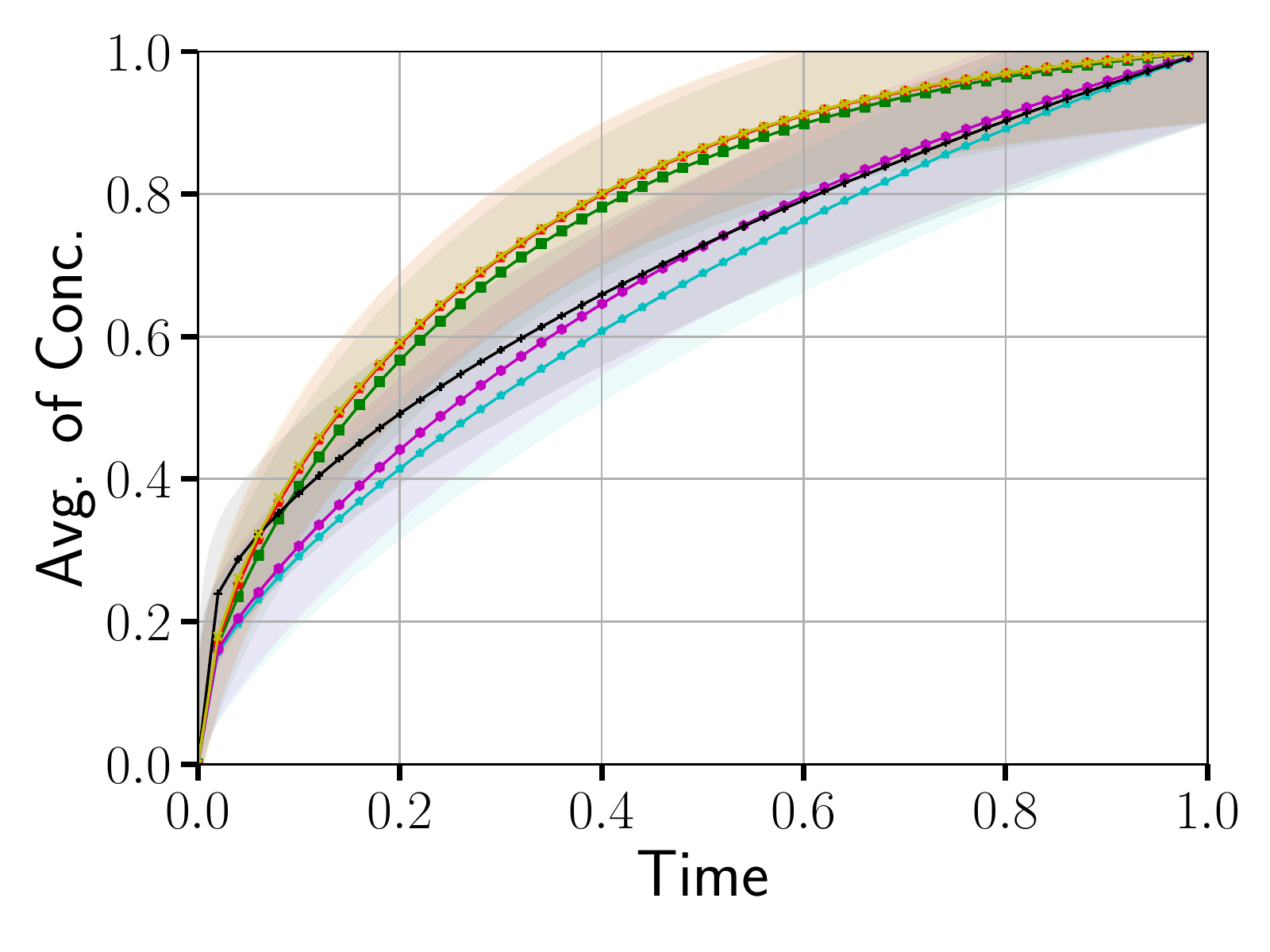}}
  \hspace{-0.1in}
  \subfigure[Species $C$:~$\mathbb{c}_C$]
    {\includegraphics[clip=true,width = 0.3\textwidth]
    {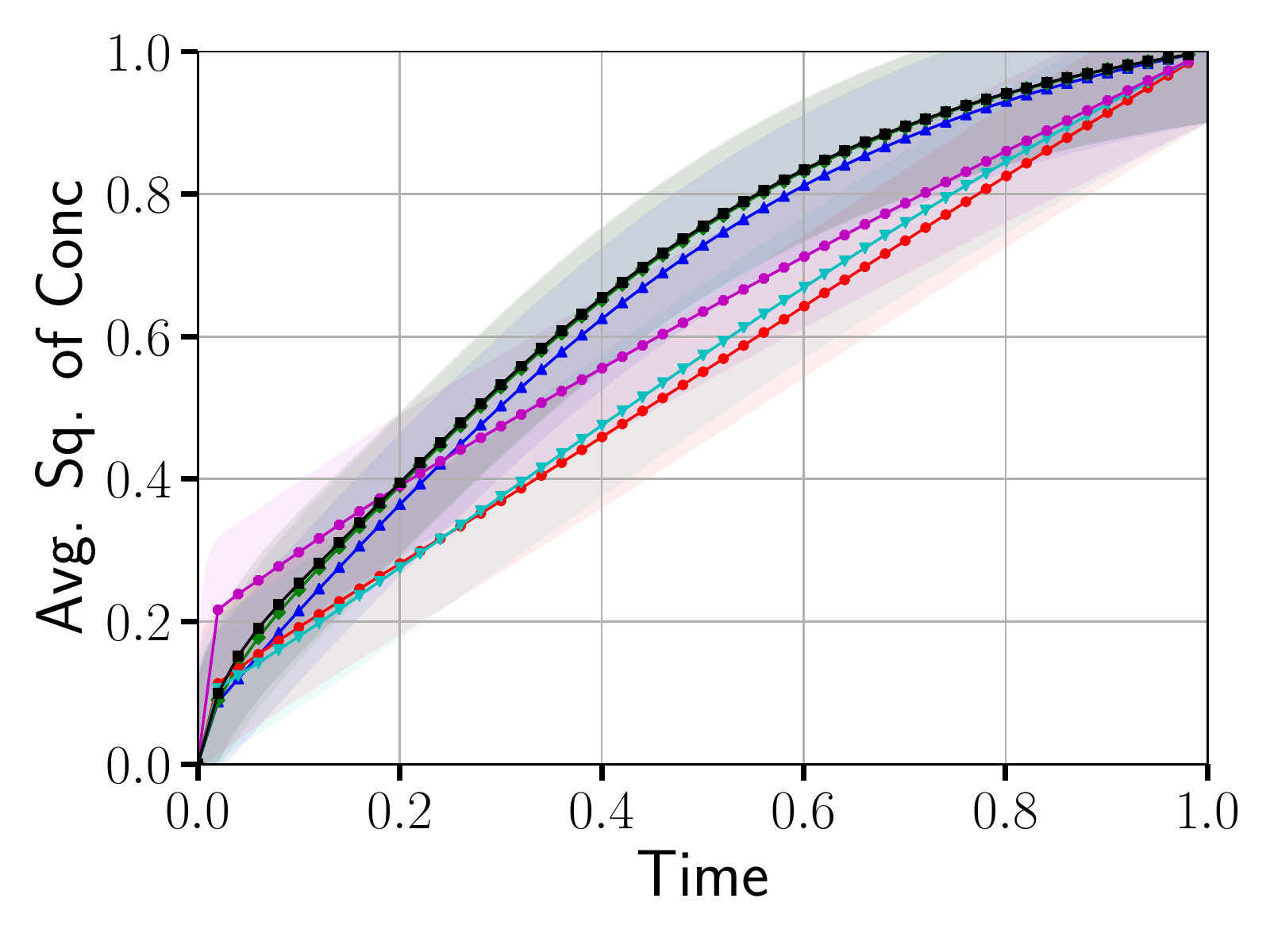}}
  \hspace{-0.1in}
  \subfigure[Species $C$:~$\sigma^2_C$]
    {\includegraphics[clip=true,width = 0.3\textwidth]
    {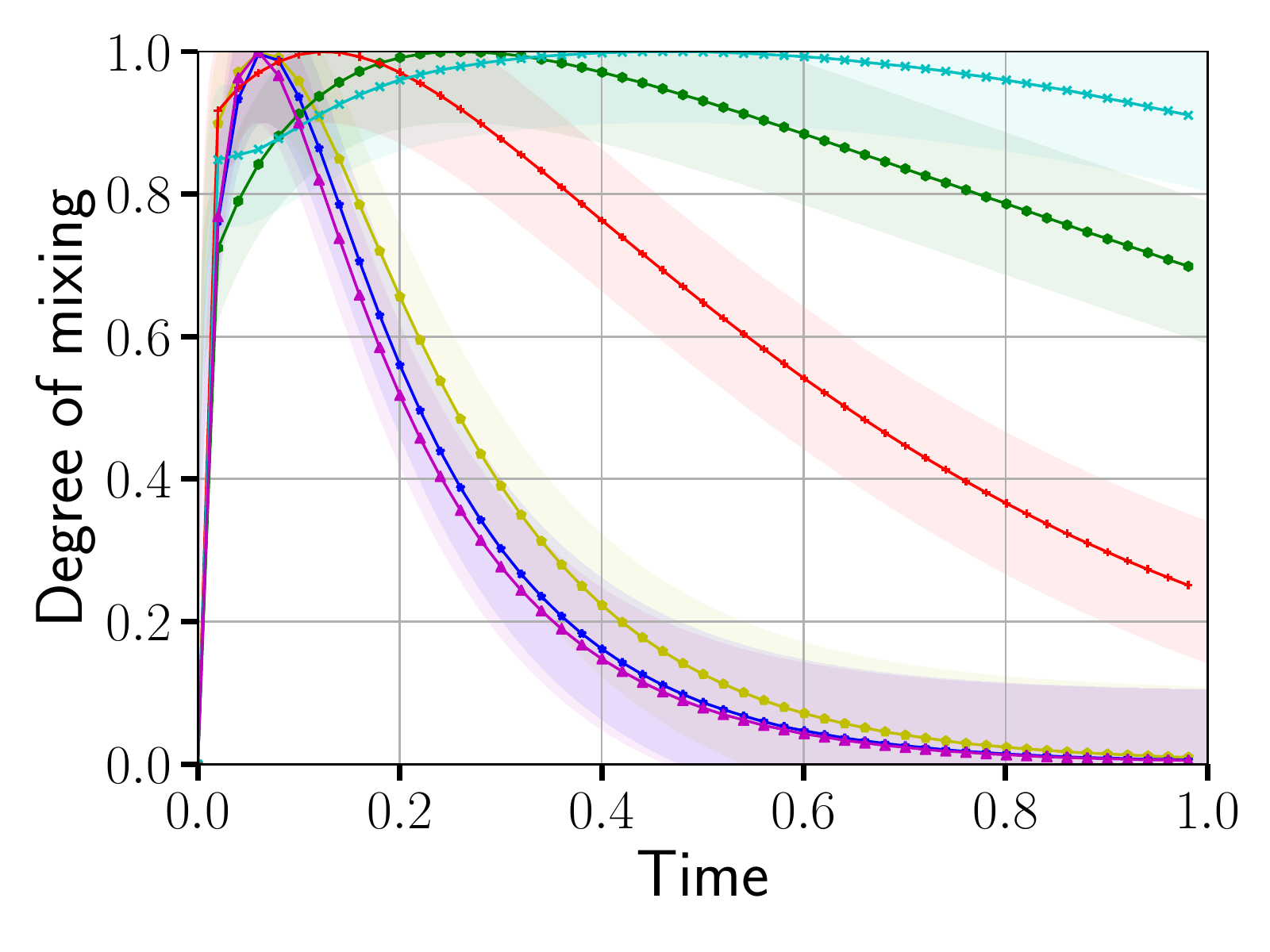}}
  \caption{\textrm{\textbf{Predictions of QoIs by Random Forest emulators for the six unseen realizations:}}~This figure shows the true (markers) and RF emulator predictions (solid curves) of average concentrations, squared of average concentrations, and degree of mixing (a)--(c) of species $A$; (d)--(f) of species $B$, and (g)--(i) of species $C$.}
  \label{Fig:RF_predictions}
\end{figure}

\begin{figure}
  \centering
  \subfigure[Species $A$:~$\mathfrak{c}_A$]
    {\includegraphics[clip=true,width = 0.3\textwidth]
    {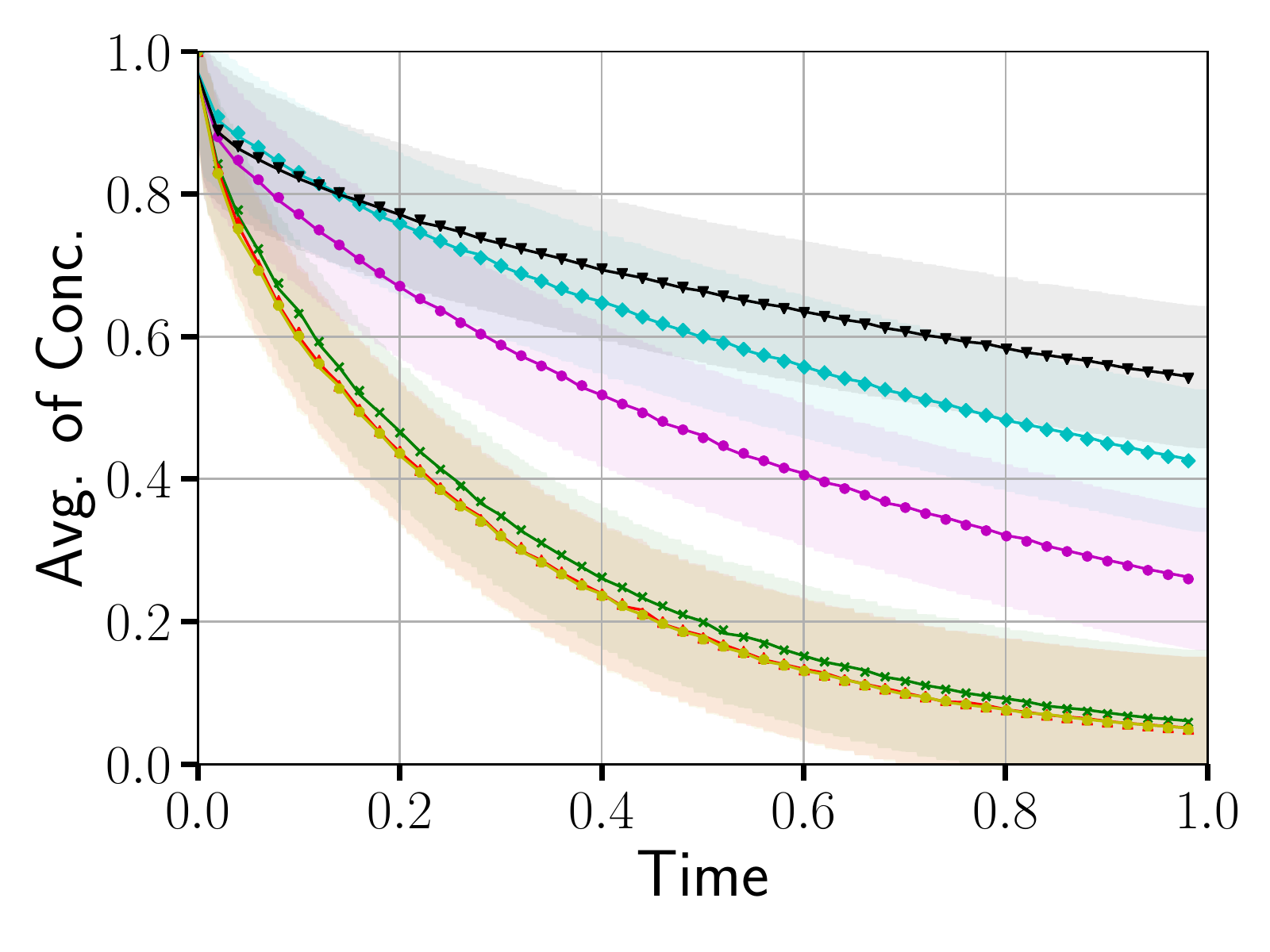}}
  \hspace{-0.1in}
  \subfigure[Species $A$:~$\mathbb{c}_A$]
    {\includegraphics[clip=true,width = 0.3\textwidth]
    {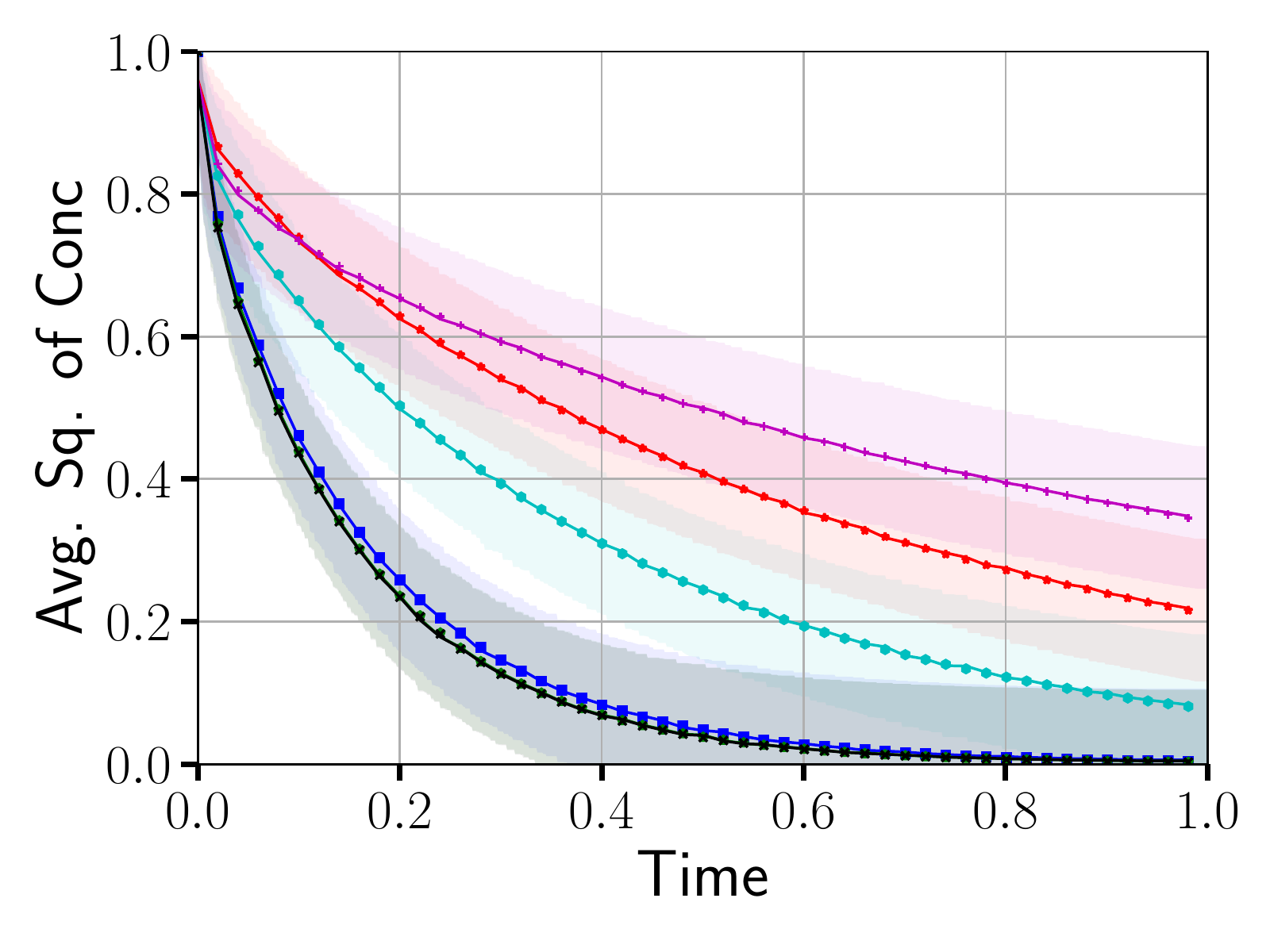}}
  \hspace{-0.1in}
  \subfigure[Species $A$:~$\sigma^2_A$]
    {\includegraphics[clip=true,width = 0.3\textwidth]
    {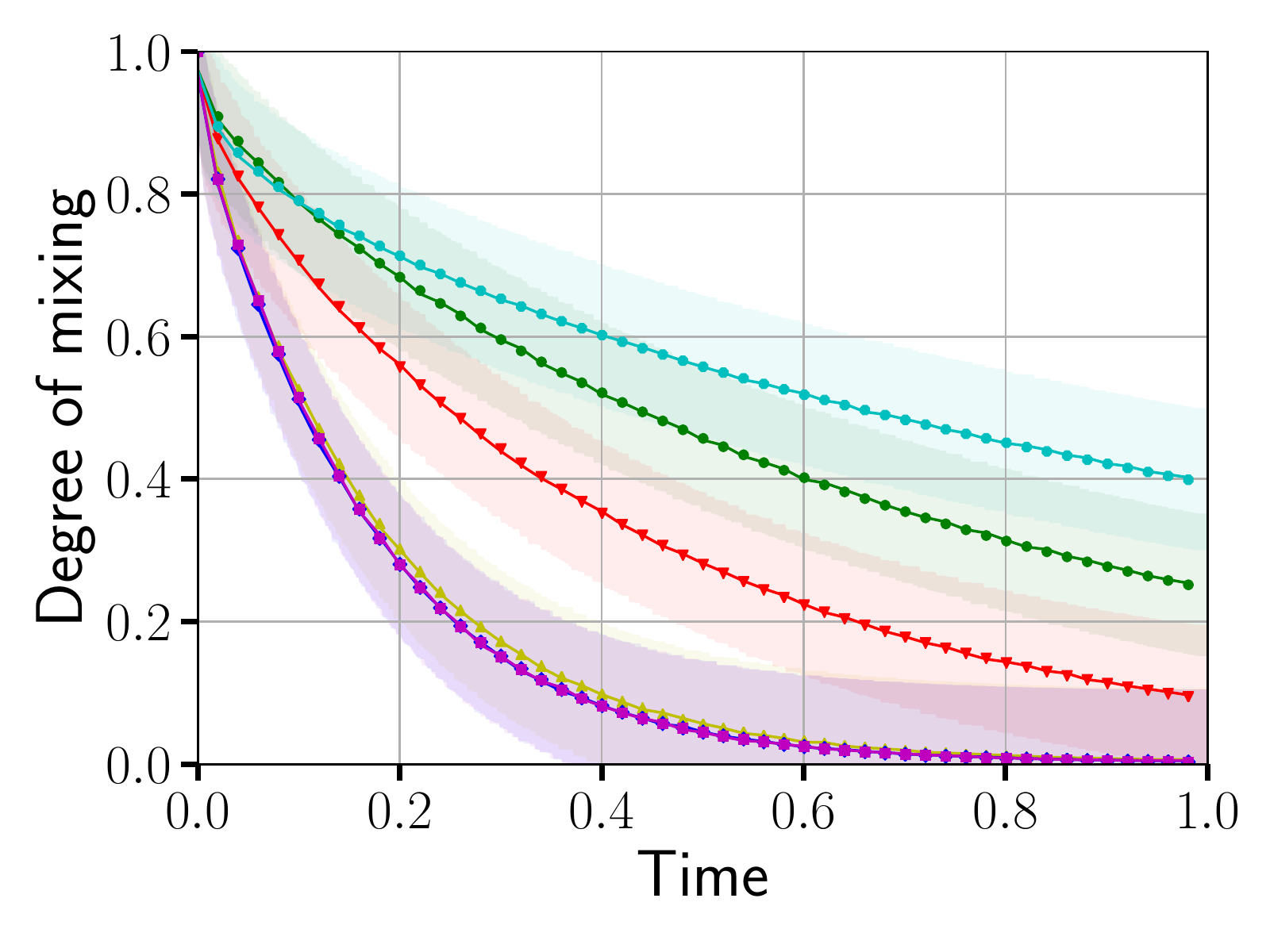}}
  \subfigure[Species $B$:~$\mathfrak{c}_B$]
    {\includegraphics[clip=true,width = 0.3\textwidth]
    {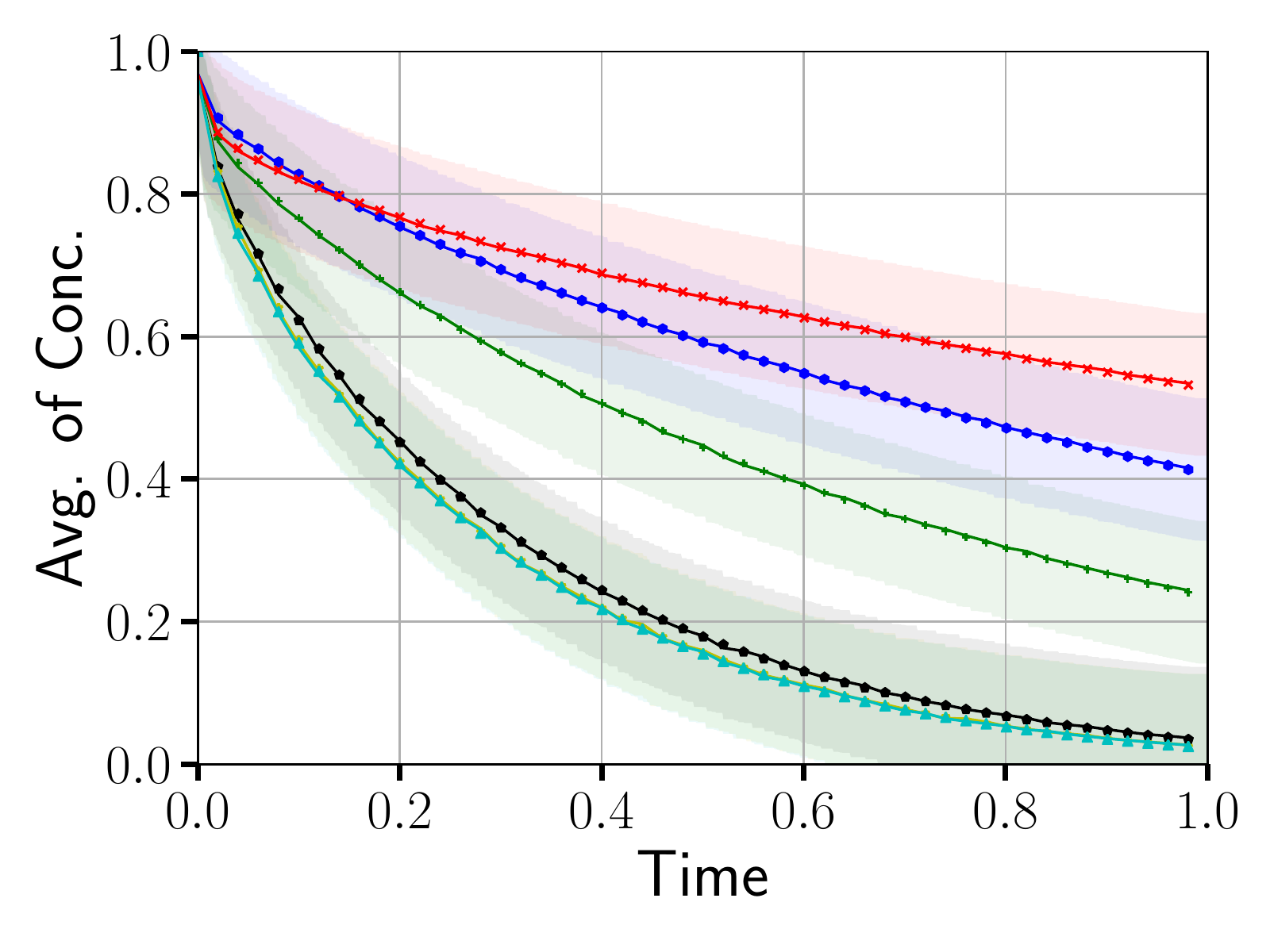}}
  \hspace{-0.1in}
  \subfigure[Species $B$:~$\mathbb{c}_B$]
    {\includegraphics[clip=true,width = 0.3\textwidth]
    {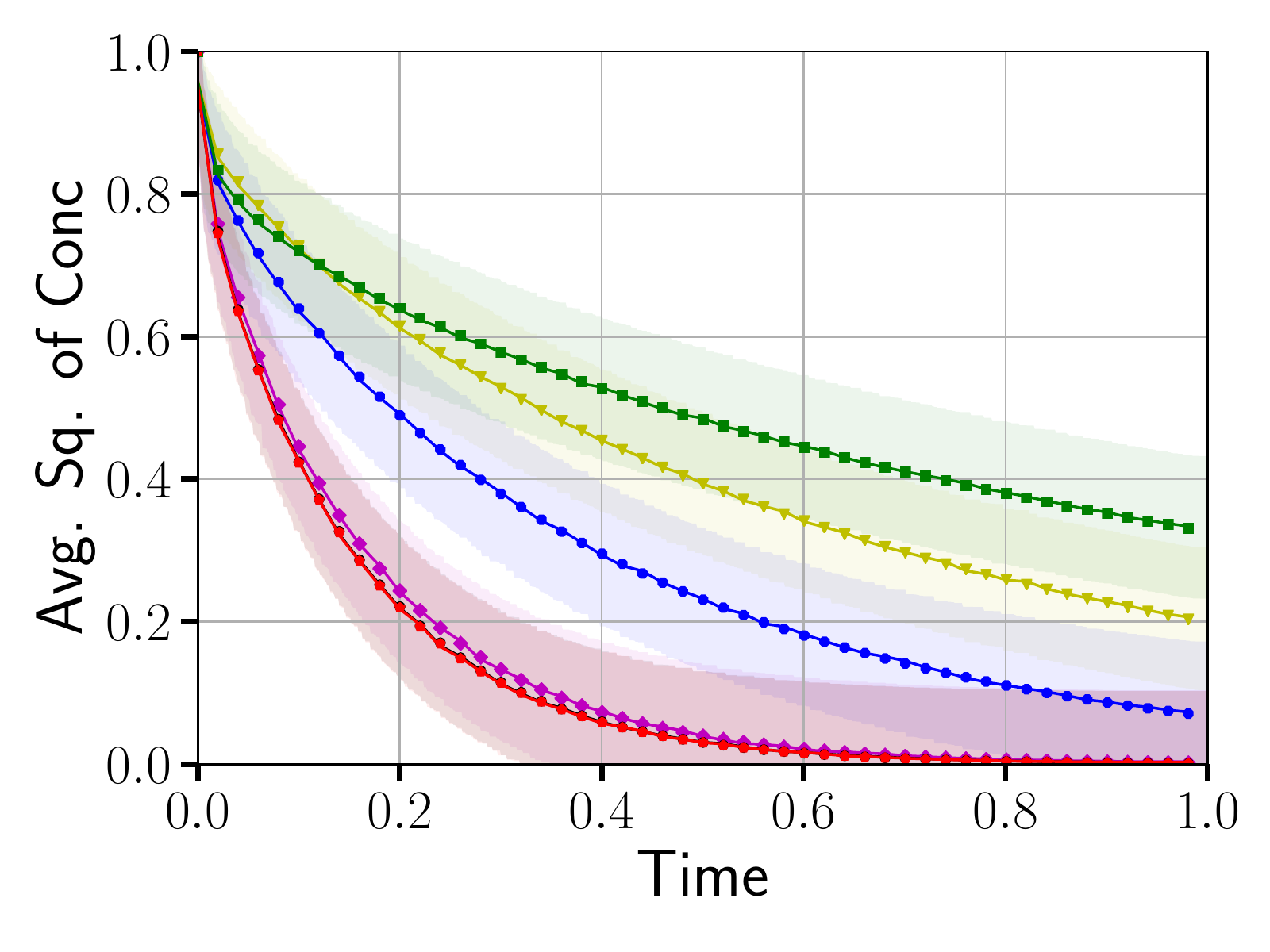}}
  \hspace{-0.1in}
  \subfigure[Species $B$:~$\sigma^2_B$]
    {\includegraphics[clip=true,width = 0.3\textwidth]
    {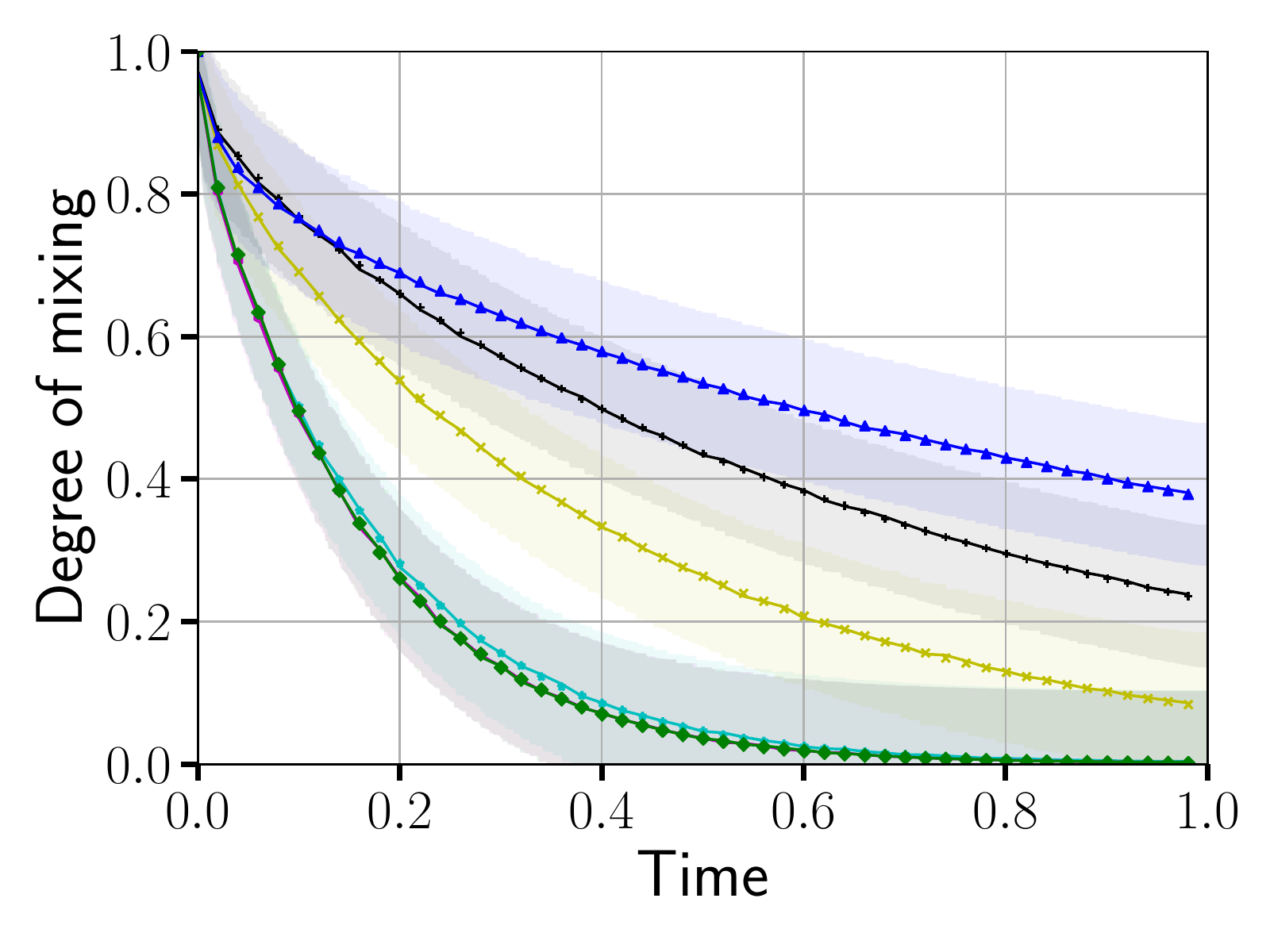}}
  \subfigure[Species $C$:~$\mathfrak{c}_C$]
    {\includegraphics[clip=true,width = 0.3\textwidth]
    {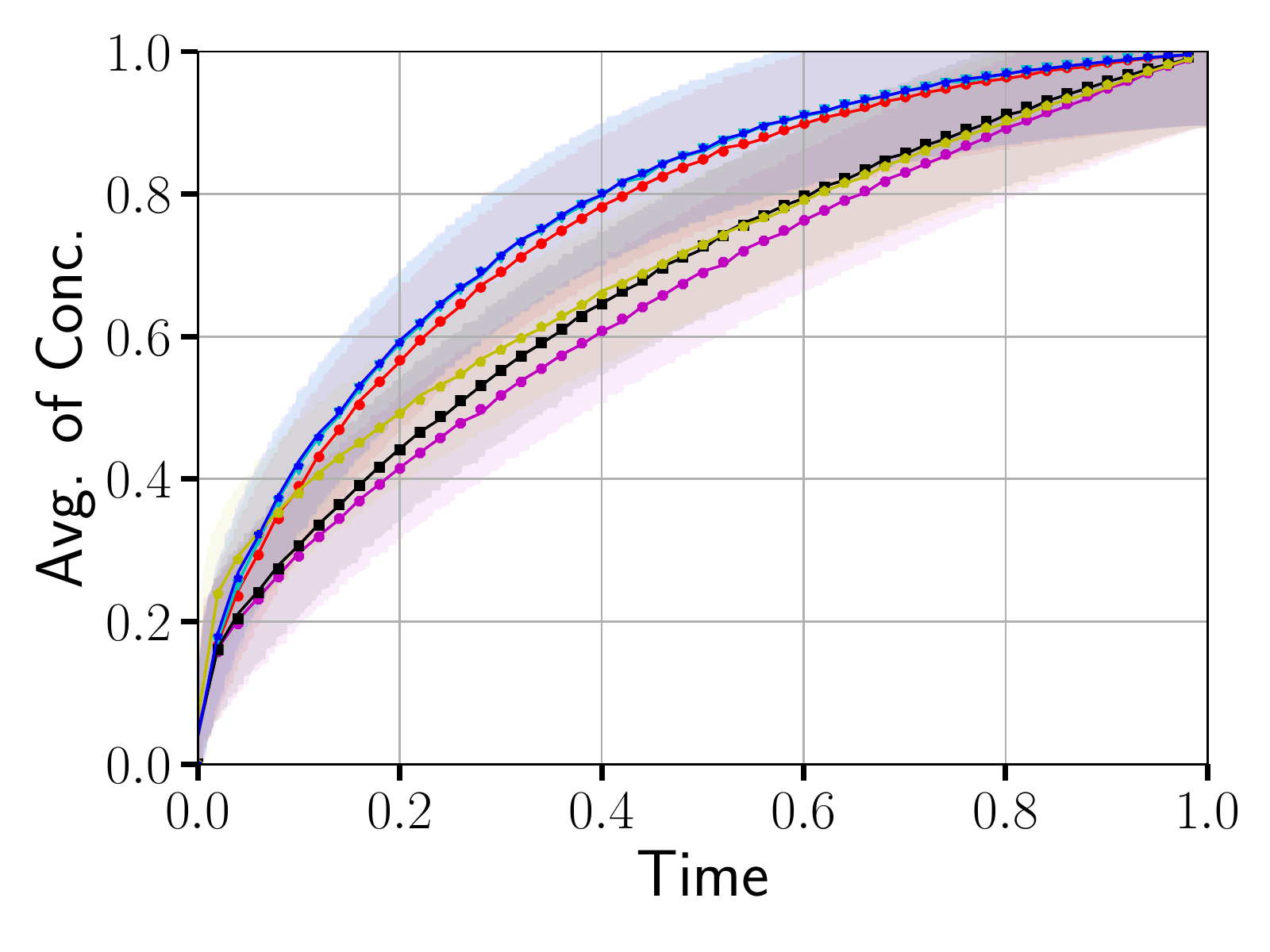}}
  \hspace{-0.1in}
  \subfigure[Species $C$:~$\mathbb{c}_C$]
    {\includegraphics[clip=true,width = 0.3\textwidth]
    {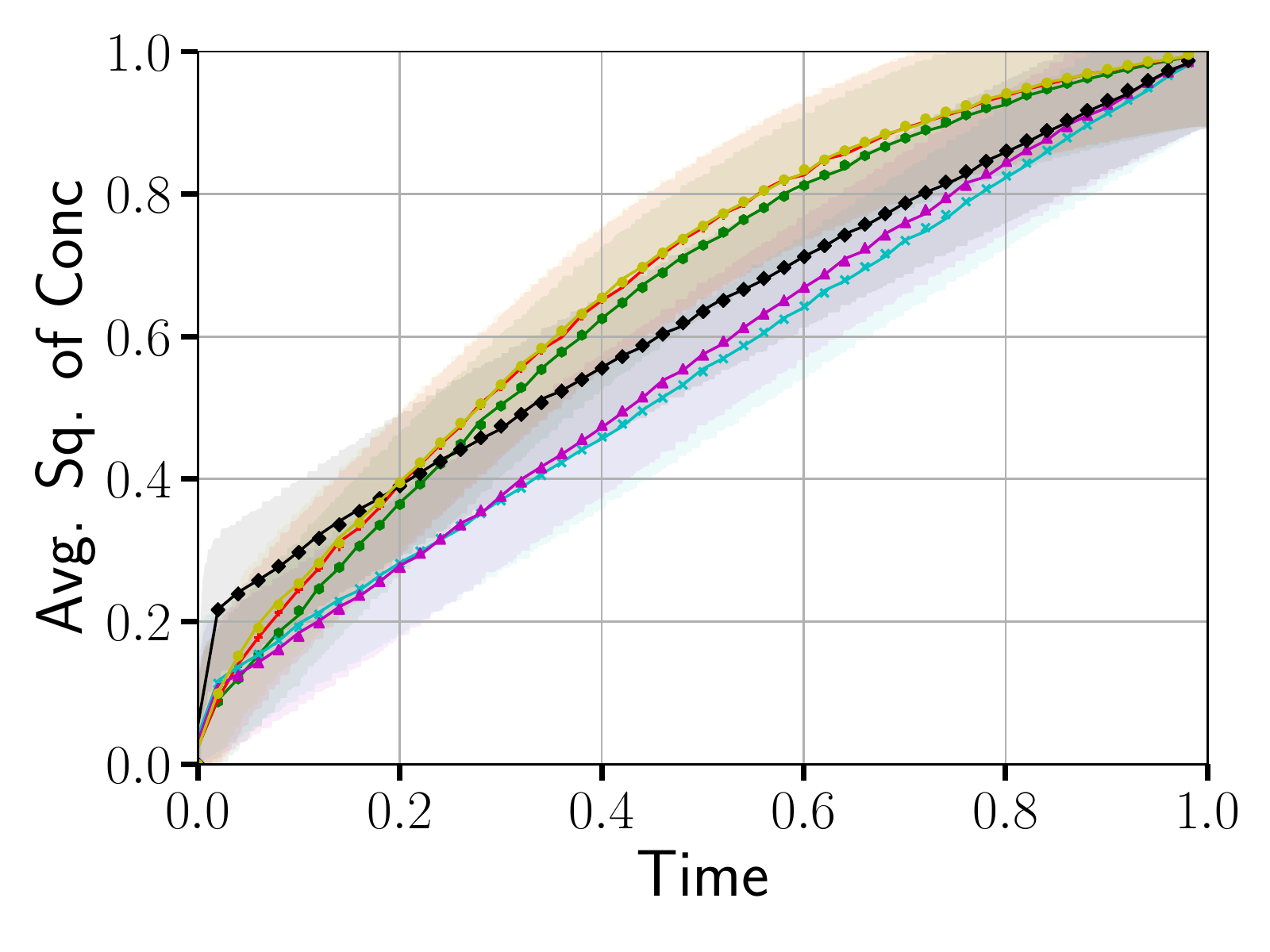}}
  \hspace{-0.1in}
  \subfigure[Species $C$:~$\sigma^2_C$]
    {\includegraphics[clip=true,width = 0.3\textwidth]
    {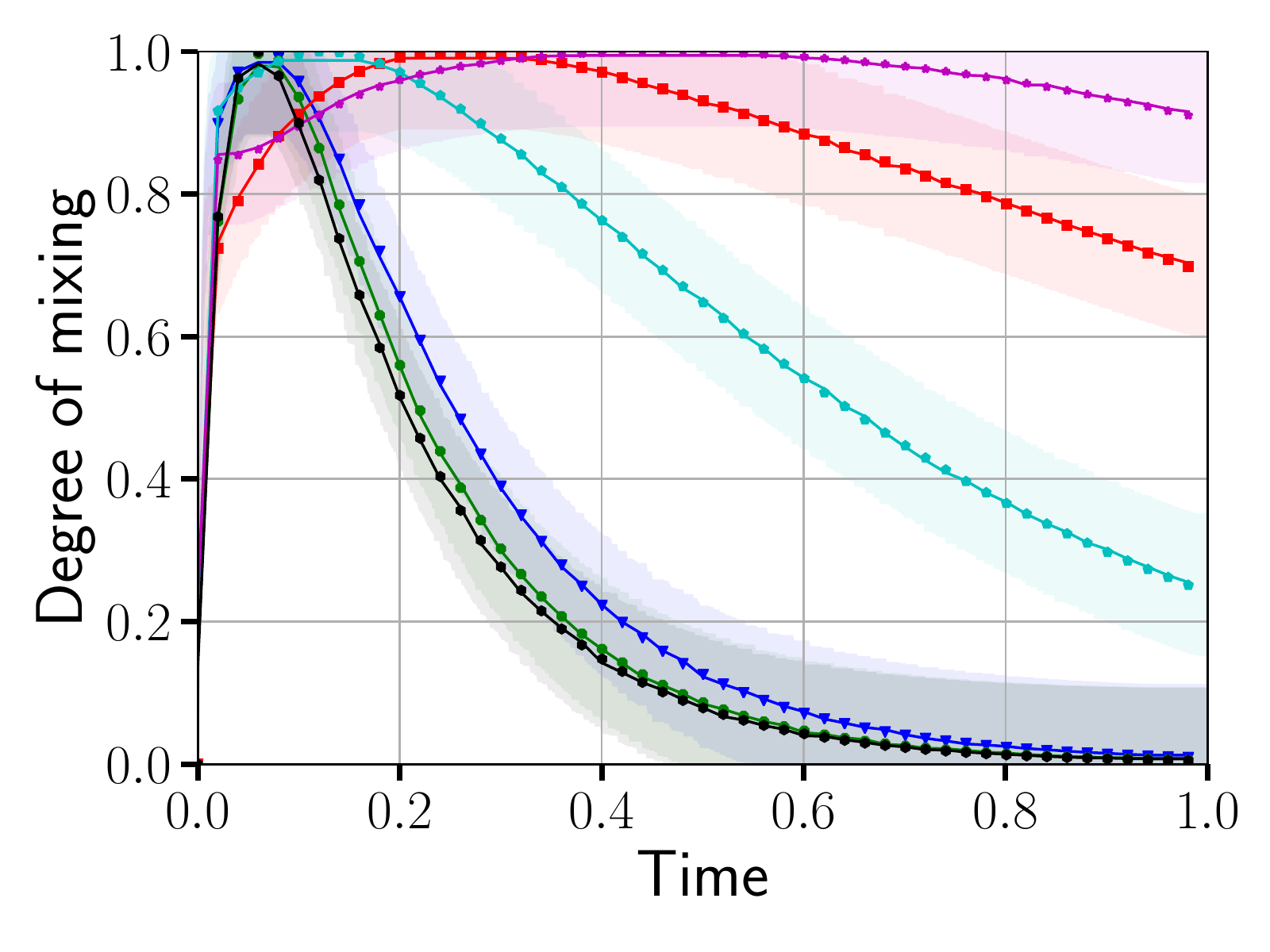}}
  \caption{\textrm{\textbf{Predictions of QoIs by GBM emulators for the six unseen realizations:}}~This figure shows the true (markers) and GBM emulator predictions (solid curves) of average concentrations, squared of average concentrations, and degree of mixing (a)--(c) of species $A$; (d)--(f) of species $B$, and (g)--(i) of species $C$.}
  \label{Fig:GBM_predictions}
\end{figure}

\begin{figure}
  \centering
  \subfigure[Species-$A$:~$\mathfrak{c}_A$]
    {\includegraphics[clip=true,width = 0.3\textwidth]
    {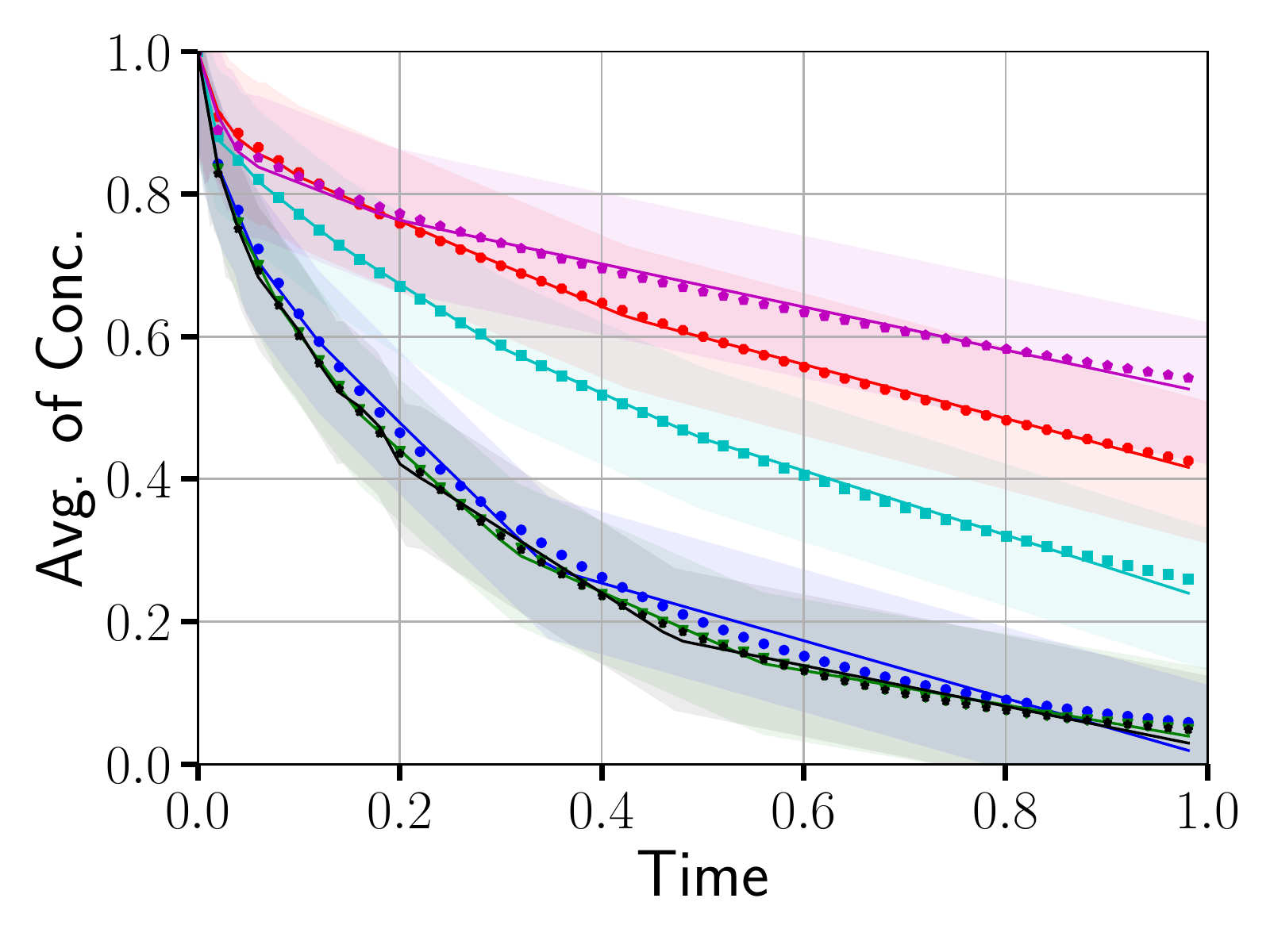}}
  \hspace{-0.1in}
  \subfigure[Species-$A$:~$\mathbb{c}_A$]
    {\includegraphics[clip=true,width = 0.3\textwidth]
    {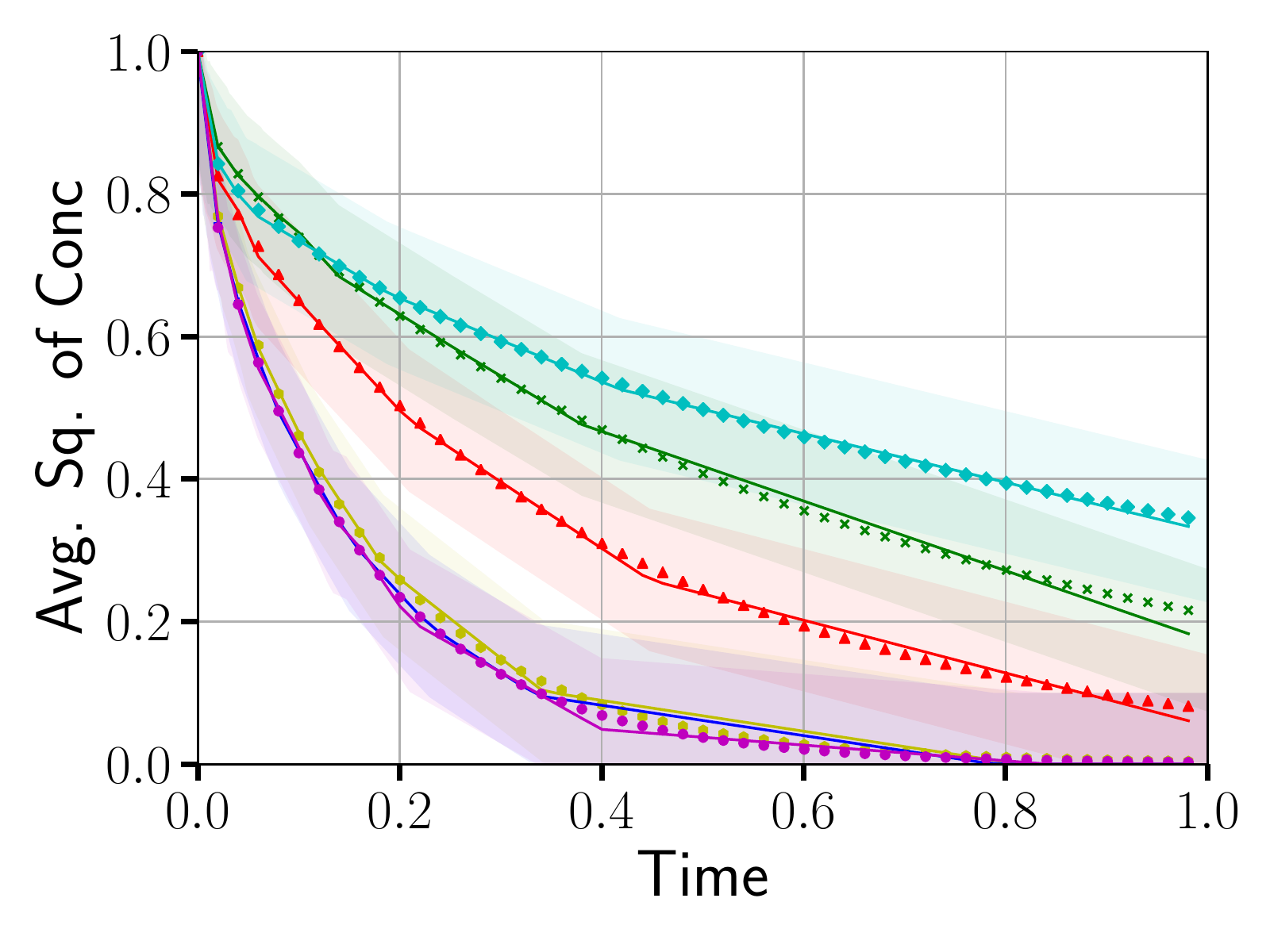}}
  \hspace{-0.1in}
  \subfigure[Species-$A$:~$\sigma^2_A$]
    {\includegraphics[clip=true,width = 0.3\textwidth]
    {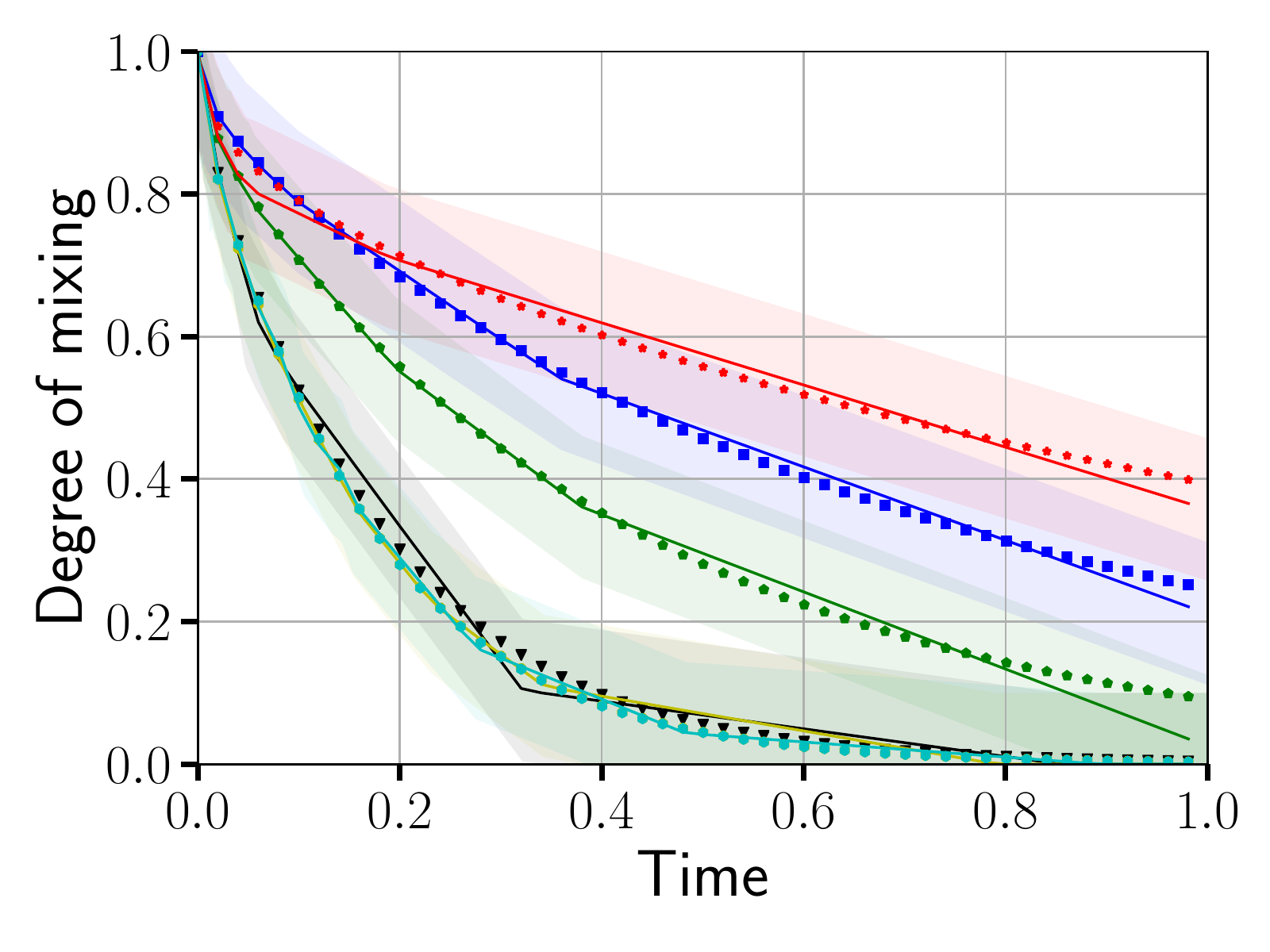}}
  \subfigure[Species-$B$:~$\mathfrak{c}_B$]
    {\includegraphics[clip=true,width = 0.3\textwidth]
    {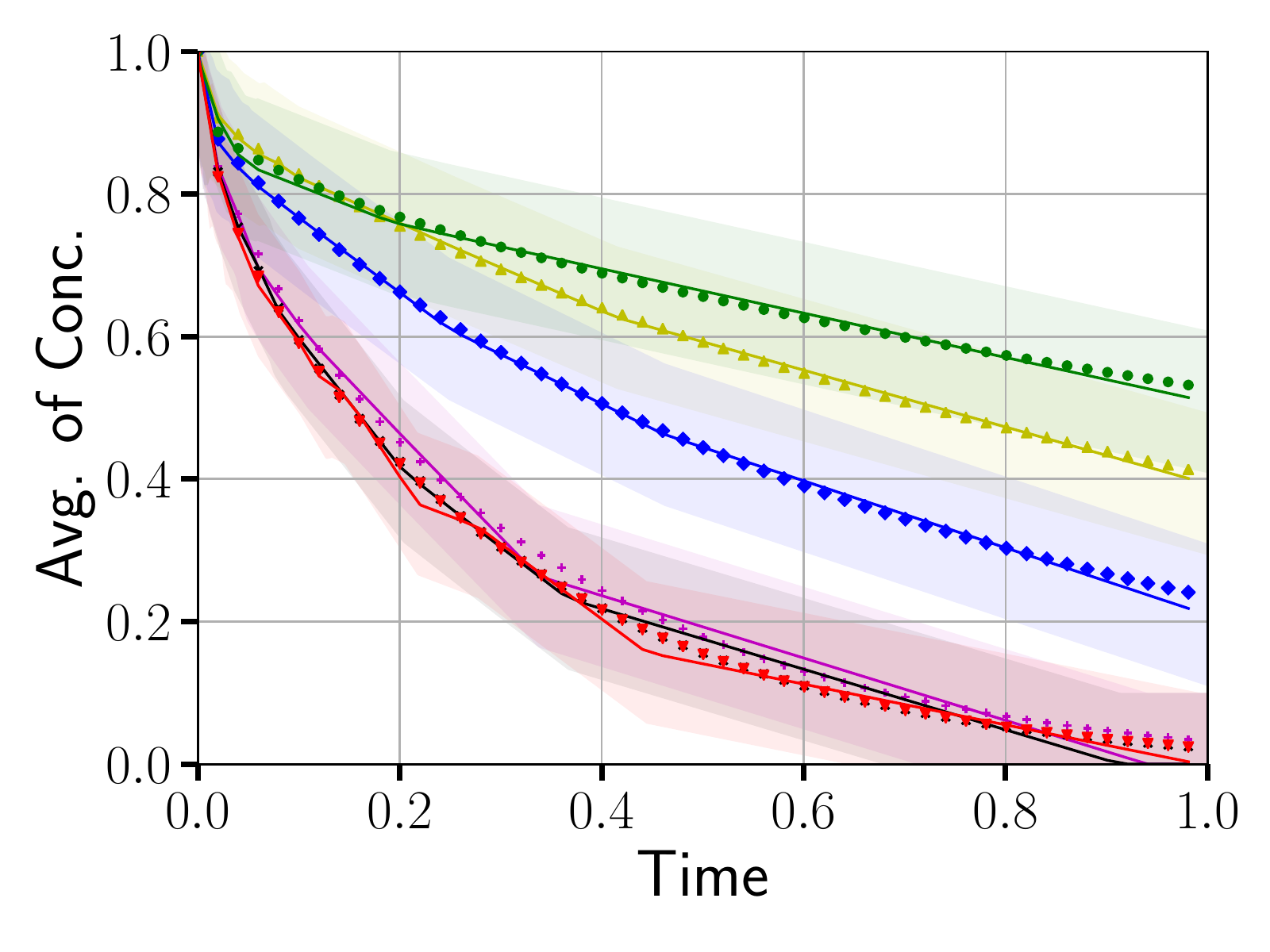}}
  \hspace{-0.1in}
  \subfigure[Species-$B$:~$\mathbb{c}_B$]
    {\includegraphics[clip=true,width = 0.3\textwidth]
    {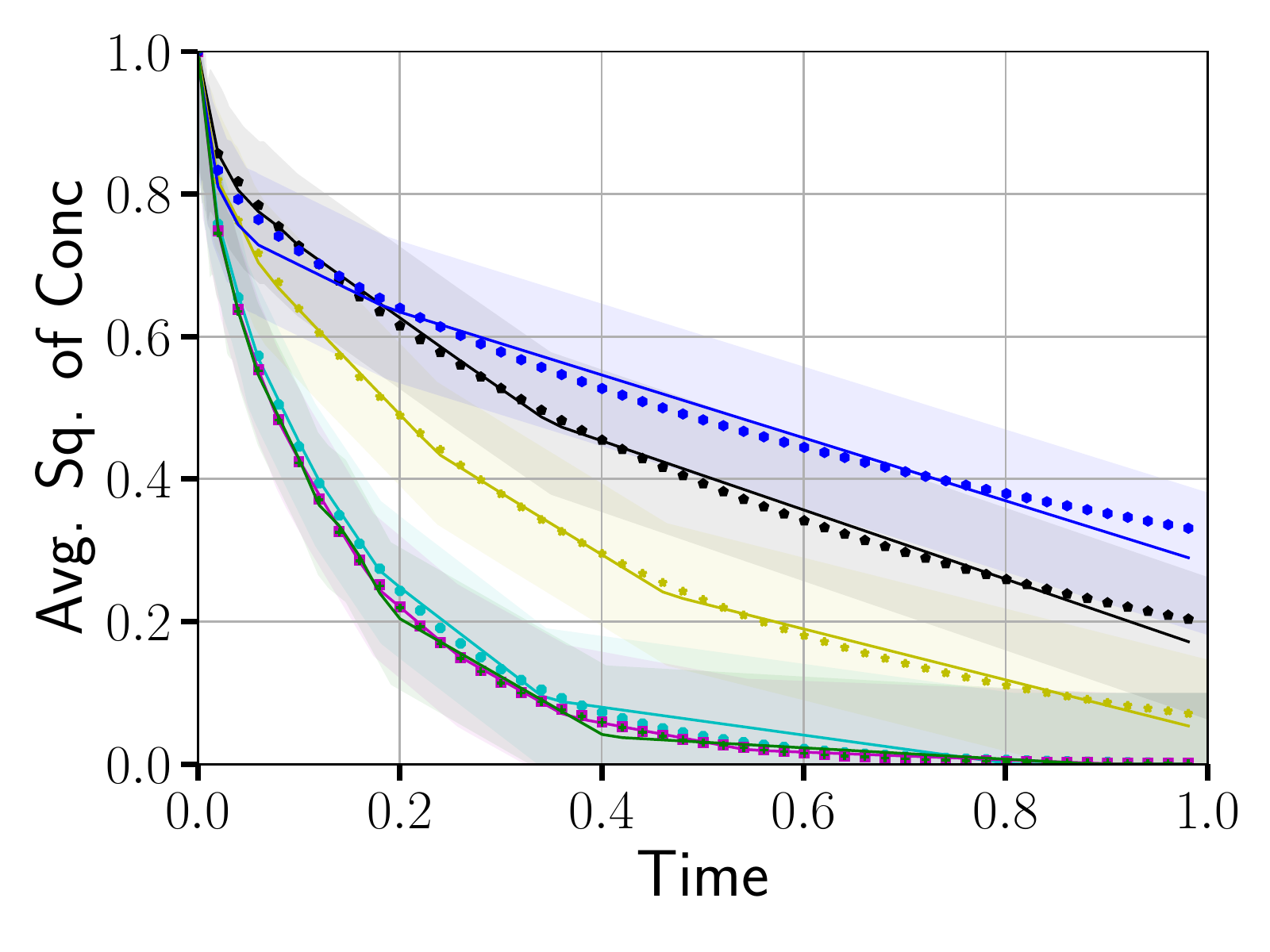}}
  \hspace{-0.1in}
  \subfigure[Species-$B$:~$\sigma^2_B$]
    {\includegraphics[clip=true,width = 0.3\textwidth]
    {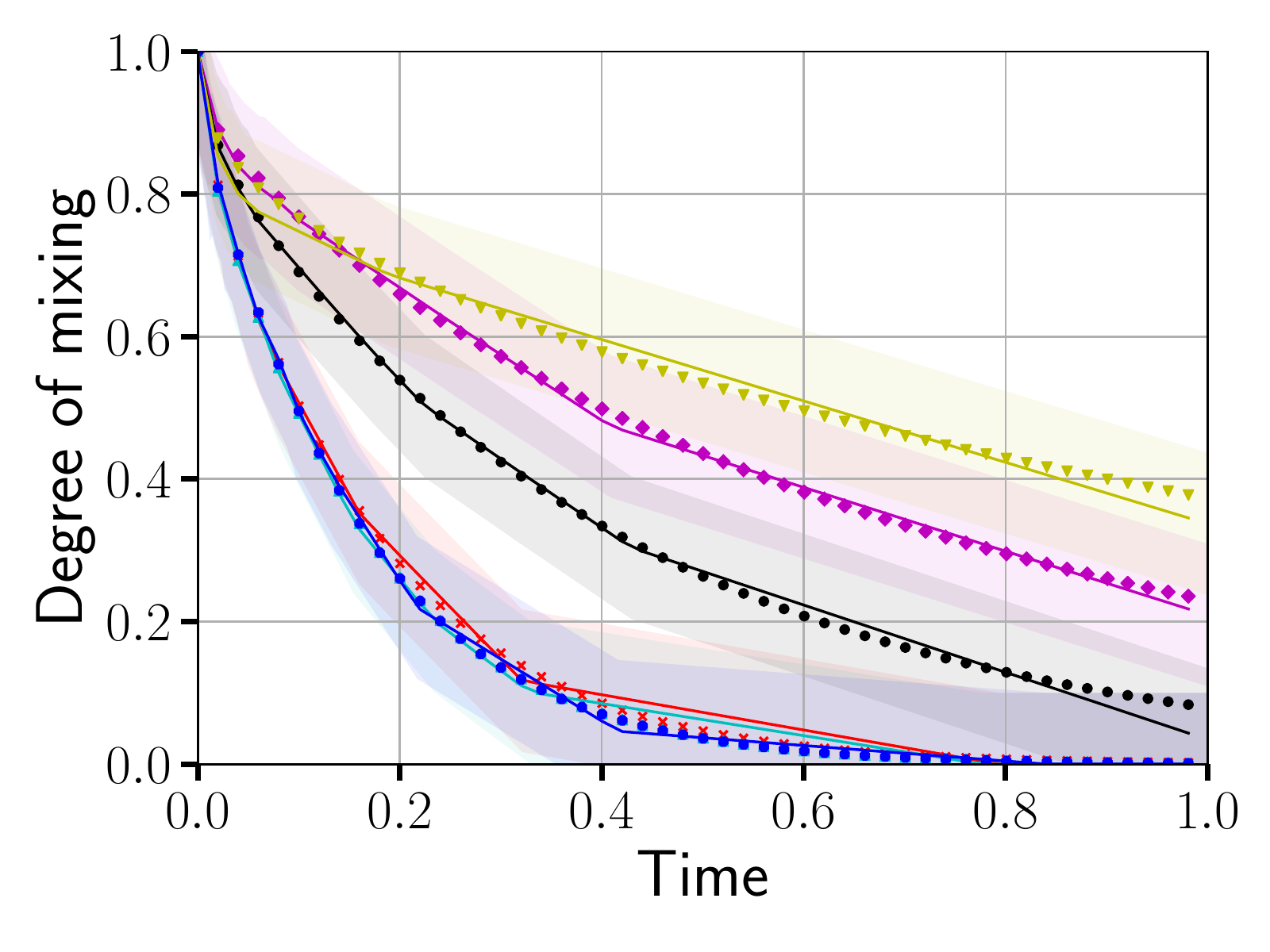}}
  \subfigure[Species-$C$:~$\mathfrak{c}_C$]
    {\includegraphics[clip=true,width = 0.3\textwidth]
    {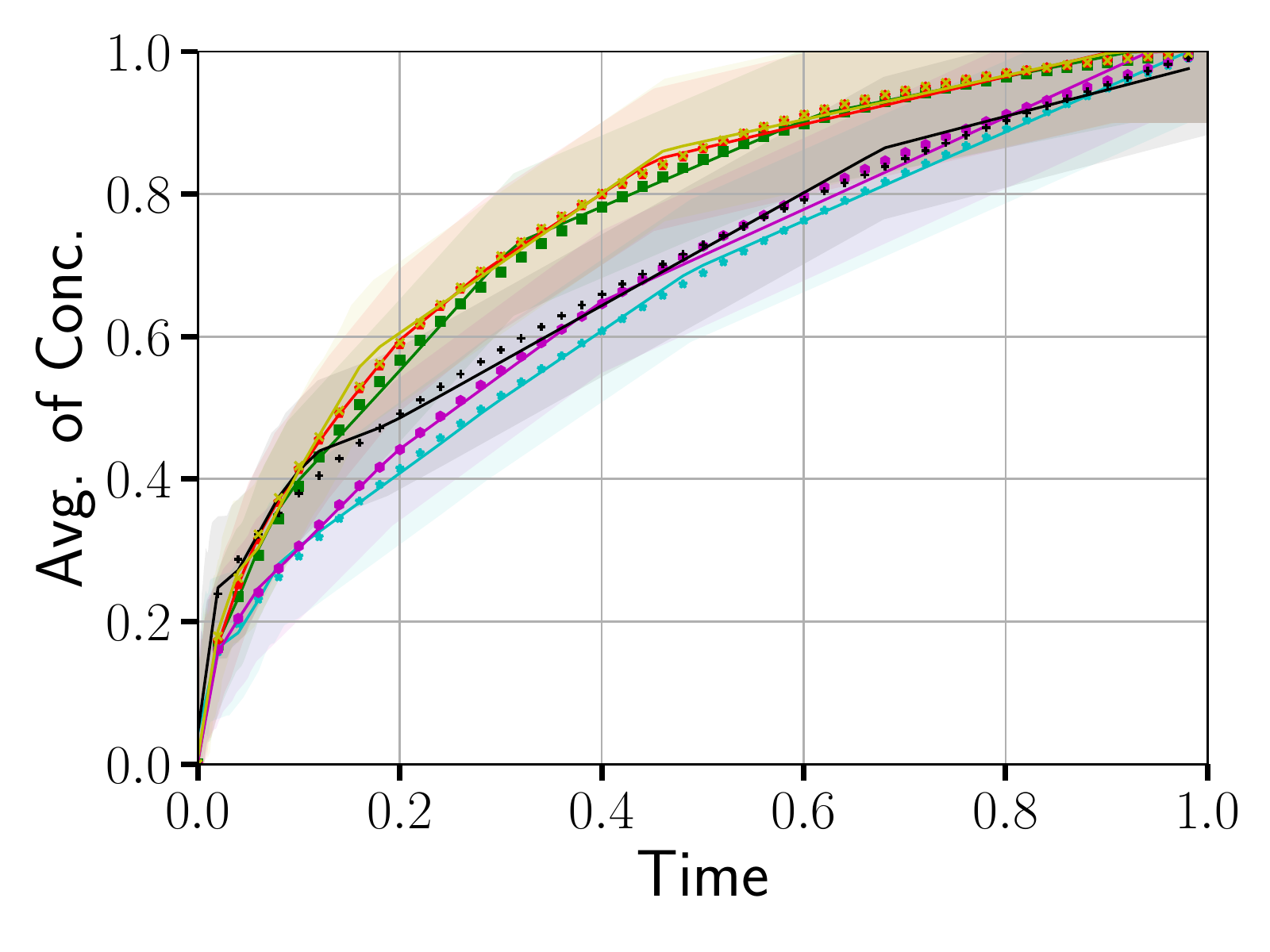}}
  \hspace{-0.1in}
  \subfigure[Species-$C$:~$\mathbb{c}_C$]
    {\includegraphics[clip=true,width = 0.3\textwidth]
    {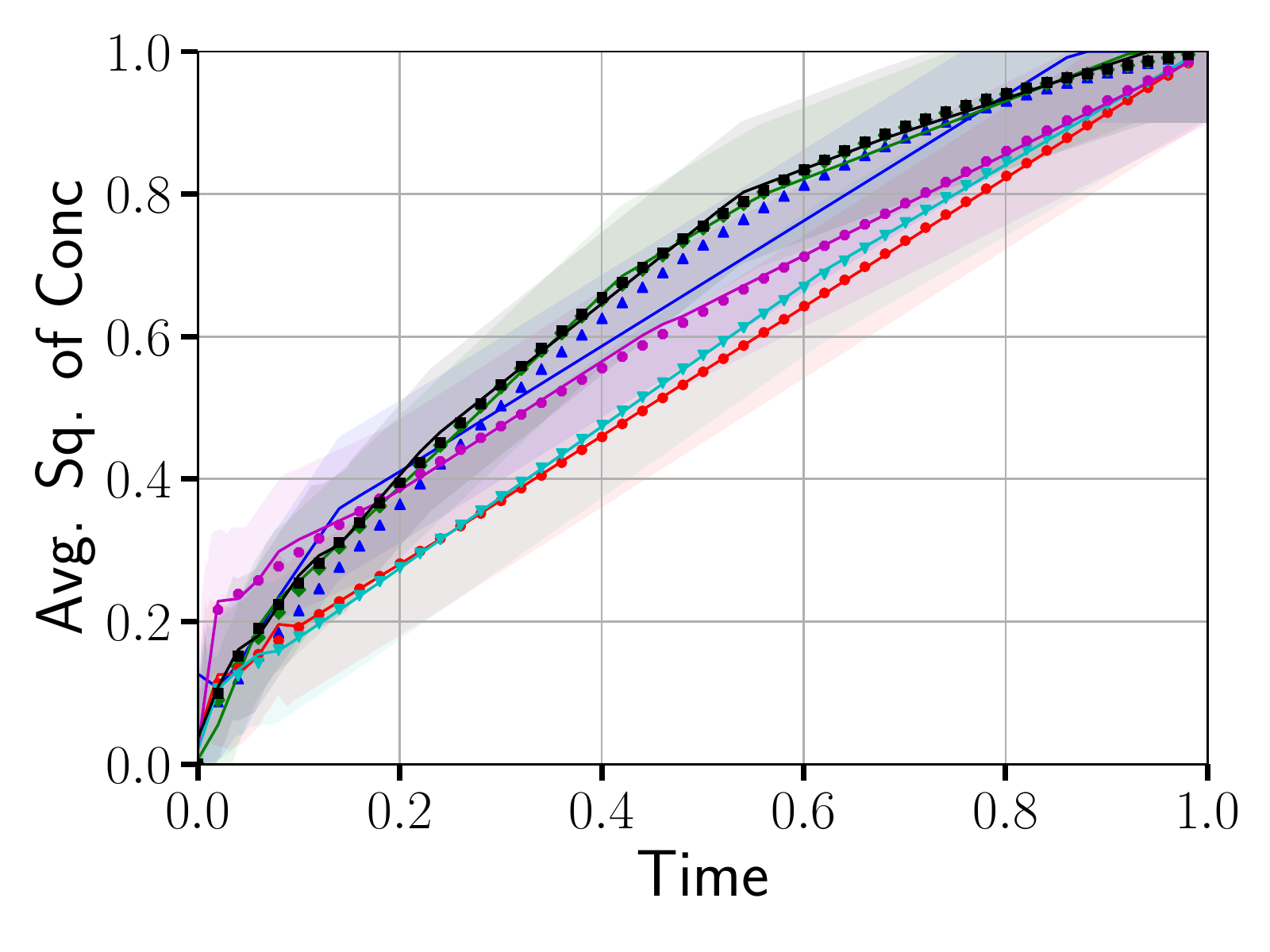}}
  \hspace{-0.1in}
  \subfigure[Species-$C$:~$\sigma^2_C$]
    {\includegraphics[clip=true,width = 0.3\textwidth]
    {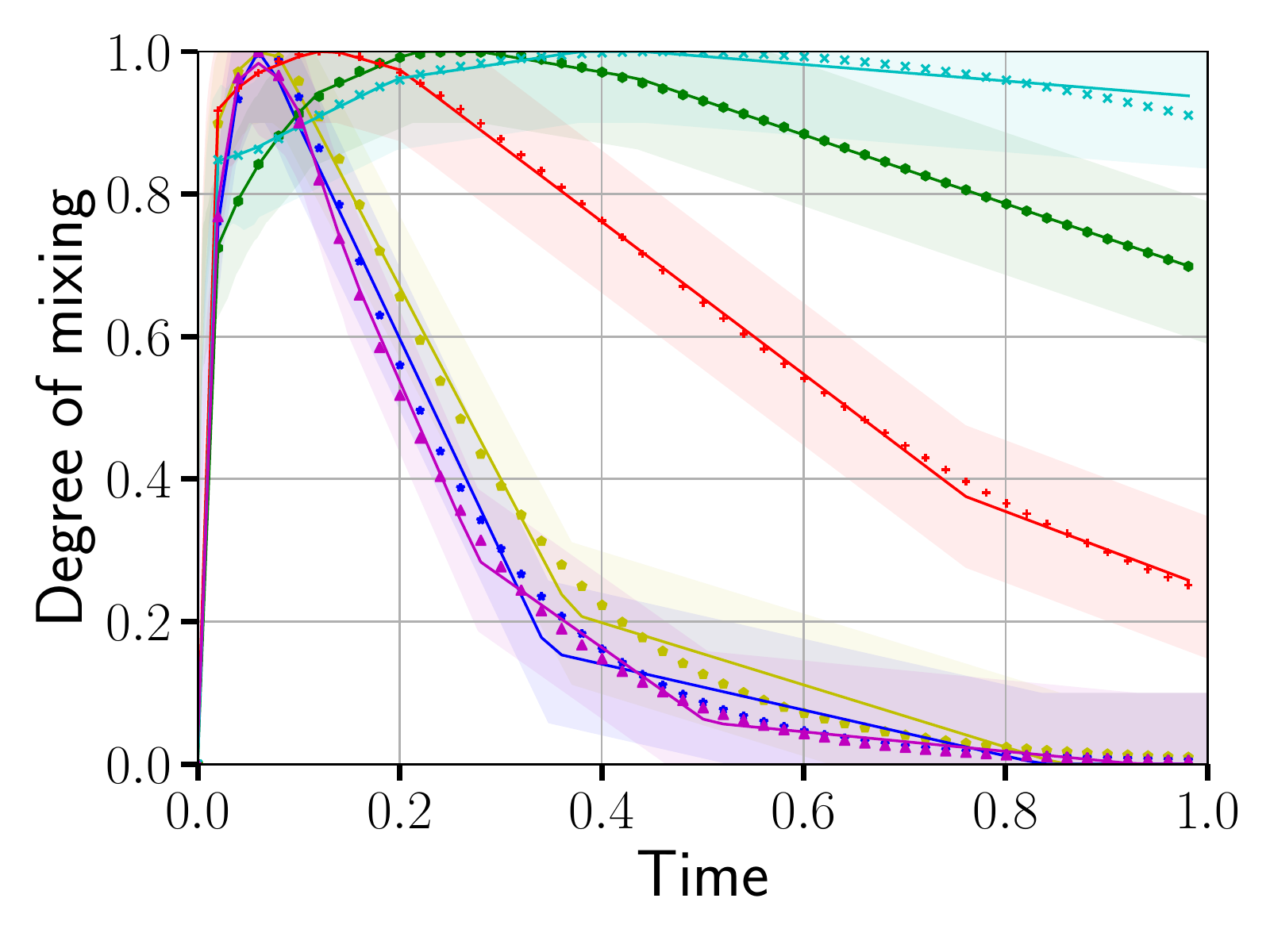}}
  \caption{\textrm{\textbf{Predictions of QoIs by ANN emulators for the six unseen realizations:}}~This figure shows the true (markers) and ANN emulator predictions (solid curves) of average concentrations, squared of average concentrations, and degree of mixing (a)--(c) of species $A$; (d)--(f) of species $B$, and (g)--(i) of species $C$.}
  \label{Fig:ANN_predictions}
\end{figure}

Table~\ref{table:ensemble_regressor_performance} compares the training and testing scores for ensemble and MLP emulators.
The ${R}^2$ scores for training and testing datasets are greater than 90\% (e.g., Bagging, DT, RF, MLP).
For six unseen (blind) realizations, Bagging, DT, RF, AdaBoost, DT-based AdaB, and GBM show astounding match between true QoIs and their corresponding predictions by RF and GBM. 
Here, only figures for RF and GBM emulators (see, Figures~\ref{Fig:RF_predictions}--\ref{Fig:GBM_predictions}) are shown here because the remaining ensemble emulators show the similar trend.
These results indicate that tree-based methods outperform linear ML methods in capturing the QoIs of the reactive-mixing system.
Also, Figure~\ref{Fig:ANN_predictions} shows the QoIs predictions by the MLP emulator for the six blind realizations.
The test ${R}^2$ score (>99\%) on different data sizes and generalized cross-validation during emulator development indicate that overfitting is not a problem.
As the size of the training dataset increases, the ensemble and MLP emulator development time increase.

Finally, the computational costs to run the high-fidelity model and the ML emulators are investigated.
Tables~\ref{table:Lin_classifiers_performance}--\ref{table:ensemble_regressor_performance} compare the computational cost of development of various ML emulators. 
These tables provide details on training time for various training dataset sizes on a 32-core processor (Intel(R) Xeon(R) CPU E5-2695 v4 \@ 2.10GHz).
A single, high-fidelity numerical simulation requires approximately 1,500\,s on a single core.
Testing an ML emulator (e.g., RF, MLP) takes 0.01--0.1\,s about 1/100,000$^{\mathrm{th}}$ of the time of the high-fidelity numerical simulation.

{\small{
\begin{table}[htbp]
    \centering
    \caption{Performance metrics of ML emulators on training and test datasets for classifying the mixing state (i.e., degree of mixing) of the reaction-diffusion system.}
    \begin{tabular}{|clllll|}\hline
      \textbf{Emulator} & \textbf{Training}    & \textbf{Testing}   & \textbf{Training}   & \textbf{Testing}    & \textbf{Training} \\ 
                       & \textbf{size (\%)}    & \textbf{size (\%)} & \textbf{score (\%)} & \textbf{score (\%)} & \textbf{time (s)} \\ \hline \hline
      \multirow{3}*{\minitab[c]{LogR}}  & 0.9  &  99 & 75 & 75  & 31 \\
        & 63        & 30      & 75   & 75  &  138 \\
        & 81        & 10      & 75   & 75  &  174 \\ \hline 
      \multirow{3}*{\minitab[c]{LDA}}  & 0.9 & 99 & 72 & 72  & 28 \\
        & 63      & 30        & 72    & 72 &  93 \\
        & 81      & 10        & 72    & 72 &  102 \\ \hline 
      \multirow{3}*{\minitab[c]{QDA}}  & 0.9  & 99 & 77 & 77  & 66 \\
        & 63       & 30        & 77    & 77 &  128 \\
        & 81       & 10        & 77    & 77 &  133 \\ \hline 
      \multirow{3}*{\minitab[c]{RF}}  & 0.9  & 99 & 100 & 98  &  6,527 \\
        & 63    & 30          & 100    & 99 &  22,161 \\
        & 81    & 10          & 100    & 99 &  24,015 \\ \hline
      \multirow{3}*{\minitab[c]{MLP}}  & 0.9  & 99 & 97 & 96  & 3,384 \\
        & 63    & 30          & 99    & 99 &  50,397 \\
        & 81    & 10          & 99    & 99 &  66,381 \\ \hline 
    \end{tabular}
    \label{table:Lin_classifiers_performance}
\end{table}
}}

{\small{
\begin{table}[htbp]
    \centering
    \caption{Performance metrics of linear and Bayesian ML emulators (regressors). Note, GP and KR failed to converge even on 1\% of training data because of a memory leak due to storage of a dense matrix.}
    \begin{tabular}{|clllll|}\hline
      \textbf{Emulator} & \textbf{Training}    & \textbf{Testing}   & \textbf{Training}   & \textbf{Testing}    & \textbf{Training} \\ 
                       & \textbf{size (\%)}    & \textbf{size (\%)} & \textbf{score (\%)} & \textbf{score (\%)} & \textbf{time (s)} \\ \hline \hline
      \multirow{3}*{\minitab[c]{LSQR}} & 0.9  & 99 & 69 & 69   &  12\\
        & 63    & 30          & 69    & 69 &  52 \\
        & 81    & 10          & 69    & 69 &  57 \\ \hline 
      \multirow{3}*{\minitab[c]{RR}} & 0.9  & 99 & 69 & 69   &  10\\
        & 63    & 30          & 69    & 69 &  42 \\
        & 81    & 10          & 69    & 69 &  50 \\ \hline 
      \multirow{3}*{\minitab[c]{LR}} & 0.9  & 99 & 69 & 69   &  95 \\
        & 63    & 30          & 69    & 69 & 330  \\
        & 81    & 10          & 69    & 69 &  368 \\ \hline 
      \multirow{3}*{\minitab[c]{ER}} & 0.9  & 99 & 69 & 69   &  121\\
        & 63    & 30          & 69    & 69 &  1,077 \\
        & 81    & 10          & 69    & 69 &  1,227 \\ \hline 
      \multirow{3}*{\minitab[c]{HR}}  & 0.9  & 99  & 69 & 69  & 14 \\
        & 63    & 30          & 69 & 69  &  185 \\
        & 81    & 10          & 69 & 69  &  195 \\ \hline
      \multirow{3}*{\minitab[c]{Polynomial}}  & 0.9  & 99 & 89 & 89  & 79 \\
        & 63    & 30          & 89    & 99 &  143 \\
        & 81     & 10          & 89    & 89  & 164 \\ \hline 
      \multirow{3}*{\minitab[c]{BR}} & 0.9  & 99 & 69 & 69   &  12\\
        & 63    & 30          & 69    & 69 &  62 \\
        & 81    & 10          & 69    & 69 &  69 \\ \hline 
      \multirow{3}*{\minitab[c]{NB}}  & 0.9  & 99 & 73 & 73  & 69 \\
        & 63    & 30          & 73 & 73  &  73 \\
        & 81    & 10          & 73 & 73  &  91 \\ \hline 
    \end{tabular}
    \label{table:Lin_regressor_performance}
\end{table}
}}

{\small{
\begin{table}[htbp]
    \centering
    \caption{Performance metrics of ensemble and MLP emulators.}
    \begin{tabular}{|clllll|}\hline
      \textbf{Emulator} & \textbf{Training}    & \textbf{Testing}   & \textbf{Training}   & \textbf{Testing}    & \textbf{Training} \\ 
                       & \textbf{size (\%)}    & \textbf{size (\%)} & \textbf{score (\%)} & \textbf{score (\%)} & \textbf{time (s)} \\ \hline \hline
      \multirow{3}*{\minitab[c]{DT}} & 0.9  & 99 & 100 & 99   &  42 \\
        & 63    & 30          & 99    & 99   &  100 \\
        & 81    & 10          & 99    & 99 &   110 \\ \hline 
      \multirow{3}{*}{Bagging} & 0.9  & 99 & 98 & 95   &  42 \\
        & 63    & 30          & 98    & 95 &  110 \\
        & 81    & 10          & 98    & 95 &   100 \\ \hline
      \multirow{3}*{\minitab[c]{RF}} & 0.9  & 99 & 100 & 99   &  1,435\\
        & 63    & 30          & 100    & 99 &  5,468\\
        & 81    & 10          & 100    & 99 &  6,044\\ \hline 
      \multirow{3}{*}{AdaB} & 0.9  & 99 & 90 & 90  &  72\\
                                & 63 & 30 & 89 & 89  &  1,378\\
                                & 81 & 10 & 89 & 89  &  1,585\\ \hline 
      \multirow{3}*{\minitab[c]{DT-based \\ AdaB}} & 0.9  & 99 & 99 & 99  & 103 \\
        & 63    & 30          & 99    & 99 &  1,648 \\
        & 81    & 10          & 99    & 99 &  1,778 \\ \hline 
      \multirow{3}*{\minitab[c]{GBM}} & 0.9  & 99 & 98 & 98  &  133\\
        & 63    & 30          & 98    & 98 &   1,533 \\
        & 81    & 10          & 98    & 98   &  2,048 \\ \hline 
      \multirow{3}*{\minitab[c]{MLP}}  & 0.9  & 99 & 99 & 99  & 688 \\
        & 63    & 30          & 99    & 99 &  4,678 \\
        & 81    & 10          & 99    & 99 &  9,691 \\ \hline 
    \end{tabular}
    \label{table:ensemble_regressor_performance}
\end{table}
}}

\section{Discussion}
\label{Sec:S5_Discussion}
A suite of linear, Bayesian, and nonlinear ML emulators are trained to classify and replicate QoIs from high-fidelity anisotropic bi-linear diffusion numerical simulations.
For this highly nonlinear system, linear and Bayesian ML emulators never exceed 70\% classification accuracy while LogR and QDA achieve only 75\% and 77\% classification accuracies, respectively.
On the other hand, nonlinear emulators perform well (95\% classification accuracies for RF and MLP).
For the regression problem (predicting the three QoIs for each chemical species), as expected, linear regressors predict QoIs at only $R^2 = 69$\%, but decision-tree-based ensembles and the MLP neural network perform remarkably well.
DTs (with and without AdaBoost), RFs, and the MLP all had $R^2 = 99$\% with GBM (98\%), bagging (95\%), and AdaBoost (85\%) performs somewhat worse. 

These results indicate that ensemble emulators outperform other ML emulators in predicting the progress of reactive mixing on unseen data.
However, not all of them perform equally.
For example, RF outperforms other averaging ensembles (e.g., Bagging, DT) while DT-based AdaB outperforms other boosting methods (e.g., AdaB, GBM).
Each bagging/averaging ensemble methods introduce randomness and voting-based evaluation metrics in unique ways; therefore, their performance is not the same.
For example, DTs often use the first feature to split; resultantly, the order of variables in the training data is critical for DT-based model construction.
Also, in DTs, trees are pruned and not fully grown.
Contrarily, RF can have unpruned and fully grown trees and are not sensitive to the feature order as in DTs.
Also, each tree in an RF learns using random sampling, and at each node, a random set of features are considered for splitting.
This random sampling and splitting introduces diversity among trees in a forest.
After randomly selecting features, RF builds a number of regression trees and averages (aka bagging) them.
With enough trees, combinations of randomly selected features and averaging (aka voting), RF emulators reduce the variance of predictions and deter the overfitting.
Resultantly, their performances are best among all averaging ensemble emulators.

Among boosting methods, DT-based AdaB outperforms AdaB and GBM because it combines DT and boosting estimators to predict QoIs.
In this study, the DT-based AdaB uses 100 trees as a base estimator to build DT-based AdaB emulator.
Two base estimators enhance the confidence on QoI predictions; resultantly, the DT-based AdaB emulator scores better than other two boosting approaches.  
Based on the ML analyses presented in Sec.~\ref{Sec:S4_Results}, linear and Bayesian ML emulators (e.g., NB, BR, GP) are a poor choice to classify and predict reactive-mixing QoIs.
Overall, RF, DT-based AdaB, GBM, and MLP emulators accurately predicted unseen realizations with average accuracies >90\%.
From the computational-cost perspective, generalized linear and Bayesian ML emulators are faster to train than ensemble and MLP emulators.
Among ensemble and boosting methods, RF and GBM emulators take longest to train.
Also, MLP emulators are more expensive to develop than other ML emulators.
However, ensemble and MLP emulators take 1/100,000$^{\mathrm{th}}$ of the time required for a high-fidelity simulation to predict equivalent QoIs.  

\section{Conclusions}
\label{Sec:S6_Conclusions}
Our primary purpose was to accurately understand reactive-mixing state and expedite predictions of species concentration (QoIs) due to reactive mixing.
A suite of linear, Bayesian, ensemble, and MLP ML emulators were compared to classify the state of reactive mixing and to predict species concentrations.
All ML emulators were developed based on high-fidelity numerical simulation datasets.
A~total of 2,315 simulations were carried out to generate data to train and test the emulators.
Data were generated by solving the anisotropic reaction-diffusion equations using the non-negative finite element method.
Because of the highly nonlinear reactive-mixing system, linear and Bayesian (except GP) ML emulators performed poorly in classifying and predicting the state of reactive mixing (e.g., $R^2$ $\approx$ 70\%).
Among Bayesian ML emulators, GP showed promise for accurate prediction of QoIs for small datasets.
On the other hand, ensemble and MLP emulators accurately classified the state of reactive-mixing and predicted associated QoIs.
For example, RF and MLP emulators classified the state of reactive-mixing with an accuracy of \textgreater90\%.
Moreover, they predicted the progress of reactive-mixing with an accuracy of \textgreater95\% on training, testing, and unseen data.
Among bagging ensemble methods, RF emulators provided comparatively better predictions than bagging and DT emulators.
Similarly, among boosting ensemble methods, DT-based AdaBoost emulators provided better predictions than AdaBoost and GBM emulators.
Computationally, for QoI predictions, ML emulators were approximately $10^{5}$ faster than a high-fidelity numerical simulation.
Finally, ensemble ML and MLP emulators proved good classifiers and predictors for interrogating the progress of reactive mixing.
Looking to the future, ensemble ML and MLP emulators will be validated on both reservoir-scale field and simulation data.

\section*{Acknowledgments}
Bulbul Ahmmed thanks the support from Mickey Leland Energy Fellowship (MLEF) awarded by U.S. Department of Energy's (DOE) Office of Fossil Energy (FE).
MKM and SK also thank the support of the LANL Laboratory Directed Research and Development (LDRD) Early Career Award 20150693ECR. 
VVV thanks the support of LANL LDRD-DR Grant 20190020DR.
Los Alamos National Laboratory is operated by Triad National Security, LLC, for the National Nuclear Security Administration of U.S. Department of Energy (Contract No. 89233218CNA000001).
Additional information regarding the simulation datasets and codes can be obtained from Bulbul Ahmmed (Email:~\texttt{ahmmedb@lanl.gov}) and Maruti Mudunuru (Email: \texttt{maruti@lanl.gov}).
\section*{Conflict of Interest}
The authors declare that they do not have conflict of interest.
\section*{Computer Code Availability}
Codes for machine learning implementation are available in the public Github repository \url{https://github.com/bulbulahmmed/ML-to-reactive-mixing-data}.
Additional information regarding the simulation datasets can be obtained from Bulbul Ahmmed (Email:~\texttt{bulbul\_ahmmed@baylor.edu}) and Maruti Kumar Mudunuru (Email: \texttt{maruti@lanl.gov}).
\bibliographystyle{unsrt}
\bibliography{Master_References/References}

\begin{thebibliography}{10}

\bibitem{R1_Lagneau2019industry}
V.~Lagneau, O.~Regnault, and M.~Descostes.
\newblock Industrial deployment of reactive transport simulation:~{A}n
  application to uranium in situ recovery.
\newblock {\em Reviews in Mineralogy and Geochemistry}, 85:499--528, 09 2019.

\bibitem{R2_Cama2019acid_water}
J.~Cama, J.~M. Soler, and C.~Ayora.
\newblock Acid water-rock-cement interaction and multicomponent reactive
  transport modeling.
\newblock {\em Reviews in Mineralogy and Geochemistry}, 85:459--498, 09 2019.

\bibitem{R3_Rolle2019reactive_fronts}
M.~Rolle and T.~Le~Borgne.
\newblock Mixing and reactive fronts in the subsurface.
\newblock {\em Reviews in Mineralogy and Geochemistry}, 85:111--142, 09 2019.

\bibitem{R4_sin2019_multiphase}
I.~Sin and J.~Corvisier.
\newblock Multiphase multicomponent reactive transport and flow modeling.
\newblock {\em Reviews in Mineralogy and Geochemistry}, 85:143--195, 09 2019.

\bibitem{R5_molins2019_multiscale}
S.~Molins and P.~Knabner.
\newblock Multiscale approaches in reactive transport modeling.
\newblock {\em Reviews in Mineralogy and Geochemistry}, 85:27--48, 09 2019.

\bibitem{lichtner2019reactive}
P.~C. Lichtner, C.~I. Steefel, and E.~H. Oelkers.
\newblock {\em Reactive {T}ransport in {P}orous {M}edia}, volume~34.
\newblock Walter de Gruyter GmbH \& Co KG, 2019.

\bibitem{lichtner2015pflotran}
P.~C. Lichtner, G.~E. Hammond, C.~Lu, S.~Karra, G.~Bisht, B.~Andre, R.~T.
  Mills, and J.~Kumar.
\newblock {PFLOTRAN} user manual: A massively parallel reactive flow and
  transport model for describing surface and subsurface processes.
\newblock Technical report, (Report No.: LA-UR-15-20403) Los Alamos National
  Laboratory, 2015.

\bibitem{chen2018pore}
L.~Chen, M.~Wang, Q.~Kang, and W.~Tao.
\newblock Pore-scale study of multiphase multicomponent reactive transport
  during {CO}\textsubscript{2} dissolution trapping.
\newblock {\em Advances in Water Resources}, 116:208--218, 2018.

\bibitem{ozturk2015plants}
M.~A. {\"O}zt{\"u}rk, M.~Ashraf, A.~Aksoy, M.~S.~A. Ahmad, and K.~R. Hakeem.
\newblock {\em Plants, Pollutants and Remediation}.
\newblock Springer, 2015.

\bibitem{ahmmedMS2015}
B.~Ahmmed.
\newblock Numerical modeling of {CO}$_2$-water-rock interactions in the
  {F}arnsworth, {T}exas hydrocarbon unit, {USA}.
\newblock Master's thesis, University of Missouri, 2015.

\bibitem{VESSELINOV201985ntfk}
V.V. Vesselinov, M.K. Mudunuru, S.~Karra, D.~O'Malley, and B.S. Alexandrov.
\newblock Unsupervised machine learning based on non-negative tensor
  factorization for analyzing reactive mixing.
\newblock {\em Journal of Computational Physics}, 395:85 -- 104, 2019.

\bibitem{mudunuru2019mixing}
M.~K. Mudunuru and S.~Karra.
\newblock Physics-informed machine learning models for predicting the progress
  of reactive mixing.
\newblock {\em arXiv preprint arXiv:1908.10929v1}, 2019.

\bibitem{wu2018seismic}
Y.~Wu, Y.~Lin, Z.~Zhou, and A.~Delorey.
\newblock Seismic-net:~{A} deep densely connected neural network to detect
  seismic events.
\newblock {\em arXiv preprint arXiv:1802.02241}, 2018.

\bibitem{hulbert2019similarity}
C.~Hulbert, B.~Rouet-Leduc, P.~A. Johnson, C.~X. Ren, J.~Rivi\'{e}re, D.~C.
  Bolton, and C.~Marone.
\newblock Similarity of fast and slow earthquakes illuminated by machine
  learning.
\newblock {\em Nature Geoscience}, 12:69, 2019.

\bibitem{viswanathan2018graph}
H.~S. Viswanathan, J.~D. Hyman, S.~Karra, D.~O'Malley, S.~Srinivasan,
  A.~Hagberg, and G.~Srinivasan.
\newblock Advancing graph-based algorithms for predicting flow and transport in
  fractured rock.
\newblock {\em Water Resources Research}, 54:6085--6099, 2018.

\bibitem{srinivasan2018robust}
S.~Srinivasan, J.~Hyman, S.~Karra, D.~O'Malley, H.~S. Viswanathan, and
  G.~Srinivasan.
\newblock Robust system size reduction of discrete fracture networks:~{A}
  multi-fidelity method that preserves transport characteristics.
\newblock {\em Computational Geosciences}, 22:1515--1526, 2018.

\bibitem{camps2018physics}
G.~C.-Valls, L.~Martino, D.~H Svendsen, M.~C.-Taberner, J.~M.-Mar{\'\i},
  V.~Laparra, D.~Luengo, and F.~J. G.-Haro.
\newblock Physics-aware gaussian processes in remote sensing.
\newblock {\em Applied Soft Computing}, 68:69--82, 2018.

\bibitem{bergen2019machine}
K.~J. Bergen, P.~A. Johnson, V.~Maarten, and G.~C. Beroza.
\newblock Machine learning for data-driven discovery in solid earth geoscience.
\newblock {\em Science}, 363:eaau0323, 2019.

\bibitem{reichstein2019deep}
M.~Reichstein, G.~C.-Valls, B.~Stevens, M.~Jung, J.~Denzler, N.~Carvalhais, and
  Prabhat.
\newblock Deep learning and process understanding for data-driven earth system
  science.
\newblock {\em Nature}, 566:195, 2019.

\bibitem{requena2018predicting}
C.~R.-Mesa, M.~Reichstein, M.~Mahecha, B.~Kraft, and J.~Denzler.
\newblock Predicting landscapes as seen from space from environmental
  conditions.
\newblock In {\em IGARSS 2018-2018 IEEE International Geoscience and Remote
  Sensing Symposium}, pages 1768--1771. IEEE, 2018.

\bibitem{james2018_1}
S.~C. James, Y.~Zhang, and F.~O'Donncha.
\newblock A machine learning framework to forecast wave conditions.
\newblock {\em Coastal Engineering}, 137:1--10, 2018.

\bibitem{james2019_2}
F.~O'Donncha, Y.~Zhang, B.~Chen, and S.~C. James.
\newblock Ensemble model aggregation using a computationally lightweight
  machine-learning model to forecast ocean waves.
\newblock {\em Journal of Marine Systems}, 199:103206, 2019.

\bibitem{james2018_3}
F.~O'Donncha, Y.~Zhang, B.~Chen, and S.~C. James.
\newblock An integrated framework that combines machine learning and numerical
  models to improve wave-condition forecasts.
\newblock {\em Journal of Marine Systems}, 186:29 -- 36, 2018.

\bibitem{rouet2017machine}
B.~R.-Leduc, C.~Hulbert, N.~Lubbers, K.~Barros, C.~J. Humphreys, and P.~A.
  Johnson.
\newblock Machine learning predicts laboratory earthquakes.
\newblock {\em Geophysical Research Letters}, 44:9276--9282, 2017.

\bibitem{reynen2017supervised}
A.~Reynen and P.~Audet.
\newblock Supervised machine learning on a network scale:~{A}pplication to
  seismic event classification and detection.
\newblock {\em Geophysical Journal International}, 210:1394--1409, 2017.

\bibitem{magana2017explosion}
S.~A. M.-Zook and S.~D. Ruppert.
\newblock Explosion monitoring with machine learning:~{A LSTM} approach to
  seismic event discrimination.
\newblock In {\em AGU Fall Meeting Abstracts}, 2017.

\bibitem{yuan2019using}
B.~Yuan, Y.~J. Tan, M.~K. Mudunuru, O.~E. Marcillo, A.~A. Delorey, P.~M.
  Roberts, J.~D. Webster, C.~N.~L. Gammans, S.~Karra, G.~D. Guthrie, and P.~A.
  Johnson.
\newblock Using machine learning to discern eruption in noisy environments:~{A}
  case study using $\mathrm{CO}_2$-driven cold-water geyser in {C}himay{\'o},
  {N}ew {M}exico.
\newblock {\em Seismological Research Letters}, 90:591--603, 2019.

\bibitem{BARZEGAR2018697}
R.~Barzegar, A.~A. Moghaddam, R.~Deo, E.~Fijani, and E.~Tziritis.
\newblock Mapping groundwater contamination risk of multiple aquifers using
  multi-model ensemble of machine learning algorithms.
\newblock {\em Science of The Total Environment}, 621:697 -- 712, 2018.

\bibitem{RODRIGUEZGALIANO2015804}
V.~R.-Galiano, M.~S.-Castillo, M.~C.-Olmo, and M.~C.-Rivas.
\newblock Machine learning predictive models for mineral prospectivity:~{A}n
  evaluation of neural networks, random forest, regression trees and support
  vector machines.
\newblock {\em Ore Geology Reviews}, 71:804 -- 818, 2015.

\bibitem{KIRKWOOD201649}
C.~Kirkwood, M.~Cave, D.~Beamish, S.~Grebby, and A.~Ferreira.
\newblock A machine learning approach to geochemical mapping.
\newblock {\em Journal of Geochemical Exploration}, 167:49 -- 61, 2016.

\bibitem{Zuo2017}
R.~Zuo.
\newblock Machine learning of mineralization-related geochemical anomalies:~{A}
  review of potential methods.
\newblock {\em Natural Resources Research}, 26:457--464, 10 2017.

\bibitem{OONK201580}
S.~Oonk and J.~Spijker.
\newblock A supervised machine-learning approach towards geochemical predictive
  modelling in archaeology.
\newblock {\em Journal of Archaeological Science}, 59:80 -- 88, 2015.

\bibitem{cracknell2014mapping}
M.~J. Cracknell, A.~M. Reading, and A.~W. McNeill.
\newblock Mapping geology and volcanic-hosted massive sulfide alteration in the
  hellyer--mt charter region, tasmania, using random forests and
  self-organising maps.
\newblock {\em Australian Journal of Earth Sciences}, 61:287--304, 2014.

\bibitem{salah2018machine}
M.~K. Salah.
\newblock {\em Machine {L}earning for {M}odel {O}rder {R}eduction}.
\newblock Springer, 2018.

\bibitem{brunton2019data}
S.~L. Brunton and J.~N. Kutz.
\newblock {\em Data-{D}riven {S}cience and {E}ngineering:~{M}achine {L}earning,
  {D}ynamical {S}ystems, and {C}ontrol}.
\newblock Cambridge University Press, 2019.

\bibitem{marquaridt1970generalized}
D.~W. Marquaridt.
\newblock Generalized inverses, ridge regression, biased linear estimation, and
  nonlinear estimation.
\newblock {\em Technometrics}, 12:591--612, 1970.

\bibitem{hastie2009elements}
T.~Hastie, R.~Tibshirani, and J.~Friedman.
\newblock {\em The Elements of Statistical Learning:~Data mining, Inference,
  and Prediction}.
\newblock Springer Series in Statistics. Springer, New York, 2009.

\bibitem{murphy2012machine}
K.~P. Murphy.
\newblock {\em Machine {L}earning:~{A} {P}robabilistic {P}erspective}.
\newblock MIT press, 2012.

\bibitem{byr_2tipping2001}
M.~E. Tipping.
\newblock Sparse bayesian learning and the relevance vector machine.
\newblock {\em Journal of machine learning research}, 1:211--244, 2001.

\bibitem{breiman1996bagging}
L.~Breiman.
\newblock Bagging predictors.
\newblock {\em Machine learning}, 24:123--140, 1996.

\bibitem{adab4_freund1997}
Y.~Freund and R.~E. Schapire.
\newblock A decision-theoretic generalization of on-line learning and an
  application to boosting.
\newblock {\em Journal of computer and system sciences}, 55:119--139, 1997.

\bibitem{rumelhart1988learning}
D.~E. Rumelhart, G.~E. Hinton, and R.~J. Williams.
\newblock Learning representations by back-propagating errors.
\newblock {\em Cognitive modeling}, 5:1, 1988.

\bibitem{montavon2018methods}
G.~Montavon, W.~Samek, and K.~M{\"u}ller.
\newblock Methods for interpreting and understanding deep neural networks.
\newblock {\em Digital Signal Processing}, 73:1--15, 2018.

\bibitem{nakshatrala2013}
K.~B. Nakshatrala, M.~K. Mudunuru, and A.~J. Valocchi.
\newblock A numerical framework for diffusion-controlled bimolecular-reactive
  systems to enforce maximum principles and the non-negative constraint.
\newblock {\em Journal of Computational Physics}, 253:278--307, 2013.

\bibitem{2015_Mudunuru_etal_ASME}
M.~K. Mudunuru, M.~Shabouei, and K.~B. Nakshatrala.
\newblock {On local and global species conservation errors for nonlinear
  ecological models and chemical reacting flows}.
\newblock In {\em Proceedings of ASME 2015 International Mechanical Engineering
  Congress and Exposition}, pages V009T12A018--V009T12A018, 2015.

\bibitem{mudunuru2012framework}
M.~K. Mudunuru and K.~B. Nakshatrala.
\newblock A framework for coupled deformation-diffusion analysis with
  application to degradation/healing.
\newblock {\em International Journal for Numerical Methods in Engineering},
  89:1144--1170, 2012.

\bibitem{mudunuru2017mesh}
M.~K. Mudunuru and K.~B. Nakshatrala.
\newblock On mesh restrictions to satisfy comparison principles, maximum
  principles, and the non-negative constraint: Recent developments and new
  results.
\newblock {\em Mechanics of Advanced Materials and Structures}, 24:556--590,
  2017.

\bibitem{Pinder_Celia}
G.~F. Pinder and M.~A. Celia.
\newblock {\em {Subsurface Hydrology}}.
\newblock John Wiley \& Sons, Inc., New Jersey, USA, 2006.

\bibitem{2002_Adrover_etal_CCE_v26_p125_p139}
A.~Adrover, S.~Cerbelli, and M.~Giona.
\newblock {A spectral approach to reaction/diffusion kinetics in chaotic
  flows}.
\newblock {\em Computers \& Chemical Engineering}, 26:125--139, 2002.

\bibitem{2009_Tsang_PRE_v80_p026305}
Y.~K. Tsang.
\newblock {Predicting the evolution of fast chemical reactions in chaotic
  flows}.
\newblock {\em Physical Review E}, 80:026305(8), 2009.

\bibitem{mudunuru2016enforcing}
M.~K. Mudunuru and K.~B. Nakshatrala.
\newblock On enforcing maximum principles and achieving element-wise species
  balance for advection--diffusion--reaction equations under the finite element
  method.
\newblock {\em Journal of Computational Physics}, 305:448--493, 2016.

\bibitem{muller2016introduction}
A.~C. M{\"u}ller and S.~Guido.
\newblock {\em Introduction to {M}achine {L}earning with {P}ython:~{A} {G}uide
  for {D}ata {S}cientists}.
\newblock O'Reilly Media, Inc., 2016.

\bibitem{scikit-learn}
F.~Pedregosa, G.~Varoquaux, A.~Gramfort, V.~Michel, B.~Thirion, O.~Grisel,
  M.~Blondel, P.~Prettenhofer, R.~Weiss, V.~Dubourg, J.~Vanderplas, A.~Passos,
  D.~Cournapeau, M.~Brucher, M.~Perrot, and E.~Duchesnay.
\newblock Scikit-learn:~{M}achine learning in {P}ython.
\newblock {\em Journal of Machine Learning Research}, 12:2825--2830, 2011.

\bibitem{sklearn_api}
L.~Buitinck, G.~Louppe, M.~Blondel, F.~Pedregosa, A.~Mueller, O.~Grisel,
  V.~Niculae, P.~Prettenhofer, A.~Gramfort, J.~Grobler, R.~Layton,
  J.~VanderPlas, B.~Holt, and G.~Varoquaux.
\newblock {API} design for machine learning software:~{E}xperiences from the
  scikit-learn project.
\newblock In {\em ECML PKDD Workshop: Languages for Data Mining and Machine
  Learning}, pages 108--122, 2013.

\bibitem{scikit-learn_software}
F.~Pedregosa, G.~Varoquaux, A.~Gramfort, V.~Michel, B.~Thirion, O.~Grisel,
  M.~Blondel, P.~Prettenhofer, R.~Weiss, V.~Dubourg, J.~Vanderplas, A.~Passos,
  D.~Cournapeau, M.~Brucher, M.~Perrot, and E.~Duchesnay.
\newblock Scikit-learn:~{M}achine learning in {P}ython.
\newblock {\em Journal of Machine Learning Research}, 12:2825--2830, 2011.

\bibitem{kohavi1995study}
R.~Kohavi.
\newblock A study of cross-validation and bootstrap for accuracy estimation and
  model selection.
\newblock In {\em Ijcai}, volume~14, pages 1137--1145, 1995.

\bibitem{chou2014machine}
J.~S. Chou, C.~F. Tsai, A.~D. Pham, and Y.~H. Lu.
\newblock Machine learning in concrete strength simulations: Multi-nation data
  analytics.
\newblock {\em Construction and Building Materials}, 73:771--780, 2014.

\bibitem{logreg_3Molnar2019}
C.~Molnar.
\newblock {\em Interpretable {M}achine {L}earning:~{A} {G}uide for {M}aking
  {B}lack {B}ox {M}odels {E}xplainable}.
\newblock Christoph Molnar, 2019.

\bibitem{byr_1mackay1992}
D.~J.~C. MacKay.
\newblock Bayesian interpolation.
\newblock {\em Neural computation}, 4:415--447, 1992.

\bibitem{rasmussen2003gaussian}
C.~E. Rasmussen.
\newblock Gaussian processes in machine learning.
\newblock In {\em Summer School on Machine Learning}, pages 63--71. Springer,
  2003.

\bibitem{santner2018design}
T.~J. Santner, B.~J. Williams, and W.~Notz.
\newblock {\em The {D}esign and {A}nalysis of {C}omputer {E}xperiments},
  volume~1.
\newblock Springer, 2018.

\bibitem{rennie2003tackling}
J.~D. Rennie, L.~Shih, J.~Teevan, and D.~R. Karger.
\newblock Tackling the poor assumptions of na\"{i}ve {B}ayes text classifiers.
\newblock In {\em Proceedings of the 20th International Conference on Machine
  Learning (ICML-03)}, pages 616--623, 2003.

\bibitem{manning2010introduction}
C.~Manning, P.~Raghavan, and H.~Sch{\"u}tze.
\newblock Introduction to information retrieval.
\newblock {\em Natural Language Engineering}, 16:100--103, 2010.

\bibitem{mccallum1998comparison}
A.~McCallum and K.~Nigam.
\newblock A comparison of event models for na\"{i}ve {B}ayes text
  classification.
\newblock In {\em AAAI-98 workshop on learning for text categorization}, volume
  752, pages 41--48. Citeseer, 1998.

\bibitem{metsis2006spam}
V.~Metsis, I.~Androutsopoulos, and G.~Paliouras.
\newblock Spam filtering with na\"{i}ve {B}ayes-which na\"{i}ve {B}ayes?
\newblock In {\em CEAS}, volume~17, pages 28--69. Mountain View, CA, 2006.

\bibitem{zhang2004optimality}
H.~Zhang.
\newblock The optimality of na\"{i}ve {B}ayes.
\newblock {\em AA}, 1:3, 2004.

\bibitem{naive_bayes_mean_var_update}
Tony~F. C., Gene~H. G, and Randall~J. L.
\newblock Algorithms for computing the sample variance: Analysis and
  recommendations.
\newblock {\em The American Statistician}, 37(3):242--247, 1983.

\bibitem{breiman2001random}
L.~Breiman.
\newblock Random forests.
\newblock {\em Machine learning}, 45:5--32, 2001.

\bibitem{breiman2017classification}
L.~Breiman, J.~H. Friedman, R.~A. Olshen, and C.~J. Stone.
\newblock {\em Classification and {R}egression {T}rees}.
\newblock Taylor \& Francis Group, New York, 2017.

\bibitem{geurts2006extremely}
P.~Geurts, D.~Ernst, and L.~Wehenkel.
\newblock Extremely randomized trees.
\newblock {\em Machine learning}, 63:3--42, 2006.

\bibitem{adab1_freund1995}
Y.~Freund.
\newblock Boosting a weak learning algorithm by majority.
\newblock {\em Information and computation}, 121:256--285, 1995.

\bibitem{adab2_freund1996}
Y.~Freund and Robert~E. S.
\newblock Experiments with a new boosting algorithm.
\newblock In {\em icml}, volume~96, pages 148--156. Citeseer, 1996.

\bibitem{adab3_freund1996}
Y.~Freund and R.~E. Schapire.
\newblock Game theory, on-line prediction and boosting.
\newblock In {\em COLT}, volume~96, pages 325--332. Citeseer, 1996.

\bibitem{adab5_schapire1995}
R.~E. Schapire and Y.~Freund.
\newblock A decision-theoretic generalization of on-line learning and an
  application to boosting.
\newblock In {\em Second European Conference on Computational Learning Theory},
  pages 23--37, 1995.

\bibitem{relu}
V.~Nair and G.~E. Hinton.
\newblock Rectified linear units improve restricted boltzmann machines.
\newblock In {\em Proceedings of the 27th international conference on machine
  learning (ICML-10)}, pages 807--814, 2010.

\end{thebibliography}

%

\appendix
\section{Appendices}
\subsection{Nomenclature}
\subsubsection{Variables}
\noindent \textbf{A} = Diagonal matrix\\
$a_n^l$ = Activation function of neuron $n$ at layer $l$ \\
$b$ = Bias \\
$c_i$ [mol\,m$^{-3}$] = The molar concentration of chemical species $i$ \\
$c^{0}_i(\mathbf{x})$ = The initial concentration of chemical species $i$ \\
$c^{\mathrm{p}}_i(\mathbf{x},t)$ [mol\,m$^{-3}$] = Prescribed molar concentration \\
$\mathbf{D}(\mathbf{x},t)$ [s$^{2}\,$m$^{-1}$] = The anisotropic dispersion tensor \\
$D_\mathrm{m}$ [m$^{3}\,$s$^{-2}$] = The molecular diffusivity \\ 
$\mathrm{det}$  = Determinant \\
$\mathbb{E}$ = Expectation \\
$F$ = Activation function \\
$f$ = Function \\
$f_m$ = $m^\mathrm{th}$ classifier \\
$G$ = Impurity \\
$H$ = Gini impurity function \\
$H_e$  =  Truncation value for Huber loss \\
$h_m$ = Base learner/tree \\
$h^{\mathrm{p}}_i(\mathbf{x},t)$ [m\,s$^{-1}$] = Flux \\
$i$ =  Index \\
$\mathbf{I}$ = The identity tensor \\
$\mathrm{inf}$  = The greatest upper bound \\
$J = \mathrm{sup}\left|\hat{y_i} - y_i\right|$ \\
$j$ = Feature index \\
$k$ = Class variable index\\
$k_{AB}$ [m$^{-1}$] = The bi-linear reaction rate coefficient \\
$L$ = Loss function \\
$L_1$  = $L_1$ norm \\
$L_2$ = $L_2$ norm \\
$l$ = Layer index \\
$\mathrm{l}$ = The length of the `wiggles' in sine function \\
$M$ = Number of training data\\
$m$ = Tree node index\\
$N$ = Gaussian or normal distribution \\
$n$  = Node number \\
$n_A$ = Stoichiometric coefficient for species $A$\\
$n_B$ = Stoichiometric coefficient for species $B$\\
$n_C$ = Stoichiometric coefficient for species $C$\\
$o$ = Random coefficient \\
$p$ [\%] = Probability \\
$\mathrm{p}$ = The distance between repetitions of the sine function \\
$p_{mk}$ [\%] = Probability at $mk^\mathrm{th}$ leaf\\
$Q$ = Data at tree node $m$ \\
$q$ = Batch or subsample \\
$r$ = Random coefficient \\
$s$ = Split \\
$\mathrm{sup}$ = The least upper bound \\
$t_m$ = Threshold at which trees split\\
$u$ = Random coefficient \\
$\mathbf{v}$ [m\,s$^{-1}$] = The velocity vector field \\
$v_0$ = The perturbation parameter \\
$w$ = Coefficient \\ 
$\mathbb{w}$ = Coefficient vector \\
$w_0$ = Intercept \\
$\mathbb{X} $ = Feature matrix \\
$\mathbb{x}$ = Feature vector \\
$y$  = Label \\
$\mathbb{y}$ = Label vector \\
$\hat{y}$ = Approximation to $y$ \\
$z$ = The dummy variable \\
\subsubsection{Greek Symbols}
\noindent $\alpha$ = Penalty/regularization parameter \\
$\alpha_1$ = Regularization parameter for $w$ \\ 
$\alpha_2$ = Regularization parameter for $w$ \\
$\alpha_\mathrm{L}$ [m$^{2}\,$s$^{-1}$] = The longitudinal diffusivity \\ 
$\alpha_\mathrm{T}$ [m$^{2}\,$s$^{-1}$] = The transverse diffusivity \\
$\beta$ = Regularization parameter\\
$\Gamma^{\mathrm{D}}_{i}$  = Dirichlet  boundary condition \\
$\Gamma^{\mathrm{N}}_{i}$  = Neumann boundary condition \\ 
$\theta_m$ = Confidence in the prediction for $m^{\mathrm{th}}$ data\\
$\kappa_\mathrm{f}L$ and $T$ [--] = The characteristic spatial and temporal scales of the flow field \\ 
$\gamma$ = Learning rate \\
$\epsilon$ = Truncation value under which no penalty is associated with the training loss \\
$\eta$ = Noise \\
$\theta$ = Confidence function for prediction \\
$\mathscr{K}$ = Node number \\
$\mathcal{K}$ = Kernel \\
$\lambda$ = Spread of kernel \\
$\mu$ = Mean \\
$\nu$  = Integer \\
$\mathrm{\pi}$ = cos\textsuperscript{--1}(--1) \\
$\Sigma$ = The covariance matrix \\
$\boldsymbol{\Psi}$ = Vector of $\mathcal{N} \times 1 $ size \\ 
$\mathrm{\omega}$ = Regularization parameter \\
$\mathbb{I}$ = Identity matrix \\
$\mathfrak{T}$ = Number of regression tree \\
$\mathbf{\Phi}$  = Design matrix of $\mathcal{N}\times(\mathcal{N}+1)$ size \\
$\omega$ = Regularization parameter \\
\subsubsection{Mathematical Symbol}
\noindent $\otimes$ = The tensor product \\
$\mathbb{1}$ = Indicator function (either 1 or 0)\\
\subsection{Brief Mathematical Description of ML Emulators}
\subsubsection{Linear Emulators}
\subsubsection*{Generalized Linear Emulators}
Suppose there are $n$~features $x_1$ through $x_n$ that correspond to a label $y$. 
LSQR calculates the closest $\hat{y}$ by finding the best linear combination of features as:
\begin{equation}\label{eq:OLSQ_1}
    \hat{y}\left(\mathbb{w}, \mathbb{x}\right)  =  w_0 + w_1x_1 + \cdots + w_nx_n = \mathbb{x}\cdot \mathbb{w}.
\end{equation}
Linear regressors minimize a loss (or cost) function. 
The cost/loss function in this case is the residual sum of squares between a set of training feature vectors $\mathbb{x}_1, \mathbb{x}_2, \cdots, \mathbb{x}_m$ and predicted targets $y_1, y_2, \cdots, y_m$ of the form: 

\begin{equation}\label{eq:linear_ML_eqs}
    L_{\mathrm{lin}} = \underset{\mathbb{w}}{\mathrm{min}}||\mathbb{X}\mathbb{w} - \mathbb{y} ||_2^2 + \alpha_1 || \mathbb{w} ||_1 + \alpha_2 || \mathbb{w} ||_2^2 + \Sigma \sum_{i=1}^m \left[1 + H_\epsilon \left(\frac{\mathbb{x}_i \cdot \mathbb{w} - y_i}{\Sigma}\right) \right].
\end{equation}
For LSQR, $\alpha_1 = \alpha_2 = \Sigma = 0$, for RR, $\alpha_1 = \Sigma = 0$, for LR, $\alpha_2 = \Sigma = 0$, for ER, $\Sigma = 0$, and for HR, $\alpha_2 = 0$.
\begin{equation}
  H_\epsilon\left(e\right) = 
  \begin{cases} e^2 & \mathrm{if} \, e 
    < \, \epsilon \\ 
    2\epsilon |e| - \epsilon^2 & \mathrm{otherwise}\\
  \end{cases}.
\end{equation}

The LSQR method minimizes $L_\mathrm{lin}$ without regularization. 
RR uses the $L_2$ norm, which does not use sparsity constraints.
However, it includes a penalty $\alpha_2$ to weights, which is known as the ridge coefficient.
This prevents weights from getting too large as well as overfitting. 
LR is another linear regressor that penalizes the $L_1$ norm. 
This penalty $\alpha_1$ on the absolute value of weights results in sparse models tending toward small weights.
The $\alpha_1$ controls the strength of the regularization penalty, and more parameters are eliminated with increasing $\alpha_1$. 
With increasing $\alpha_1$, bias increases, but variances decrease and vice versa.
ER is another linear regressor that combines the $L_1$ and $L_2$ penalties of RR and LR.
It is useful for data with multiple features that are correlated with each other.
LR likely picks one of these correlated features at random, but ER picks all the correlated features.
HR is a generalized linear regression method that put a sample as an inlier, if the absolute error of that sample is less than the specified threshold.
HR puts a sample as an outlier, if absolute errors go beyond the specified threshold.
Polynomial regression applies the LSQR formula on quadratic scaled data.
%
\subsubsection*{Logistic Regression}
Despite its name, logistic regression is a classifier; it uses the linear regression scheme to correlate a probability for each class.
Logistic regression predicts the outcome in terms of probability and provides a meaningful threshold at which distinguishing between classes is possible \cite{logreg_3Molnar2019}.
Multi-class classification is achieved through either One-vs-One or One-vs-Rest strategy \cite{scikit-learn}.
A simple linear ML emulator fails to provide multi-class output as probabilities.
But the logistic regression provides the probabilities through the logistic function.
Consider an ML model with two features $x_1$ and $x_2$ with one label $y$, which is classified with a probability $p$.
If we assume a linear relationship between predictor variables and the log-odds of the event:
\begin{equation}\label{eq:logisticreg_1}
    \mathrm{ln}\frac{p}{1 - p} = w_0 + w_1 x_1 + w_2 x_2.
\end{equation}
With simple algebraic manipulation, the probability $p$ of classifying the predictor variable can be recast as:
\begin{equation}\label{eq:logisticreg_2}
    p  =  \frac{1}{1 + \mathrm{e}^{-\left(w_0 + w_1 x_1 + w_2 x_2\right)}}.
\end{equation}
Here, the loss function is defined by cross-entropy loss as:
\begin{equation}\label{eq:cross-entropy}
    L_{\operatorname{cross-entropy}} = -\frac{1}{n} \sum_{n=1}^n \left[y_n \mathrm{log}(p_n) + \left(1 - y_n\right) \mathrm{log}\left(1 - p_n\right) \right].
\end{equation}
%
\subsubsection*{Kernel Ridge (KR) Regression}
KR regression combines RR with kernel tricks \cite{murphy2012machine} to learn a linear function induced by both the kernel and data.
The kernel trick enables a linear ML emulator to learn nonlinear functions without explicitly mapping a linear learning algorithm.
The kernel function is applied on each label to map the original nonlinear observations into a higher-dimensional space.
In this work, the stationary radial basis function (RBF) kernel is the optimized kernel. 
The RBF kernel on two different feature vectors, $\mathbb{x_1}$ and $\mathbb{x_2}$, is:
\begin{equation}\label{eq:KRR_RBF}
    \mathcal{K}_{\mathrm{RBF}}\left(\mathbb{x_1}, \mathbb{x_2}\right)  =  \mathrm{exp}\left(-\lambda \left|\left|\mathbb{x_1} - \mathbb{x_2}\right|\right|^{2} \right).
\end{equation}
%
If the kernel is Gaussian then high $\lambda$ shrinks the spread of Gaussian distribution and vice versa.
The squared-loss function is used to learn the linear mapping function:
\begin{equation}
    L_{\mathrm{squared}} = \left(y - \hat{y}\right)^2.
\end{equation}
%
\subsubsection{Bayesian Emulators}
\subsubsection*{Bayesian Ridge (BR) Regression}
Using Bayes' Rule, BR formulates a probabilistic model of the regression problem.
BR assumes labels $\mathbb{y}$ as normally distributed around $\mathbb{X} \mathbb{w}$ and obtains a probabilistic model by:
\begin{equation}\label{eq:Bayesian_1}
    p\left(\mathbb{y}|\mathbb{X}, \mathbb{w}, \beta \right)  =  N\left(\mathbb{y}|\mathbb{X}\mathbb{w}, \beta \right).
\end{equation}
%
The prior for the coefficient vector $\mathbb{w}$ is given by a spherical Gaussian distribution:
\begin{equation}\label{eq:Bayesian_2}
    p\left(\mathbb{w}|\omega \right)  =  N \left(\mathbb{w}|0, \omega^{-1} \mathbf{\mathbb{I}}\right).
\end{equation}
The $\beta$ and $\omega$ are selected to be conjugate priors and gamma distributions. 
The parameters $\beta$ and $\omega$ are estimated by maximizing the log-marginal likelihood \cite{byr_1mackay1992,byr_2tipping2001} as:
\begin{equation}\label{eq:lml}
    L_{\mathrm{lml}} = -\frac{1}{2}\left[\log_{10}\|\omega^{-1}\mathbb{I} + \mathbf{\Phi} \mathbf{A}^{-1} \mathbf{\Phi}^\intercal \| + \mathbf{\Psi}^{\intercal}\left(\omega^{-1}\mathbb{I} + \mathbf{\Phi} \mathbf{A}^{-1} \mathbf{\Phi}^\intercal\right)^{-1}\mathbf{\Psi}\right] + \sum_{i=0}^n \left(o \mathrm{log}\beta_i - \mathrm{r}\beta_i\right) + u\log\omega - w\omega.
\end{equation}
%
\subsubsection*{Gaussian Process (GP)}
GPs are generic supervised learning methods for prediction and probabilistic classification that use properties inherited from the normal distribution.
GP has the capability of using kernel tricks, which differentiate GP from BR.
GP emulators are not sparse; as a result they are computational inefficient when developing models in high-dimensional spaces. 
That is, computing GP emulators are difficult to implement if features exceed a few dozens \cite{rasmussen2003gaussian,santner2018design} in \textsf{Scikit-learn} and our training sample size is $\mathcal{O}(10^5)$.
In this study, the RBF kernel (Eq.~(\ref{eq:KRR_RBF})) is used to obtain GP emulators by maximizing the first term on the right of Eq.~(\ref{eq:lml}) to predict $\mathbb{\hat{y}}$. 
\subsubsection*{Na\"{\i}ve Bayes (NB)} 
NB emulators are supervised ML methods that also apply Bayes' Theorem with the na\"{\i}ve assumption of conditional independence between every pair of features given the label value \cite{rennie2003tackling,manning2010introduction,mccallum1998comparison,metsis2006spam}. 
NB maximizes $p\left(x_i\,|\,y\right)$ and $p\left(y\right)$ by maximizing the a posteriori function \cite{scikit-learn}. 
Various na\"{\i}ve Bayes regressions differ by the assumptions they make regarding the distribution of $p\left(x_i\,|\,y\right)$ \cite{zhang2004optimality}. 
Herein, we use Gaussian-na\"{\i}ve Bayes emulator:
\begin{equation}\label{eq:Naive_bayes_6}
    p\left(x_i\,|\,y\right)  =  \frac{1}{\sqrt{2 \pi \sigma_y^2}} \exp{\left[-\frac{\left(x_i - \mu_y\right)^2}{2 \sigma_y^2}\right]}.
\end{equation}
NB updates model parameters such as feature means and variance using different batch sizes, which makes NB computationally efficient \cite{naive_bayes_mean_var_update}. 
%
\subsubsection*{Linear and Quadratic Discriminant Analyses (LDA/QDA)}
LDA and QDA are classifiers that use Bayes rule.
They compute the class conditional distribution of data $p\left(\mathbb{x}|y=k\right)$ for each class $k$. 
Based on $p\left(\mathbb{x}|y=k\right)$, for partition $y = q$ of sample space, predictions are made using Bayes' rule:
\begin{equation}\label{eq:lda_1}
    p\left(y=k|\mathbb{x}\right)  = \frac{p\left(\mathbb{x}|y=k\right) p\left(y=k\right)}{p\left(\mathbb{x}\right)}  = \frac{p\left(\mathbb{x}|y=k\right) p\left(y=k\right)}{\sum \limits_q p\left(\mathbb{x}|y=q\right) p\left(y=q\right)}.
\end{equation}
Later, class $k$ is selected to maximize the conditional probability. 
Specifically, $p\left(\mathbb{x}|y\right)$ is modeled using a multivariate Gaussian distribution with density:
\begin{equation}\label{eq:lda_2}
    p\left(y = k|\mathbb{x}\right)  = \frac{1}{\left(2 \pi \right)^{j/2} \left| \mathrm{det}[\sum_k] \right|^{1/2}} \exp{\left[-\frac{1}{2} \left(\mathbb{x} - \mu_k \mathbb{1} \right) \cdot ({\scriptstyle\sum}_k)^{-1} \left(\mathbb{x} - \mu_k \mathbb{1} \right)\right]}.
\end{equation}
%
Using training data, it estimates the class priors $p\left(y=k\right)$, class means $\mu_k$, and covariance matrices ${\scriptstyle\sum}_k$ either by the empirical sample class covariance matrices or by a regularized estimator. 
In LDA, each class shares the same covariance matrix (i.e.,~${\scriptstyle\sum}_k = {\scriptstyle\sum}$), which leads to linear decision surface \cite{hastie2009elements}:
\begin{equation}\label{eq:lda_3}
\begin{split}
        \mathrm{log}\left(\frac{p\left(y=k|\mathbb{x}\right)}{p\left(y=q|\mathbb{x}\right)}\right)  =  \mathrm{log}\left(\frac{p\left(\mathbb{x}|y=k\right) p \left(y=k\right)}{p\left(\mathbb{x}|y=q\right) p \left(y=q\right)}\right)  = 0 \iff \\ \left(\mu_k - \mu_q\right) \mathbb{1} \cdot {\scriptstyle\sum}^{-1} \mathbb{x} = \frac{1}{2} \left(\mu_k \mathbb{1} \cdot {\scriptstyle\sum}^{-1} \mu_k \mathbb{1} - \mu_q \mathbb{1} \cdot {\scriptstyle\sum}^{-1} \mu_q \mathbb{1} \right) - \mathrm{log}\frac{p\left(y=k\right)}{p\left(y=q\right)}.
\end{split}
\end{equation}
However, QDA does not assume covariance matrices of the Gaussian's, which leads to a quadratic decision surface \cite{hastie2009elements}. 
Both LDA and QDA use the cross-entropy loss function Eq.~(\ref{eq:cross-entropy}). 
%
\subsubsection{Ensemble ML Emulators}
\subsubsection*{Decision Tree (DT)}
DT is interpretable as a weak ML classifier and regressor. 
DTs split leaves in a tree and find the best or optimal split $s^*$ that increases the purity/accuracy of the resulting tree \cite{RODRIGUEZGALIANO2015804,breiman2001random,breiman2017classification,geurts2006extremely}. 
A single tree reduces error in a locally optimal way during feature space splitting while a regression tree minimizes the residual squared error.
For $n$ pairs of training samples, the DT recursively partitions the space to bring the same labels under the same group.
Let data at node $m$ be represented by $Q$.
For each candidate split $s = \left(j, \epsilon_m\right)$ consisting of feature $j$ and threshold $\epsilon_m$, DT splits data into $Q_{\mathrm{left}}(s)$ and $Q_{\mathrm{right}}(s)$ subsets.
For regression, the impurity at $m$ is computed using the Gini impurity function $H\left(\mathbb{X}_m\right) = \frac{1}{\mathfrak{T}_m} \sum\limits_{i \in \mathfrak{T}_m} \left(y_i - \hat{y}_i\right)^2$ using:
\begin{equation}\label{eq:DT_splitting}
    G\left(Q, s\right) = \frac{n_{\mathrm{left}}}{\mathfrak{T}_m} H\left(Q_{\mathrm{left}}(s)\right) + \frac{n_{\mathrm{right}}}{\mathfrak{T}_m} H\left(Q_{\mathrm{right}}(s)\right),
\end{equation}
where $\mathfrak{T}_m \le \mathrm{min}_{\mathrm{samples}}$ or $\mathfrak{T}_m = 1$.
Then, the DT selects the parameters that minimize the impurity:
\begin{equation}
    s^* = \underset{s}{\mathrm{argmin}} \, G\left(Q, s\right).
\end{equation}
DT recursively find $Q_{\mathrm{left}}\left(s^*\right)$ and $Q_\mathrm{right}\left(s^*\right)$ till to reach maximum allowable depth, $\mathfrak{T}_m < \mathrm{min}_\mathrm{samples}$ or $\mathfrak{T}_m = 1$.
For regression, the loss function, $L_{\mathrm{MSE}}$, is defined as mean squared error (MSE) between the high-fidelity simulations and the ML emulators:
\begin{equation}\label{eq:mse}
    L_{\mathrm{MSE}} = \frac{1}{n} \sum_{i=1}^n \left( y_i - \hat{y_i}\right)^2.
\end{equation}
%
\subsubsection*{Bagging Emulator}
Bagging is a simple ensemble technique that builds on many independent tree/predictors and combines them using various model averaging techniques such as a weighted average, majority vote, or arithmetic average.
For $n$ pairs of training samples, bagging (bootstrap aggregating) selects $M$ set of samples from $n$ with replacement. 
Based on each sample, it trains functions $f_1\left(\mathbb{x}_1\right), ...,f_M\left(\mathbb{x}_M\right)$. 
Then, these individual functions or trees are aggregated for regression as:
\begin{equation}
    \label{eq:bagging_1}
    \hat{f}  =  \sum_{i=1}^M f_i (\mathbb{x}_i).
\end{equation}
The optimized regression criteria, or loss function, to select locations for splits is $L_\mathrm{MSE}$ (see Eq.~(\ref{eq:mse})).
%
\subsubsection*{Random Forest (RF)}
RF is a model-free ensemble emulator, which provides good accuracy by combining the performance of numerous DTs to classify or predict the value of a variable \cite{breiman2001random,breiman2017classification}. 
For given input data (e.g., feature vector $\mathbb{x}$), RF builds a number of regression trees ($M$) and averages the results. 
For each tree $\mathfrak{T}_m(\mathbb{x})$ for all $m = 1, 2, \cdots, M$, the RF prediction is:
\begin{equation}
    \hat{f}_{rf}^M = \frac{1}{M} \sum_{m=1}^M \mathfrak{T} \left( \mathbb{x} \right).
\end{equation}
For classification, the Gini impurity function is used for the loss function and for $k$ class variables, the Gini impurity is:
\begin{equation}
    H\left(\mathbb{x}_m\right) = \sum_k p_{mk} \left(1 - p_{mk}\right).
\end{equation}
%
For regression, $L_\mathrm{MSE}$ (Eq.~(\ref{eq:mse})) is used for the loss function.
%
\subsubsection*{AdaBoost (AdaB)} 
AdaB (aka Adaptive Boosting) converts weak learners into a strong learners \cite{adab1_freund1995,adab2_freund1996,adab3_freund1996,adab4_freund1997,adab5_schapire1995}. 
Weak learners are DTs with a single split that are also known as decision stumps.
AdaB is a greedy and forward stage-wise additive model (adding up multiple models to create a composite model) with an exponential loss function that iteratively fits a weak classifier to improve the current estimator.
AdaB puts more weight on difficult-to-learn labels and less on others. 
AdaB construct a tree regressor, $f_m$, from training data so that $f_m:\mathbb{x} \rightarrow y$.
Every pair of training data is passed through $f_m$.
Then, $f_m$ calculates a loss for each training datum using the square-loss function:
\begin{equation}
    L_i = \frac{\left|\hat{y_i} - y_i\right|^2}{J^2}.
\end{equation}
%
Then, the $L_i$ is averaged by $\hat{L} = \sum\limits_{i=1}^{n} L_i p_i$ to measure confidence in the prediction as:
\begin{equation}
    \theta = \frac{\hat{L}}{1 - \hat{L}}.
\end{equation}
%
The resulting $\theta$ is used to update weights: $w_i \rightarrow w_i \theta \mathrm{exp}\left(1 - L_i\right)$.
For $\mathbb{x}_i$, each of $M$ trees/regressors makes a prediction $h_m$, $m = 1,\cdots,
T$, to form a cumulative function:
\begin{equation}
    f = \mathrm{inf}\left[y \in \mathbb{y}: \sum_{m:h_m \leq y} \log\left(\frac{1}{\theta_m}\right) \geq \frac{1}{2} \sum_m \log\left(\frac{1}{\theta_m}\right) \right].
\end{equation}

DT-based AdaB is a heterogeneous emulator that applies both DT and boosting base estimators to learn a prediction function.

\subsubsection*{Gradient Boosting Method (GBM)}
GBM learns function like AdaB, but it generalizes the model by allowing optimization of an arbitrary differentiable loss function.
GBM builds learning function $f$ for $M$ trees as:
\begin{equation}\label{eq:AdaBoost_1}
    f = \sum_{m=1}^M \gamma_m h_m \left(\mathbb{x}_m\right).
\end{equation}
%
After learning each weak model, the additive model ($f_m$) is built in a greedy fashion:
\begin{equation}
    f_m = f_{m-1} + \gamma_m h_m,
\end{equation}
where the newly added tree minimizes the least-squared function, $L_\mathrm{lsqr}$, for previous model $f_{m-1}$ by:
\begin{equation}\label{eq:lsqr_loss}
    L_{\mathrm{lsqr}} = \underset{w}{\mathrm{min}} \sum_{i=1}^n \left(\mathbb{x}_i w - y_i \right)^2,
\end{equation}
where $i=1,\cdots,n$. 
The new learner is:
\begin{equation}
    h_m = \underset{h}{\mathrm{arg\,min}}\,\sum_{1=1}^n L_{\mathrm{lsqr}}\left[y_i, f_{m-1}\left(\mathbb{x}_i\right) + h\left(\mathbb{x}_i\right)\right].
\end{equation}
GBM minimizes the $L_{\mathrm{lsqr}}$ (optimal loss function for this work) by using steepest descent where the steepest descent direction is the negative gradient of the $L_{\mathrm{lsqr}}$ determined at the $f_{m-1}$.
The steepest gradient direction and rate is calculated by:
\begin{equation}
    \gamma_m = \underset{\gamma}{\mathrm{arg\,min}}\,\sum_{i=n}^n L_{\mathrm{lsqr}} \left(y_i, f_{m-1}\left(\mathbb{x}_i\right) - \gamma \frac{\partial L_{\mathrm{lsqr}} \left(y_i, f_{m-1}\left(\mathbb{x}_i\right)\right)}{\partial f_{m-1}\left(\mathbb{x}_i\right)}\right).
\end{equation}
%
\subsubsection{Multi-layer Perceptron (MLP)}
An MLP is a supervised ML method for classification and prediction.
MLPs are feed-forward neural networks or neural networks that are generalizations of linear models for prediction after multi-processing stages \cite{muller2016introduction}.
MLPs consist of numerous simple computation elements called neurons arranged in layers.
Neuron output is calculated as the result from a nonlinear activation function whose input is the sum of weighted inputs from all neurons in the preceding layer.
The out from neuron $n$ in layer $l$ is: 
\begin{equation}
    a_n^{\left(l\right)} = F\left(\sum_{\mathscr{K}=1}^{\mathscr{N}_{l-1}}w_{\mathscr{K},n}^{\left(l\right)} a_\mathscr{K}^{\left(l-1\right)} + b_n^{\left(l\right)}\right).
\end{equation}
%
%
The rectified linear unit (ReLU) \cite{relu} is the optimal activation function for this work: 
\begin{equation}
    f(z) = \mathrm{max}\left(0, z\right).
\end{equation}
%
The MSE function (see Equations~(\ref{eq:mse})) and cross-entropy function (see Equation~(\ref{eq:cross-entropy})) are optimal loss functions for regression and classification, respectively. 

\end{document}